%% file: main.tex
\pdfoutput=1

\documentclass[11pt]{article}
\usepackage[table,dvipsnames]{xcolor}
\usepackage{collcell}

\usepackage[]{acl}

\usepackage{times}
\usepackage{latexsym}

\usepackage[T1]{fontenc}

\usepackage[utf8]{inputenc}
\usepackage{paralist}
\usepackage{microtype}

\usepackage{inconsolata}

\usepackage{times,latexsym}
\usepackage{url}
\usepackage{pgf}
\usepackage{booktabs}

\usepackage{expex}
\usepackage[linguistics]{forest}

\usepackage{extdash} 

\definecolor{Gray}{gray}{0.9}
\definecolor{cb-blue-green} {RGB}{ 0,  073,  073}
\definecolor{cb-green-sea}  {RGB}{ 0, 146, 146}
\definecolor{cb-rose}       {RGB}{255, 109, 182}
\definecolor{cb-salmon-pink}{RGB}{255, 182, 119}
\definecolor{cb-purple}     {RGB}{ 73,   0, 146}
\definecolor{cb-blue}       {RGB}{ 0, 109, 219}
\definecolor{cb-lilac}      {RGB}{182, 109, 255}
\definecolor{cb-blue-sky}   {RGB}{109, 182, 255}
\definecolor{cb-blue-light} {RGB}{182, 219, 255}
\definecolor{cb-burgundy}   {RGB}{146,   0,   0}
\definecolor{cb-brown}      {RGB}{146,  73,   0}
\definecolor{cb-clay}       {RGB}{219, 209,   0}
\definecolor{cb-green-lime} {RGB}{ 36, 255,  36}
\definecolor{cb-yellow}     {RGB}{255, 255, 109}
\definecolor{cb-grey}       {RGB}{233, 233, 233}

\usepackage{hhline}
\usepackage{amsmath}
\usepackage{amssymb}
\usepackage{multirow}
\usepackage{arydshln}
\usepackage{enumitem}
\usepackage{multicol}
\usepackage{xspace}
\usepackage{wasysym}
\usepackage{hyperref}

\usepackage{graphicx,scalerel} 
\graphicspath{{./images/}} 
\usepackage{adjustbox}
\usepackage{graphicx}
\usepackage{subcaption}

\newcommand\rotation{30}

\usepackage{tikz}
\usepackage{collcell}

\newcommand*{\MinNumber}{0.0}%
\newcommand*{\MidNumber}{60.0} %
\newcommand*{\MaxNumber}{100.0}%

\newcommand{\ApplyGradient}[1]{%
        \ifdim #1 pt > \MidNumber pt
            \pgfmathsetmacro{\PercentColor}{max(min(100.0*(#1 - \MidNumber)/(\MaxNumber-\MidNumber),100.0),0.00)} %
            \hspace{-0.33em}\colorbox{SeaGreen!\PercentColor!Goldenrod!50}{#1}
        \else
            \pgfmathsetmacro{\PercentColor}{max(min(100.0*(\MidNumber - #1)/(\MidNumber-\MinNumber),100.0),0.00)} %
            \hspace{-0.33em}\colorbox{Red!\PercentColor!Goldenrod!50}{#1}
        \fi
}

\newcolumntype{R}{>{\collectcell\ApplyGradient}c<{\endcollectcell}}

\newcommand{\logo}[0]{\raisebox{-.25\height}{\includegraphics[width=.05\textwidth]{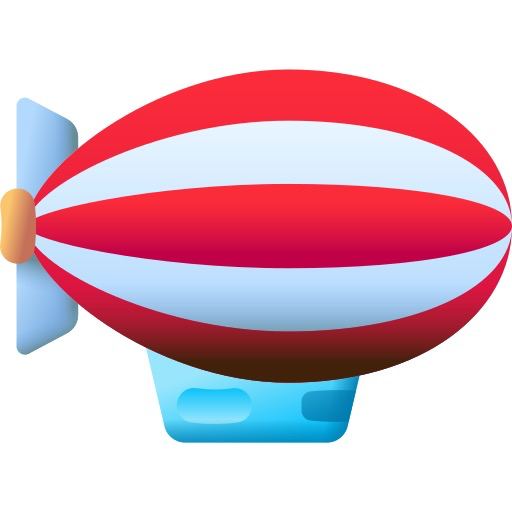}}}

\input{parts/notations}

\title{\logo \hspace{0.1em} RuBLiMP: Russian Benchmark of Linguistic Minimal Pairs}

\author{Ekaterina Taktasheva$^{1}$\thanks{\ \ Equal contribution.}, Maxim Bazhukov$^{2*}$, Kirill Koncha$^{3,4*}$\thanks{\ \ Work is partially done while at HSE University.},  \\
    \textbf{Alena Fenogenova$^{5}$}, \textbf{Ekaterina Artemova$^{6}$},  \textbf{and Vladislav Mikhailov$^{7}$} \\
    \textsuperscript{1}University of Edinburgh,
    \textsuperscript{2}HSE University,
    \textsuperscript{3}University of Groningen, \\
    \textsuperscript{4}Ghent University,
    \textsuperscript{5}SaluteDevices,
    \textsuperscript{6}Toloka AI,
    \textsuperscript{7}University of Oslo
}

\begin{document}
\maketitle
\begin{abstract}
Minimal pairs are a well-established approach to evaluating the grammatical knowledge of language models. However, existing resources for minimal pairs address a limited number of languages and lack diversity of language-specific grammatical phenomena. This paper introduces the \textbf{Ru}ssian \textbf{B}enchmark of \textbf{Li}nguistic \textbf{M}inimal \textbf{P}airs (RuBLiMP), which includes 45k pairs of sentences that differ in grammaticality and isolate a morphological, syntactic, or semantic phenomenon. In contrast to existing benchmarks of linguistic minimal pairs, RuBLiMP is created by applying linguistic perturbations to automatically annotated sentences from open text corpora and decontaminating test data. We describe the data collection protocol and present the results of evaluating 25 language models in various scenarios. We find that the widely used LMs for Russian are sensitive to morphological and agreement-oriented contrasts, but fall behind humans on phenomena requiring the understanding of structural relations, negation, transitivity, and tense. RuBLiMP, the codebase, and other materials are publicly available.
\end{abstract}

\input{parts/introduction}

\input{parts/design}

\begin{figure*}[ht!]
    \centering
    \includegraphics[width=\textwidth]{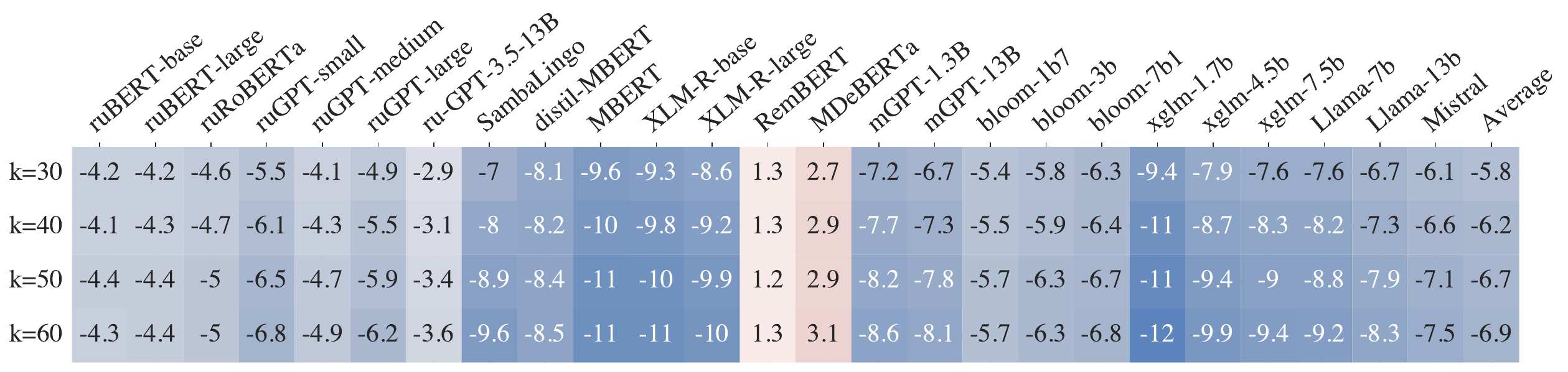}
    \caption{$\Delta$-scores ({$\downarrow$}) for each LM and \textsc{K\%} $\in \{30, 40, 50, 60\}$. All values are in \%.}
    \label{fig:ablation_k}
\end{figure*}

\input{parts/setup}

\input{parts/min_kablations}

\input{parts/monolingual_experiments}

\input{parts/multilingual_experiments}

\input{parts/related_work}

\input{parts/conclusion}

\input{parts/limitations}

\input{parts/ethics_statement}


\bibliography{anthology,custom}

\newpage
\appendix

\input{parts/appendix_examples}

\newpage
\clearpage
\input{parts/appendix_documentation}

\newpage
\clearpage
\input{parts/appendix_data_validation_guidelines}

\newpage
\clearpage
\input{parts/appendix_data_validation_results}

\newpage
\clearpage
\input{parts/appendix_data_stats}

\newpage
\clearpage
\input{parts/appendix_human_baseline}

\newpage
\clearpage

\input{parts/appendix_add_results}

\newpage
\clearpage

\twocolumn

\input{tables/cross_ling_all}

\input{parts/appendix_multilingual_experiments}

\end{document}

%% file: parts/notations.tex
\def\rublimp{RuBLiMP\xspace}
\def\Nparadigms{45\xspace}
\def\Nphenomena{12\xspace}

\def\dmbert{\texttt{distil-MBERT}\xspace}
\def\mbert{\texttt{MBERT}\xspace}
\def\xlmrb{\texttt{XLM-R}$_{\small{\texttt{base}}}$\xspace}
\def\xlmrl{\texttt{XLM-R}$_{\small{\texttt{large}}}$\xspace}
\def\rembert{\texttt{RemBERT}\xspace}
\def\mdeberta{\texttt{MDeBERTa}\xspace}
\def\mgpt{\texttt{mGPT-1.3B}\xspace}
\def\mgptb{\texttt{mGPT-13B}\xspace}
\def\blooms{\texttt{bloom-1b7}\xspace}
\def\bloomm{\texttt{bloom-3b}\xspace}
\def\blooml{\texttt{bloom-7b1}\xspace}
\def\xglms{\texttt{xglm-1.7B}\xspace}
\def\xglmm{\texttt{xglm-4.5B}\xspace}
\def\xglml{\texttt{xglm-7.5B}\xspace}
\def\rubertb{\texttt{ruBERT-base}\xspace}
\def\rubertl{\texttt{ruBERT-large}\xspace}
\def\ruroberta{\texttt{ruRoBERTa}\xspace}
\def\smblingorb{\texttt{SambaLingo}\xspace}
\def\rugpts{\texttt{ruGPT-small}\xspace}
\def\rugptm{\texttt{ruGPT-medium}\xspace}
\def\rugptl{\texttt{ruGPT-large}\xspace}
\def\rugpttpf{\texttt{ruGPT-3.5-13B}\xspace}
\def\llamas{\texttt{Llama-7b}\xspace}
\def\llamam{\texttt{Llama-13b}\xspace}
\def\mistral{\texttt{Mistral}\xspace}

\def\blimp{BLiMP\xspace}
\def\climp{CLiMP\xspace}
\def\sling{SLING\xspace}
\def\clams{CLAMS\xspace}
\def\jblimp{JBLiMP\xspace}

%% file: parts/introduction.tex
\section{Introduction}
\begin{figure*}[t!]
    \centering
    \includegraphics[width=0.99\textwidth]{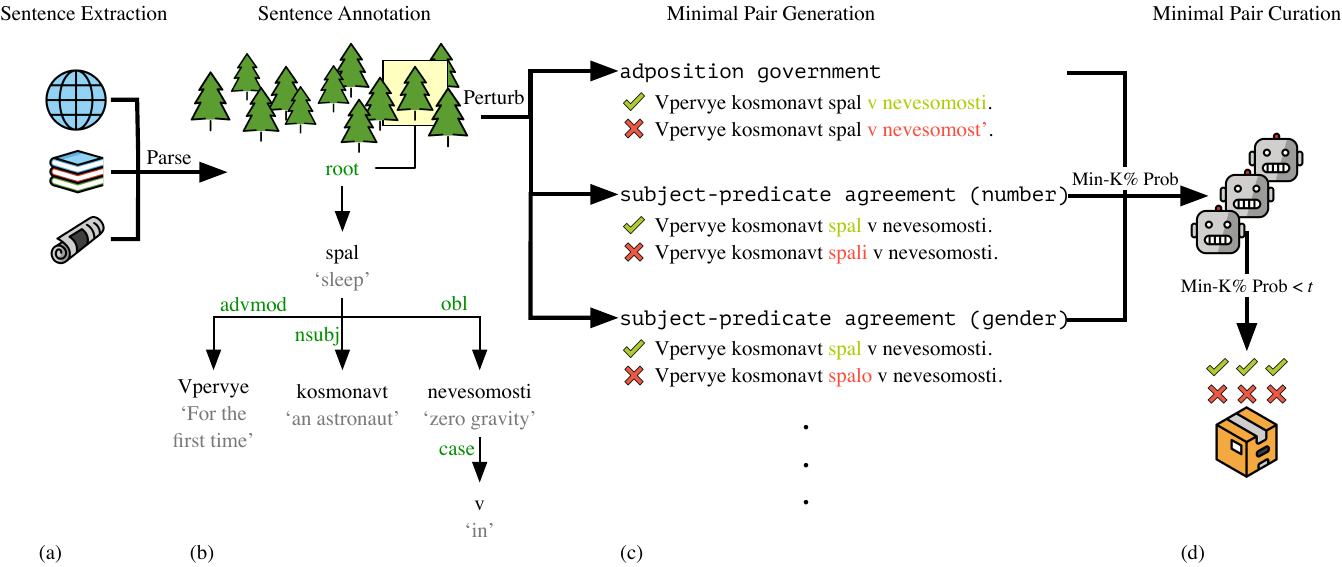}
    \caption{Overview of the RuBLiMP's minimal pair generation approach. \underline{Example}: \textit{Vpervye kosmonavt spal v nevesomosti} ``For the first time an astronaut slept in zero gravity''. (a) Extract sentences from publicly available corpora of Wikipedia texts, news articles, and books. (b) Annotate each extracted sentence in the Universal Dependencies scheme \cite{nivre-etal-2017-universal} with a multidomain morphosyntactic parser for Russian \cite{anastasyev2020exploring}. (c) Search the dependency trees for specific lexical units and linguistic structures and apply expert-written perturbation rules to create a pool of minimal pairs for a target paradigm. (d) Compute \textsc{Min-K\% Prob} \cite{shi2023detecting} for each grammatical sentence in the pool using a set of LMs. Select $t$ (the threshold for the maximum \textsc{Min-K\% Prob} value), which allows to find an intersection of 1k minimal pairs between the LMs. The minimal pairs in the intersection contain grammatical sentences that are not detected as the LMs' pretraining examples.
    }
    \label{fig:data_generation}
\end{figure*}

Acceptability judgments are the main empirical test in generative linguistics for assessing humans' linguistic competence and language acquisition \cite{chomsky1965aspects,schutze1996empirical}. One of the well-established approaches to judging a sentence's acceptability is a forced choice between \emph{minimal pairs} of sentences, where a native speaker is expected to prefer a grammatical sentence to an ungrammatical one, as in Example \ref{ex}.


\pex[*,labeloffset=0.3em,interpartskip=0.2ex,aboveexskip=2ex,belowexskip=2ex]
\a The cat \colorbox{cb-brown!20}{is} on the mat. (grammatical)
\a \ljudge{*}The cat \colorbox{cb-brown!20}{are} on the mat. (ungrammatical)
\xe
\label{ex}

\vspace{-1.3em}

The paradigm of minimal pairs has been widely adopted for evaluating the grammatical knowledge of language models (LMs) across various linguistic phenomena \cite{linzen-etal-2016-assessing,marvin-linzen-2018-targeted,wilcox-etal-2018-rnn,warstadt-etal-2019-investigating,warstadt-etal-2020-blimp-benchmark}. The evaluation design implies that an LM assigns a higher probability to the grammatical sentence than the ungrammatical one if it is sensitive to the isolated phenomenon. Over the last few years, a broad range of LMs has been analyzed via this paradigm in typologically diverse languages, except for Russian \citep[e.g.,][]{hartmann-etal-2021-multilingual,perez-mayos-etal-2021-assessing,leong2023bhasa}. 

\input{tables/blimps} 

This paper introduces the \textbf{Ru}ssian \textbf{B}enchmark of \textbf{Li}nguistic \textbf{M}inimal \textbf{P}airs (\rublimp), which consists of {\Nparadigms} datasets, each including 1k minimal pairs. Our benchmark covers morphological, syntactic, and semantic phenomena well-represented in Russian theoretical linguistics. In contrast to existing benchmarks of linguistic minimal pairs (see \autoref{tab:blimps}), RuBLiMP is created by (i) extracting sentences from open text corpora across multiple domains, (ii) annotating the sentences with one of the state-of-the-art multidomain morphosyntactic parsers, (iii) creating minimal pairs by perturbing the annotated sentences with expert-written rules, and (iv) discarding the pairs if the grammatical sentence is detected as a pretraining corpus example for at least one of 25 widely used LMs for Russian. Our method allows for generating minimal pairs at scale and ensures high customizability w.r.t. domain, dataset size, and LMs. Validating RuBLiMP by 20 native speakers with a background in linguistics confirms that the generated minimal pairs unambiguously isolate the target phenomenon and contrast in grammaticality.

Our main \emph{contributions} are: (i) we create RuBLiMP, the first diverse and large-scale benchmark of minimal pairs in Russian, (ii) we conduct ablation studies to analyze the effect of pretraining data decontamination on the model performance, (iii) we evaluate 25 monolingual and cross-lingual Transformer LMs~\cite{vaswani2017attention} and crowdsourcing workers, (iv) we release RuBLiMP\footnote{\href{https://huggingface.co/datasets/RussianNLP/rublimp}{\texttt{hf.co/datasets/RussianNLP/rublimp}}}, our codebase\footnote{\href{https://github.com/RussianNLP/RuBLiMP}{\texttt{github.com/RussianNLP/RuBLiMP}}}, and all data collection, data annotation, and other materials.

%% file: tables/blimps.tex
\begin{table}[t!]
\centering

\resizebox{\columnwidth}{!}{
    \begin{tabular}{llrrl}
    \toprule
    & \textbf{Language} &  \textbf{Size} & \textbf{\# Paradigm} & \textbf{Method}  \\
    \midrule
    \textbf{BLiMP} & English & 67k & 67 & Dictionary \& templates \\
    \textbf{CLiMP} & Chinese & 16k & 16 & Translation \& templates  \\
    \textbf{JBLiMP} & Japanese & 331 & 39 & Extract from articles \\
    \textbf{SLING} & Chinese & 38k & 38 & UD Treebank \& templates \\
    \textbf{NoCoLA}$_\text{zero}$ & Norwegian & 99.1k & 11 & Extract from an L2 corpus \\
    \textbf{DaLAJ} & Swedish & 4.8k & 4 & Extract from an L2 corpus \\
    \vspace{-1em}\\
    \hdashline
    \vspace{-1em}\\
    \multirow{2}{*}{\textbf{LINDSEA}} & Indonesian & 380 & 38 & \multirow{2}{*}{Expert-written min. pairs}  \\ 
     & Tamil & 200 & 20  \\
    \vspace{-1em}\\
    \hdashline
    \vspace{-1em}\\
    \multirow{5}{*}{\textbf{CLAMS}} & English & 153.5k & 13 & \multirow{5}{*}{Translation \& templates} \\
        & French & 49.3k & 7  \\
        & German & 47.8k & 7  \\
        & Hebrew & 40.8k & 7   \\
        & Russian & 40.1k & 7  \\
    \vspace{-1em}\\
    \hdashline
    \vspace{-1em}\\
    \multirow{3}{*}{\textbf{RuBLiMP}} & \multirow{3}{*}{Russian} & \multirow{3}{*}{45k} & \multirow{3}{*}{45} &  Open text corpora, rules, \\ & & & & automatic UD annotation, \\
    & & & & pretraining data detection \\
    \bottomrule
    \end{tabular}
}
\caption{Comparison of benchmarks of linguistic minimal pairs for different languages: BLiMP \cite{warstadt-etal-2020-blimp-benchmark}, CLiMP \cite{xiang-etal-2021-climp}, JBLiMP \cite{someya-oseki-2023-jblimp}, SLING \cite{song-etal-2022-sling}, NoCoLA$_\text{zero}$ \cite{jentoft-samuel-2023-nocola}, DaLAJ \cite{volodina-etal-2021-dalaj}, LINDSEA \cite{leong2023bhasa}, CLAMS \cite{mueller-etal-2020-cross}, and RuBLiMP (ours).}
\label{tab:blimps}
\end{table}

%% file: parts/design.tex
\section{RuBLiMP}
\autoref{fig:data_generation} outlines our approach to generating minimal pairs for RuBLiMP, which includes the following stages: sentence extraction and annotation (\S\ref{subsec:corpora_annotation}), minimal pair generation (\S\ref{subsec:pair_generation}) and curation (\S\ref{subsec:curation}). Our framework allows the user to customize each component and provides the foundation to mitigate the limitations of static benchmarks \cite{bowman-dahl-2021-will} through continuous generation of minimal pairs for a domain of interest and decontaminating the data for specific Russian LMs.

\subsection{Corpora Annotation} 
\label{subsec:corpora_annotation}
\paragraph{Sentence Extraction} Three open text corpora are used as the source of grammatical sentences: Wikipedia\footnote{\href{https://dumps.wikimedia.org/ruwiki/latest/}{\texttt{dumps.wikimedia.org/ruwiki/latest}}}, Wikinews\footnote{\href{https://dumps.wikimedia.org/ruwikinews/latest/}{\texttt{dumps.wikimedia.org/ruwikinews/latest}}}, and Librusec, a collection of digitalized Russian books \cite{Panchenko:17:aist}. We extract articles from Wikipedia and Wikinews using WikiExtractor \cite{Wikiextractor2015} and literary texts from Librusec using \texttt{corus}\footnote{\href{https://github.com/natasha/corus/tree/master}{\texttt{github.com/natasha/corus}}}. Next, we segment the documents into sentences and tokenize the sentences with the help of \texttt{natasha}\footnote{\href{https://github.com/natasha/natasha}{\texttt{github.com/natasha/natasha}}}. We filter out the sentences based on the number of tokens (6-to-50) and shallow heuristics to avoid the sentence segmentation errors.

\paragraph{Sentence Annotation} Each extracted sentence is annotated in the Universal Dependencies scheme \cite{nivre-etal-2017-universal} with a multidomain morphosyntactic parser for Russian \cite{anastasyev2020exploring}.

\subsection{Minimal Pair Generation}
\label{subsec:pair_generation}
We search the dependency trees for specific lexical units and linguistic structures and edit them using expert-written perturbation rules to create a pool of minimal pairs for a target paradigm (\S \ref{sec:phenomena}). Our rules are written by three authors of this paper (native Russian computational linguistics) based on theoretical works on Russian morphology, syntax, and semantics. Each set of rules undergoes a peer-review stage by one of the authors. Below, we provide a general description of the minimal pair generation procedure, which involves four main edit operations: addition, replacement, swapping, and movement. These operations ensure the equal length of the grammatical and ungrammatical sentences. The implementation details and a complete list of the literature are documented in \autoref{app:generation}.

\paragraph{Morphology} Our morphological perturbations violate the principles of the affix order \cite{Greenberg_63,reynolds2013order} and properties of inflectional classes. We introduce derivational and inflectional errors using \texttt{pymorphy2}\footnote{A rule-based morphological analyzer, which allows for inflecting a word w.r.t. a given set of grammatical features and searching a word and its grammatical properties in the supported dictionaries.} \cite{pymorphy2}, morphological dictionaries \cite{bocharov2013crowdsourcing} available in \texttt{pymorphy2}, and word formation dictionaries \cite{Bolshakova2021}.

\paragraph{Syntax} Here, we corrupt adpositional and verbal government, negative concord rules, and agreement in number, gender, person, or case \cite{Testelets:2001}. We search for a word from curated lists or with specific morphosyntactic features in relevant syntactic constructions and move it to a different constituent or change its form using \texttt{pymorphy2}. We consider various types of the subject (a noun phrase, genitive, and clause) and additional contexts with attractors, which introduce contextual ambiguity in the ungrammatical sentence.

\paragraph{Semantics} Our semantic perturbations alter the verb's argument structure and introduce temporal and aspectual violations across the entire sentence. \cite{hopper1980transitivity,Paducheva:2010}. We search for a word or phrases with certain morphosyntactic features (e.g., a transitive verb) and semantic properties using a manually curated list of temporal markers and word co-occurrence and semantic dictionaries from the Russian National Corpus \cite{savchuk-ruscorpora-2024}.

\begin{figure}[t!]
    \centering
    \includegraphics[width=\columnwidth]{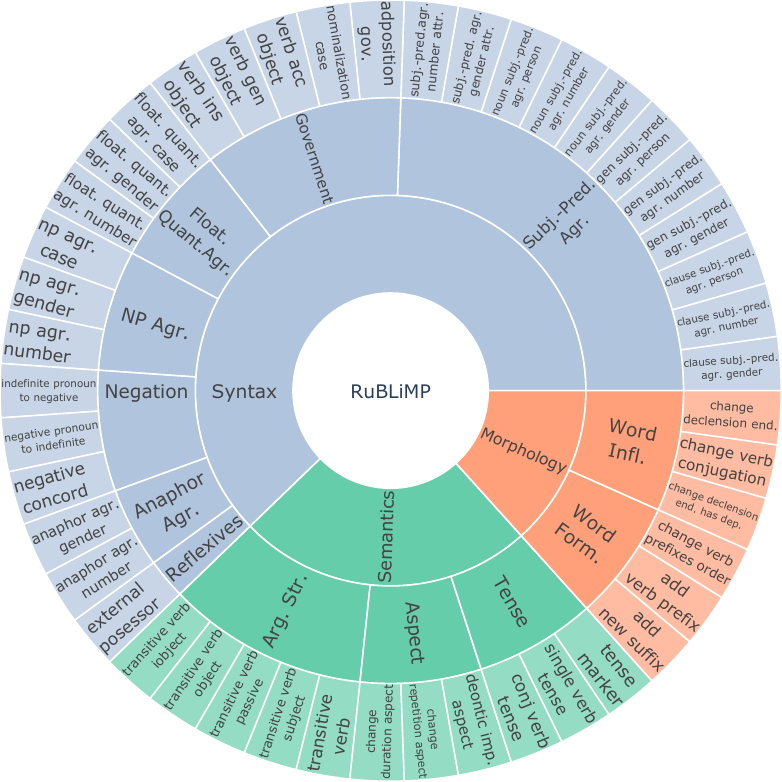}
    \caption{Distribution of phenomena in \rublimp.}
    \label{fig:phenomena}
\end{figure}

\subsection{Phenomena}
\label{sec:phenomena}
\rublimp includes \Nparadigms minimal pair types or \textit{paradigms}, each containing 1k minimal pairs. All paradigms are grouped into 12 \emph{phenomena} (see \autoref{fig:phenomena}), which are well represented in Russian theoretical and corpus linguistics. We provide a minimal pair example for each paradigm in \autoref{app:dataset_examples} and describe each phenomenon below.

\begin{itemize}[leftmargin=*,topsep=0pt]
    \setlength\itemsep{0pt}
    
     \item \textsc{Word Formation}: uninterpretable combinations of derivational affixes and violation of verb prefix stacking rules. 

    \item \textsc{Word Inflection}: incorrect use of declension affixes or verb conjugation endings.

    \item \textsc{Government}: incorrect use of a word governed by a nominalization, preposition, or verb.

    \item \textsc{Subject-Predicate Agreement}: violations of the subject-predicate agreement w.r.t. number, gender, person, or case. We include more complex agreement violation contexts with attractors. 

    \item \textsc{Anaphor Agreement}: incorrect agreement between an anaphoric relative pronoun and its antecedent in number or gender.
    
    \item \textsc{Noun Phrase Agreement}: agreement violation between the head of a noun phrase and its modifiers, such as adjectives and determiners, w.r.t. number, gender, or case.

    \item \textsc{Floating Quantifier Agreement}: lack of number, gender, or case agreement between a floating quantifier and a noun.

    \item \textsc{Reflexives}: incorrect use of a reflexive pronoun in constructions with an external possessor.
    
    \item \textsc{Negation}: negative particle movement and inappropriate use of negative and indefinite pronouns.

    \item \textsc{Argument Structure}: violations of the animacy requirement for a transitive verb's arguments via the replacement of a subject, direct or indirect object, and predicate in the active or passive voice.

    \item \textsc{Aspect}: incorrect use of perfective verbs in contexts with semantics of duration and repetition and in negative constructions with deontic verbs. 

    \item \textsc{Tense}: incorrect choice of (i) a single or conjoined verb form in a sentence with temporal adverbial (an adverb or a noun phrase) and (ii) a temporal adverbial in a sentence with a tense-marked verb.
    
\end{itemize} 

\subsection{Minimal Pair Curation} 
\label{subsec:curation}
Detecting pretraining data helps measure test data contamination and becomes a necessary component of the evaluation design \cite{brown2020language,eval-harness}. In our work, we employ a pretraining data detection method as a filtering stage when creating RuBLiMP. In particular, we use \textsc{Min-K\% Prob} \cite{shi2023detecting}, which relies on the hypothesis that a pretraining example is less likely to include outlier tokens with low probability compared to a non-pretraining example. The main idea is to compute the average log-likelihood of \textsc{K\%} tokens with minimum probability and determine a threshold \textit{t} used to classify an example as pretraining or non-pretraining.  \textsc{Min-K\% Prob} does not require an access to an LM's pretraining corpus and is highly efficient for scoring-based evaluation, since both \textsc{Min-K\% Prob} and a sentence's probability are computed via a single forward pass.

\noindent We compute \textsc{Min-K\% Prob} for each grammatical sentence in a pool of generated minimal pairs using 25 LMs described in \S\ref{sec:exps}. For each paradigm, we then run a grid search for \textit{t}, which allows to find an intersection of 1k minimal pairs between \emph{all} LMs. The minimal pairs in the intersection contain unique grammatical sentences, which are not detected as pretraining examples for any LM\footnote{We limit the maximum number of the generated minimal pairs for each paradigm to 350k. If the threshold search allows us to find more than 1k pairs in the LMs' intersection, we downsample the decontaminated pairs to 1k in a stratified fashion w.r.t. domain, length, and paradigm-specific features.}. We conduct ablation studies on choosing \textsc{K\%} in \S\ref{sec:ablation_mink}.

\subsection{Human Validation}
\label{subsec:human_validation}
\paragraph{Annotation Design} We conduct an in-house human validation to verify that the generated minimal pairs unambiguously isolate a target phenomenon and illustrate a grammaticality contrast. We create a team of 20 undergraduate BA and MA students in fundamental and computational linguistics from several Russian universities. We collaborate closely with the students over the course of the annotation project and maintain communication in a group chat. Our project includes a training phase and a main annotation phase. Each student is given detailed annotation guidelines available at any time during both annotation phases. We train the students to perform the task on 10 examples with explanations and ensure that their training performance is above 70\% \cite{nangia-bowman-2019-human}. The main annotation phase counts 2,350 examples (50 minimal pairs per paradigm). Each student receives a page with 5 minimal pairs, one of which is a honeypot example\footnote{Honeypot examples are a standard practice to estimate the annotation quality \cite{ustalov2024learning}. Three authors of this paper prepare 250 honeypot minimal pairs by manually labelling the generated pairs as ``positive'' and ``negative''. Various inconsistencies are manually introduced to balance the number of ``negative'' examples, such as violation of several phenomena, perturbing multiple sentence units, or usage of ambiguous word forms. An annotator labels a honeypot example without knowing the ground truth label, and then the annotator’s labels are compared against the authors’ labels in order to measure the annotator’s performance.}. The pay rate is on average \$20/hr, the minimum response time per page is 25 seconds, and the average honeypot performance exceeds 75\%. A shortened version of the guidelines and an example of the web interface are in Appendix \ref{sec: app_data_validation_guidelines}.

\input{tables/dawid_skene_by_paradigm} 

\paragraph{Vote Aggregation} The students' votes are aggregated with the Dawid-Skene  method~\cite{dawid_skene} using \texttt{Crowd-Kit} \cite{ustalov2024learning}. We compute the inter-annotator agreement using the Worker Agreement with Aggregate (WAWA) coefficient \cite{ning-etal-2018-joint}, which indicates the average fraction of the annotators’ votes that agree with the aggregated vote for each pair.

\paragraph{Results} We report the per-phenomenon results in \autoref{tab: ds_phenomenon} and per-paradigm results in \autoref{tab:ds_paradigm} (see Appendix \ref{sec: app_data_validation_results}). Overall, we observe a high ratio of plausible minimal pairs (94.35\%), with more than 85\% of correctly generated pairs for most of the paradigms. The average IAA as measured by WAWA is 92.5, indicating a strong agreement.

\subsection{General Statistics}
\label{subsec:statistics}
The \rublimp's general statistics are summarized in \autoref{tab:rublimp_sentences_stats} and compared with the Russian subset of \clams \cite{mueller-etal-2020-cross}, a pattern-generated benchmark for subject-predicate agreement. 

\paragraph{Length and Frequency} We compute the ratio of high-frequency tokens in the grammatical sentences as follows. We divide the number of tokens whose number of instances per million in our corpus (\S\ref{subsec:corpora_annotation}) is $\geq$ 1 by the sentence length in tokens. The sentences contain on average 11.3 tokens and 87.4\% of high-frequency tokens. In \clams, the sentences are shorter on average (7.55 tokens) and similar in terms of the high-frequency tokens ratio (86.3\%). We also observe that the overall number of unique tokens in \clams's 40.1k grammatical sentences is 126, which indicates its low lexical diversity. In contrast, \rublimp's subset for the syntactic phenomena counts 57.9k unique tokens.

\paragraph{Syntactic Diversity} We compute the dependency tree depth and the number of unique POS 5-grams and syntactic patterns at the benchmark- and sentence-level. The sentences vary in terms of the word order, with the number of unique POS 5-grams ranging between 18.9k (morphology) and 50k (syntax). The average tree depth in \rublimp is 4.18, and there are  24.6k unique syntactic patterns, with the average pattern frequency of 1.82 (see \autoref{app:dataset_stats}). Comparing RuBLiMP's minimal pairs for the syntactic phenomena to CLAMS, we find that CLAMS has significantly less variety, with 70 unique syntactic patterns, and their average frequency of 573.1. The number of unique POS 5-grams and average tree depth  are smaller: 99 and 2.94, respectively. This confirms that utilizing open text corpora promotes high linguistic diversity. We report the CLAMS's manual analysis results in \S\ref{sec:multilingual_exps}.

\input{tables/rublimp_clams_stats}

%% file: tables/dawid_skene_by_paradigm.tex
\begin{table}[t!]
\centering
\resizebox{0.85\columnwidth}{!}{
\begin{tabular}{lrr}
\toprule
\textbf{Paradigm}                   & \textbf{\%} &  \textbf{WAWA} \\
\midrule
\textsc{Word Formation}             &  95.77 &              92.83 \\
\textsc{Word Inflection}            &  95.33 &              93.90 \\
\textsc{Government}                 &  91.83 &              91.84 \\
\textsc{Subject-Predicate Agreement}     &  95.87 &              92.46 \\
\textsc{Anaphor Agreement}               &  94.06 &              93.00 \\
\textsc{Noun Phrase Agreement}           &  96.50 &              94.33 \\
\textsc{Floating Quantifier Agreement}   &  97.28 &              92.37 \\
\textsc{Reflexives}                 & 100.0 &              96.50 \\
\textsc{Negation}                   &  93.33 &              92.60 \\
\textsc{Argument Structure}         &  93.51 &              89.94 \\
\textsc{Aspect}                     &  95.28 &              92.97 \\
\textsc{Tense}                      &  93.79 &              92.10 \\ \hdashline
\textsc{Average} & 94.35 & 92.51 \\
\bottomrule
\end{tabular}}
\caption{The ratio of plausible minimal pairs (\%) by phenomenon and per-phenomenon WAWA inter-annotator agreement rates.}
\label{tab: ds_phenomenon}
\end{table}

%% file: tables/rublimp_clams_stats.tex
\begin{table}[t!]
\resizebox{\columnwidth}{!}{%
    \begin{tabular}{lrrrr@{\hskip 2em}r}
    \toprule
    & \multicolumn{4}{c}{\textbf{\rublimp}} & \multirow{2}{*}{\textbf{CLAMS}} \\
    & \textbf{Morphology} & \textbf{Semantics} & \textbf{Syntax} & \textbf{Overall} &   \\
    \midrule
    \multicolumn{6}{c}{Benchmark-level}\\
    \midrule
    \textbf{\# Pairs} & 6k & 11k & 28k & 45k & 40k \\
    \textbf{\# Patterns} & 3,9k & 7,4k & 15,9k & 24,6k & 70 \\
    \textbf{Pattern Frequency} & 1.52 & 1.48 & 1.76 & 1.82 & 573.1 \\
    \textbf{\# Unique Tokens} & 20.7k & 33.8k & 57.9k & 86.5k & 126 \\
    \textbf{\# POS 5-Grams} & 18,9k & 30,9k & 50k & 64,9k & 99 \\
    \midrule
    \multicolumn{6}{c}{Sentence-level}\\
    \midrule
    \textbf{Frequency (\%)} & 86.6 & 88.9 & 87.0 & 87.4 & 86.3 \\
    \textbf{Depth} & 4.02 & 4.41 & 4.12 & 4.18 & 2.94 \\
    \textbf{\# Tokens} & 10.46 & 12.23 & 11.14 & 11.31 & 7.55\\
    \textbf{\# POS 5-Grams} & 6.46 & 8.23 & 7.14 & 7.31 & 3.56 \\
    \bottomrule
    \end{tabular}}
\caption{Benchmark- and sentence-level general statistics in comparison with CLAMS.}
\label{tab:rublimp_sentences_stats}
\end{table}

%% file: parts/setup.tex
\section{Experimental Setup}
\label{sec:exps}

\input{tables/llms_mini_tab}

\vspace{-0.5em}
\paragraph{Language Models} \autoref{tab:lms} summarizes a broad range of 25 pretrained decoder- and encoder-only LMs used in our work and accessed via \texttt{Transformers} \cite{wolf-etal-2020-transformers}. Each LM is used in our \textsc{Min-K\% Prob} ablation studies (\S\ref{sec:ablation_mink}) and empirical evaluation experiments in monolingual (\S\ref{sec:monolingual}) and cross-lingual scenarios (\S\ref{sec:multilingual_exps}). 

\paragraph{Method} The sentences in a minimal pair are ranked based on their perplexity (PPL) or pseudo-perplexity (PPPL). The PPL of a sentence $s$ is inferred with a decoder-only LM as~\autoref{eqn:ppl}, where $|s|$ is the sentence length in tokens and $\Theta$ denotes the LM’s parameters.

\vspace{-13pt}
\begin{equation}
  PPL(s) = exp(-\frac{1}{|s|}\sum_{i=0}^{|s|} 
\log {P_{\Theta}}(x_i|x_{<i}))
\label{eqn:ppl}
\end{equation}

\noindent The PPPL \cite{salazar-etal-2020-masked} is computed with an encoder-only LM as in~\autoref{eqn:pppl}. Each token $x_j$ in $s$ is masked out and predicted based on the past and future tokens $x_{\backslash i} = (x_1,\ldots,x_{i-1},\ldots,x_{i+1},\ldots,x_{|s|})$.

\vspace{-13pt}
\begin{equation}
  PPPL(s) = exp(-\frac{1}{|s|}\sum_{i=0}^{|s|} 
\log {P_{\Theta}}(x_i|x_{\backslash i}))
\label{eqn:pppl}
\end{equation}


\paragraph{Human Baseline} We establish the human baseline on 5\% of \rublimp (2,350 pairs; 50 pairs per paradigm) using ABC\footnote{Available only in Russian: \href{https://elementary.center}{\texttt{elementary.center}}}, a crowdsourcing platform. Each of the 144 hired workers is certified as a native Russian speaker and paid \$15/hr on average. The annotation task is to select a grammatical sentence in a given pair (see \autoref{sec: app_human_baseline}). The sentences in a pair are randomly shuffled. We use 10 training and 100 honeypot examples and aggregate the votes using the Dawid-Skene method. The average response time per one pair is 10 seconds, and the average honeypot performance exceeds 90\%.

%% file: tables/llms_mini_tab.tex
\begin{table}[htp!]
    \centering
    \resizebox{\columnwidth}{!}{ %
    \begin{tabular}{llrc}
    \toprule
    \textbf{Model} &  \textbf{Source} & \textbf{Size} & \textbf{Corpus}  \\ 
    \midrule
    \multicolumn{4}{c}{Encoder-only LMs} \\
    \midrule
    \href{https://huggingface.co/ai-forever/ruBert-base}{\rubertb}  &   \multirow{2}{*}{\citet{zmitrovich-etal-2024-family-pretrained}}  & 178M & \multirow{2}{*}{Wikipedia, news}   \\
        \href{https://huggingface.co/ai-forever/ruBert-large}{\texttt{\rubertl}} &  & 427M &     \\

    \vspace{-1em}\\
    \hdashline
    \vspace{-1em}\\

    \href{https://huggingface.co/ai-forever/ruRoberta-large}{\ruroberta} & \citet{zmitrovich-etal-2024-family-pretrained} &  355M &  Wikipedia, news, books  \\

    \vspace{-1em}\\
    \midrule
    \vspace{-1em}\\

    \href{https://huggingface.co/distilbert/distilbert-base-multilingual-cased}{\dmbert} & \citet{Sanh2019DistilBERTAD} & 134M  & \multirow{2}{*}{Wikipedia} \\
    \href{https://huggingface.co/google-bert/bert-base-multilingual-cased}{\mbert} & \citet{devlin-etal-2019-bert} & 177M &      \\
    
    \vspace{-1em}\\
    \hdashline
    \vspace{-1em}\\
    
    \href{https://huggingface.co/FacebookAI/xlm-roberta-base}{\xlmrb}  &  \multirow{2}{*}{\citet{conneau-etal-2020-unsupervised}}  & 279M & \multirow{2}{*}{C4}
    \\
    \href{https://huggingface.co/FacebookAI/xlm-roberta-large}{\xlmrl} &  & 560M &  \\

    \vspace{-1em}\\
    \hdashline
    \vspace{-1em}\\
    \href{https://huggingface.co/google/rembert}{\rembert} & \citet{Chung2021Rethinking} & 575M &  Wikipedia  \\
    \href{https://huggingface.co/microsoft/mdeberta-v3-base}{\mdeberta} & \citet{he2021debertav3} & 276M  & C4    \\

    \midrule
    \multicolumn{4}{c}{Decoder-only LMs} \\
    \midrule

    \href{https://huggingface.co/ai-forever/rugpt3small_based_on_gpt2}{\rugpts} &   \multirow{3}{*}{\citet{zmitrovich-etal-2024-family-pretrained}} & 125M & \multirow{3}{*}{Wikipedia, C4, news, books}  \\
    \href{https://huggingface.co/ai-forever/rugpt3medium_based_on_gpt2}{\rugptm}  &  &  355M &    \\
    \href{https://huggingface.co/ai-forever/rugpt3large_based_on_gpt2}{\rugptl}&   &  760M &    \\

    \vspace{-1em}\\
    \hdashline
    \vspace{-1em}\\

    \href{https://huggingface.co/ai-forever/ruGPT-3.5-13B}{\rugpttpf}
     & N/A &  13B &  Wikipedia, news, books, other  \\

    \vspace{-1em}\\
    \hdashline
    \vspace{-1em}\\

    \href{https://huggingface.co/sambanovasystems/SambaLingo-Russian-Base}{\smblingorb} & \citet{csaki2023efficiently} &  7B & CulturaX 
    \\
    
    \vspace{-1em}\\
    \midrule
    \vspace{-1em}\\

    \href{https://huggingface.co/ai-forever/mGPT}{\mgpt} & \multirow{2}{*}{\citet{shliazhko2024mgpt}} & 1.3B  & \multirow{2}{*}{Wikipedia, C4}  \\

    \href{https://huggingface.co/ai-forever/mGPT-13B}{\mgptb} &  & 13B  &   \\
    
    \vspace{-1em}\\
    \hdashline
    \vspace{-1em}\\

    \href{https://huggingface.co/bigscience/bloom-1b7}{\blooms}
     & \multirow{3}{*}{\citet{workshop2023bloom}} & 1.7B & \multirow{3}{*}{ROOTS}
     \\
    \href{https://huggingface.co/bigscience/bloom-3b}{\bloomm} &  & 3B  &  \\
    \href{https://huggingface.co/bigscience/bloom-7b1}{\blooml} &  & 7.1B &  \\
    
    \vspace{-1em}\\
    \hdashline
    \vspace{-1em}\\

    \href{https://huggingface.co/facebook/xglm-1.7B}{\xglms}
     &  \multirow{3}{*}{\citet{lin-etal-2022-shot}} & 1.7B & \multirow{3}{*}{C4}  \\
    \href{https://huggingface.co/facebook/xglm-4.5B}{\xglmm} & & 4.5B &    \\
    \href{https://huggingface.co/facebook/xglm-7.5B}{\xglml} &  & 7.5B &    \\

    \vspace{-1em}\\
    \hdashline
    \vspace{-1em}\\

    \href{https://huggingface.co/meta-llama/Llama-2-7b}{\llamas}
     & \multirow{2}{*}{\citet{touvron2023llama}} & 7B  & \multirow{2}{*}{Web corpora}  \\
        
   \href{https://huggingface.co/meta-llama/Llama-2-13b}{\llamam} & \multirow{2}{*}{}  & 13B  \\

    \vspace{-1em}\\
    \hdashline
    \vspace{-1em}\\

    \href{https://huggingface.co/mistralai/Mistral-7B-v0.1}{\mistral} & \citet{jiang2023mistral} & 7B & Web corpora  \\

    \bottomrule
    \end{tabular}%
    } 
     \caption{The LMs used in our work. Corpora references: C4 \cite{raffel2020exploring}, CulturaX \cite{nguyen-etal-2024-culturax}, and ROOTS \cite{laurenccon2022bigscience}.}
    \label{tab:lms}
\end{table}

%% file: parts/min_kablations.tex
\section{\textsc{Min-K\%}: Ablation Studies}
\label{sec:ablation_mink}
We begin with ablation studies on the effect of the minimal pair curation stage and the hyperparameter \textsc{K\%} $\in \{30, 40, 50, 60\}$. For each paradigm in \rublimp, (i) we randomly sample 1k generated minimal pairs and evaluate the LMs to get the reference scores (the accuracy scores are averaged over 100 runs), and (ii) decontaminate the generated minimal pairs through a greed search for $t$ and select 1k pairs with the maximum \textsc{Min-K\% Prob} as described in \S\ref{subsec:curation} and evaluate the LMs' performance. We then compute the $\Delta$-score between (i) and (ii) for each LM, which measures the performance drop when using \textsc{Min-K\% Prob} with certain \textsc{K\%}.

\paragraph{Higher \textsc{K\%} is More Effective} 
\autoref{fig:ablation_k} shows that \textsc{Min-K\% Prob} ensures adversarial filtering of the pool of generated minimal pairs. In general, the higher \textsc{K\%} value, the lower the $\Delta$-score for most LMs. We find that the overall performance can drop from 2.9\%  to 12\% and the $\Delta$-score can depend on the model size (e.g., \texttt{ruGPT}, \texttt{bloom}, and \texttt{Llama-2}). However, the $\Delta$-scores for \rembert and \mdeberta are positive; we relate it to the fact that these LMs perform close to random guessing on \rublimp (\S \ref{sec:monolingual}) and other related benchmarks (\S \ref{sec:multilingual_exps}). We select \textsc{K\%} of $60$ to create \rublimp.

%% file: parts/monolingual_experiments.tex
\section{Results on RuBLiMP}
\label{sec:monolingual}
\input{tables/rublimp_phenomena_scores}
This section describes the empirical evaluation results on \rublimp. We report the results by phenomenon in \autoref{tab:phenomena_scores_rublimp} and by paradigm in \autoref{app:add_res}. Overall, we find that the best performing and the largest monolingual LM (\rugpttpf) still falls short compared to humans, whose performance exceeds $95\%$ on all \rublimp's paradigms. Analyzing the results for the monolingual and multilingual LMs, we observe that the former generally perform better, and the latter can achieve the random baseline performance (e.g., \rembert, \mdeberta, \xglms). We evaluate the multilingual LMs on five related BLiMP-style benchmarks to explore this behavior in more detail (\S\ref{sec:multilingual_exps}). Below, we discuss our findings from the perspective of the LM size, phenomenon, domain, and length.

\paragraph{Larger $\neq$ Better} We find that smaller LMs can outperform or perform on par with larger LMs. In particular, \rugptm performs close to \rugptl on average, while \rubertb \& \rubertl, \xglmm \& \xglml, and \mgpt \& \mgptb perform on par on certain phenomena (e.g.,  \textsc{Word Formation}, \textsc{Anaphor Agreement}, and \textsc{Tense}). This finding aligns with \citet{warstadt-etal-2020-blimp-benchmark,song-etal-2022-sling}.

\paragraph{Higher Sensitivity to Local Edits} The LMs are robust to local perturbations for \textsc{Word Inflection} and \textsc{Word Formation}. We observe that the LMs can perform on par with humans in identifying an incorrect order of the verb prefixes. The presence of a modifier helps the LMs resolve an incorrect word's declension, improving the accuracy by up to 5\% (e.g., \texttt{ruBERT}, \texttt{ruGPT}, and \texttt{mGPT}).

\paragraph{Lower Sensitivity to Structural Relations} The LMs achieve lower performance on the structural phenomena  \cite{reinhart2016anaphora}. The behavior is more pronounced for the multilingual LMs, which fall behind humans by up to 40\% on \textsc{Anaphor Agreement} and 45\% on \textsc{Reflexives}. 

\begin{figure*}[t!]
    \centering
    \subfloat[Wikipedia]{\includegraphics[width=0.32\textwidth]{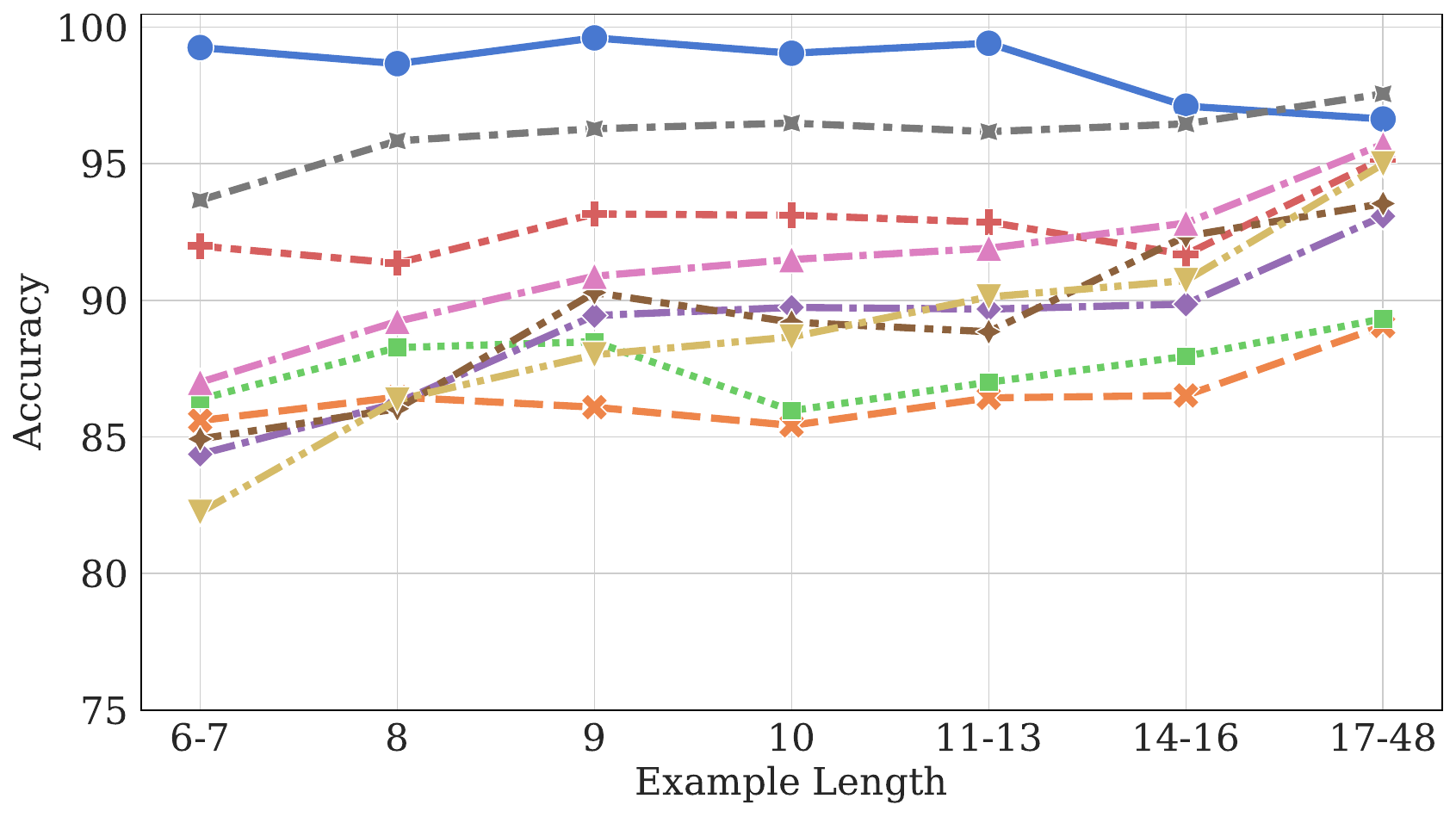}}
    \subfloat[Wikinews]{\includegraphics[width=0.32\textwidth]{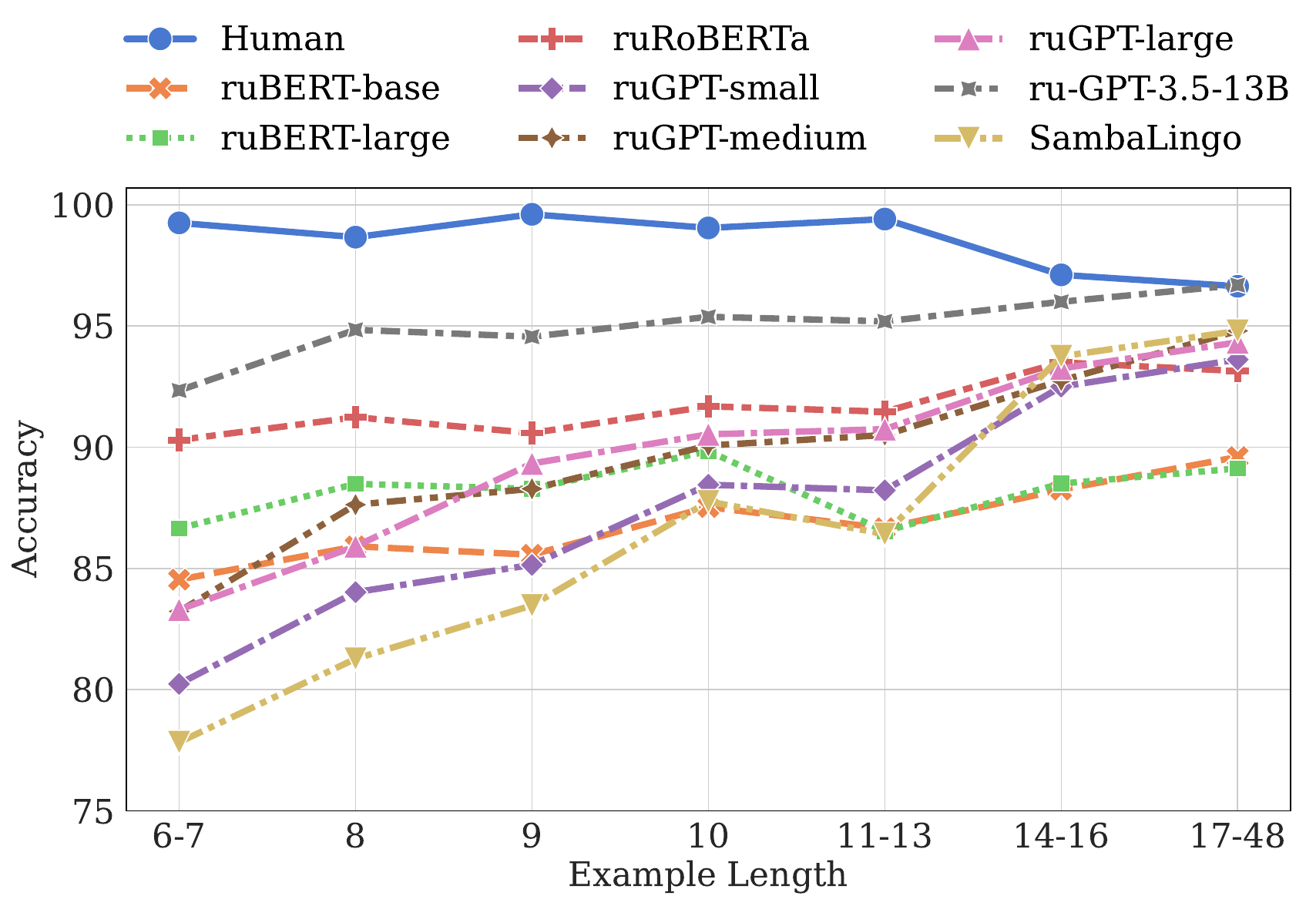}}
    \subfloat[Librusec]{\includegraphics[width=0.32\textwidth]{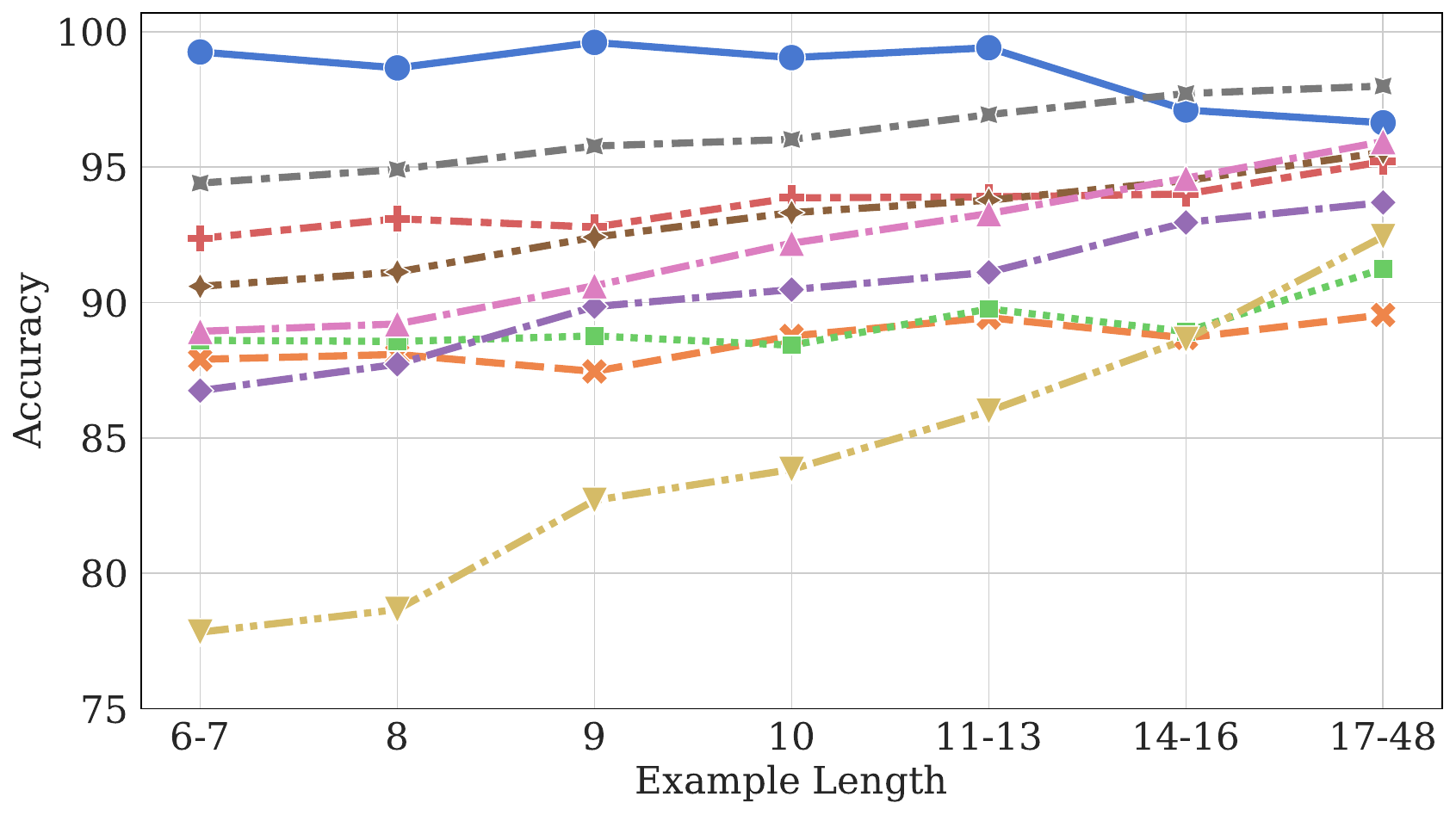}}
    \caption{Results on \rublimp for the monolingual LMs per domain grouped by seven quintiles of the length.}
    \label{fig:length}
\end{figure*}

\paragraph{LMs Struggle with Negative Pronouns} \textsc{Negation} is one of the challenging phenomena in \rublimp. In particular, most LMs are least sensitive to the replacement of a negative pronoun with an indefinite one (see \autoref{app:add_res}), which requires understanding of the pronoun licensing conditions (e.g., \textit{On nikogda/$^{*}$kogda-nibud' ne hodit v teatr} ``He never/$^{*}$ever goes to the theatre''). However, the LMs distinguish well between the sentences without a negative particle \textit{ne} ``not'' where an indefinite pronoun is replaced with a negative one (e.g., \textit{Petr kogda-to/$^{*}$nikogda byl v Moskve} ``Petter was once/$^{*}$never in Moscow'').

\paragraph{Attractors Confuse LMs} Analyzing the effect of the attractor presence (see Appendix \ref{app:syntax_docs} for details), we observe that the LMs' performance can drop by up to 10\% on \textsc{Subject-predicate Agreement} if an attractor is added (see \autoref{app:add_res}; e.g., \texttt{bloom}, \texttt{SambaLingo}, and \texttt{mGPT}).

\paragraph{LMs are Less Sensitive to Tense} Another finding is that the LMs struggle to identify a violated tense form of a single verb, with the accuracy ranging between the random guessing (\xglms) to 90.7\% (\rugpttpf). However, having a conjoined verb increases the performance by up to 17.3\% (\mgpt), which indicates that the LMs utilize the context reliably. 

\paragraph{Effect of Length \& Domain} We estimate the effect of length per domain by dividing \rublimp into 7 length groups of equal size. The results for the monolingual and multilingual LMs are in \autoref{fig:length} and Figures \ref{fig:length_multi_enc}-\ref{fig:length_multi_dec} (\autoref{app:add_res}), respectively. While human performance is consistent, the LMs' performance improves as the length increases. The first length groups (6-10 tokens) contain pairs related to the most challenging phenomena for the LMs (syntax: \textsc{Negation},  \textsc{Reflexives}; semantics: \textsc{Argument Structure},  \textsc{Aspect}). We find that some LMs are more domain-sensitive (e.g., \texttt{SambaLingo}, \texttt{ruGPT}), while others receive similar scores (e.g., \texttt{ruGPT-3.5}, \texttt{XLMR}).

%% file: tables/rublimp_phenomena_scores.tex
\begin{table*}[ht!]
\setlength{\tabcolsep}{1.5pt}
\resizebox{\textwidth}{!}{
    \begin{tabular}{lRRRRRRRRRRRRR}
    
  \textbf{Model} & \multicolumn{1}{p{2.5ex}}{\rotatebox{\rotation}{\textsc{Word Form.}}} & \multicolumn{1}{p{2.5ex}}{\rotatebox{\rotation}{\textsc{Word Infl.}}} & \multicolumn{1}{p{2.5ex}}{\rotatebox{\rotation}{\textsc{Government}}} & \multicolumn{1}{p{2.5ex}}{\rotatebox{\rotation}{\textsc{Subj.-Pred. Agr.}}} & \multicolumn{1}{p{2.5ex}}{\rotatebox{\rotation}{\textsc{Anaphor Agr.}}} & \multicolumn{1}{p{2.5ex}}{\rotatebox{\rotation}{\textsc{NP Agr.}}} & \multicolumn{1}{p{2.5ex}}{\rotatebox{\rotation}{\textsc{Float. Quant. Agr.}}} & \multicolumn{1}{p{2.5ex}}{\rotatebox{\rotation}{\textsc{Reflexives}}} & \multicolumn{1}{p{2.5ex}}{\rotatebox{\rotation}{\textsc{Negation}}} & \multicolumn{1}{p{2.5ex}}{\rotatebox{\rotation}{\textsc{Arg. Structure}}} & \multicolumn{1}{p{2.5ex}}{\rotatebox{\rotation}{\textsc{Aspect}}} & \multicolumn{1}{p{2.5ex}}{\rotatebox{\rotation}{\textsc{Tense}}}  & \multicolumn{1}{c}{\rotatebox{\rotation}{\textsc{Average}}} \\\\
\midrule
\rubertb & 81.90 & 84.87 & 90.72 & 91.16 & 85.90 & 83.57 & 91.40 & 78.70 & 77.77 & 88.52 & 96.07 & 87.17 & 86.48 \\
\rubertl & 82.83 & 86.03 & 90.66 & 91.43 & 86.35 & 84.70 & 91.23 & 81.00 & 82.13 & 89.20 & 96.47 & 88.17 & 87.52 \\
\ruroberta & 89.67 & 91.63 & 96.68 & 93.62 & 95.60 & 88.83 & 96.17 & 91.10 & 89.83 & 91.64 & 97.20 & 92.40 & 92.86 \\
\rugpts & 88.53 & 91.57 & 92.94 & 90.33 & 94.30 & 95.33 & 87.63 & 83.20 & 73.17 & 88.82 & 93.93 & 84.30 & 88.67 \\
\rugptm & 91.77 & 86.37 & 94.88 & 91.57 & 95.90 & 97.37 & 96.17 & 79.90 & 80.53 & 91.98 & 95.60 & 88.70 & 90.89 \\
\rugptl & 89.23 & 91.37 & 94.58 & 91.51 & 95.80 & 96.77 & 90.83 & 87.80 & 78.60 & 92.24 & 95.53 & 87.03 & 90.94 \\
\rugpttpf & 94.33 & 95.20 & 97.10 & 96.12 & 97.05 & 98.47 & 98.17 & 94.70 & 87.53 & 96.34 & 97.77 & 95.37 & 95.68 \\
\smblingorb & 79.87 & 85.73 & 89.20 & 80.85 & 92.95 & 89.83 & 90.43 & 96.20 & 77.63 & 82.74 & 87.40 & 80.47 & 86.11 \\
\midrule

\dmbert & 83.97 & 79.63 & 70.84 & 75.90 & 52.35 & 79.43 & 83.13 & 56.00 & 75.27 & 55.88 & 59.47 & 55.83 & 68.98 \\
\mbert & 88.83 & 84.63 & 78.88 & 80.37 & 86.35 & 87.07 & 82.77 & 52.90 & 66.30 & 61.22 & 59.77 & 52.70 & 73.48 \\
\xlmrb & 88.57 & 90.57 & 88.42 & 87.55 & 91.85 & 92.67 & 92.97 & 69.90 & 72.50 & 75.48 & 81.57 & 74.67 & 83.89 \\
\xlmrl & 88.80 & 91.03 & 90.52 & 87.73 & 93.15 & 94.37 & 93.07 & 79.10 & 80.67 & 81.30 & 87.70 & 79.77 & 87.27 \\
\rembert & 51.40 & 54.70 & 48.90 & 49.77 & 32.05 & 51.17 & 62.63 & 45.30 & 51.17 & 49.28 & 52.40 & 52.20 & 50.08 \\
\mdeberta & 52.57 & 43.63 & 47.50 & 36.77 & 75.35 & 41.03 & 37.43 & 40.20 & 43.57 & 41.90 & 44.10 & 53.53 & 46.47 \\
\mgpt & 94.37 & 95.97 & 89.64 & 87.69 & 92.15 & 94.20 & 85.13 & 82.50 & 67.80 & 79.94 & 85.43 & 79.53 & 86.20 \\
\mgptb & 94.53 & 95.53 & 92.08 & 88.75 & 94.35 & 95.33 & 88.50 & 85.60 & 68.27 & 84.46 & 87.70 & 83.03 & 88.18 \\
\blooms & 86.10 & 89.70 & 67.86 & 85.55 & 69.75 & 79.10 & 66.87 & 13.10 & 65.20 & 53.78 & 54.67 & 68.87 & 66.71 \\
\bloomm & 89.53 & 90.23 & 71.44 & 86.66 & 65.85 & 81.10 & 68.33 & 19.30 & 67.83 & 54.22 & 51.13 & 72.30 & 68.16 \\
\blooml & 88.90 & 91.87 & 73.62 & 88.91 & 73.10 & 84.63 & 75.37 & 23.30 & 68.40 & 57.64 & 55.07 & 77.43 & 71.52 \\
\xglms & 37.70 & 45.03 & 51.72 & 44.26 & 65.70 & 61.57 & 47.23 & 64.10 & 54.63 & 38.20 & 75.27 & 51.93 & 53.11 \\
\xglmm & 92.40 & 92.17 & 87.96 & 82.70 & 92.80 & 92.70 & 91.30 & 82.90 & 73.97 & 82.62 & 90.47 & 80.57 & 86.88 \\
\xglml & 92.80 & 93.43 & 88.46 & 83.75 & 93.45 & 93.70 & 91.03 & 90.80 & 74.43 & 83.12 & 90.27 & 82.37 & 88.13 \\
\llamas & 94.70 & 90.83 & 85.20 & 89.45 & 48.35 & 89.23 & 72.10 & 84.80 & 72.40 & 79.96 & 81.20 & 81.93 & 80.85 \\
\llamam & 95.83 & 93.50 & 88.50 & 91.23 & 56.00 & 91.53 & 76.97 & 89.00 & 74.00 & 83.08 & 85.50 & 85.63 & 84.23 \\
\mistral & 96.87 & 95.00 & 88.16 & 92.99 & 72.10 & 93.20 & 87.83 & 32.40 & 72.40 & 83.28 & 86.60 & 88.13 & 82.41 \\
\midrule
\textbf{Human} & 100.0 & 99.33 & 98.80 & 98.53 & 98.0 & 98.67 & 99.33 & 98.00 & 100.0 & 100.0 & 100.0 & 99.33 & 98.62\\
\bottomrule
\end{tabular}}
\caption{The average accuracy scores (\%) of the 25 LMs and human baseline by phenomenon. Random baseline is 50\%. The monolingual and multilingual LMs are separated by a line.}
\label{tab:phenomena_scores_rublimp}
\end{table*}

%% file: parts/multilingual_experiments.tex
\section{Multilingual Analysis}
\label{sec:multilingual_exps}
To analyze the multilingual LMs in more detail, we evaluate their sensitivity to linguistic phenomena in six benchmarks: BLiMP, CLiMP, SLING, JBLiMP, CLAMS, and RuBLiMP (see \autoref{tab:blimps} for statistics). We detail the experimental setup and empirical evaluation results in \autoref{appendix:multi} and outline our key findings here. (i) no single LM performs consistently well in all languages, (ii) the LMs' performance for \textsc{Agreement} in a given language depends on the benchmark, and the $\Delta$-scores between the benchmarks can be up to 15\% for English, 20\% for Chinese, and 35\% for Russian, and (iii) the manual analysis of  CLAMS reveals its concerning quality: 20\% of Russian minimal pairs are semantically implausible, 15\% do not isolate a phenomenon, and 5\% contain repetitive text (a native Russian speaker will unlikely say or write this way). Besides, there are only 126 unique tokens in the 40.1k grammatical sentences, which limits the sentence diversity. These findings raise the need for a more detailed comparison of LMs on peer-reviewed evaluation resources and their additional validation, which aligns with \citet{song-etal-2022-sling}.

%% file: parts/related_work.tex
\section{Related Work}
\paragraph{Evaluating Russian LMs' Grammatical Knowledge} Earlier studies introduce mono- and multilngual probing suites to explore how the LMs' representations encode Russian grammatical phenomena, 
ranging from a word's part of speech to gapping \citep[e.g.,][]{ravishankar-etal-2019-probing,csahin2020linspector,mikhailov-etal-2021-rusenteval,choenni-shutova-2022-investigating,serikov-etal-2022-universal}. RuCoLA \cite{mikhailov-etal-2022-rucola} includes expert-written and machine-generated (un-)acceptable sentences and aims to test the LM's linguistic competence via supervised acceptability classification. Our work extends the direction of evaluating Russian LM's grammatical knowledge and focuses on unsupervised acceptability judgments over linguistic minimal pairs.

\paragraph{Benchmarks of Linguistic Minimal Pairs}
The idea of discriminating between linguistic minimal pairs has gained visibility in NLP due to its several advantages, such as controlling the sentences' length and lexical units and providing a local view of an LM's decision boundary \cite{lau2017grammaticality,warstadt2022artificial}. With the creation of \blimp \cite{warstadt-etal-2020-blimp-benchmark}, similar resources have been proposed to evaluate LMs' acquisition of grammatical phenomena in languages other than English (see \autoref{tab:blimps}). In contrast to these benchmarks, \rublimp cover diverse phenomena in Russian morphology, syntax, and semantics beyond subject-verb agreement in CLAMS and includes pairs generated from naturally occurring and decontaminated sentences across three domains.

%% file: parts/conclusion.tex
\section{Conclusion and Future Work}
This work introduces \rublimp, the first large-scale multidomain benchmark of 45k minimal pairs for the Russian language. \rublimp covers \Nparadigms minimal pair types grouped into \Nphenomena linguistic phenomena in morphology, syntax, and semantics. The \rublimp creation approach ensures the linguistic diversity and high quality of the minimal pairs and minimizes the data contamination risk. We conduct an extensive empirical evaluation of 25 widely used monolingual and multilingual LMs for Russian and analyze their performance w.r.t. various criteria. Our results show that the LMs are better at identifying morphological and agreement-oriented contrasts than violations of structural relations, negation, transitivity, and tense. Furthermore, we analyse the 17 multilingual LMs in seven languages and find that no single LM performs well in all languages. Our \emph{future work} includes (i) comparison of pretraining data detection methods (ii) implementation of new phenomena (e.g., islands), and (iii) a more detailed multilngual study of the LMs' linguistic abilities. By releasing \rublimp, we hope to foster further research on how the Russian language is acquired by LMs.

%% file: parts/limitations.tex
\section*{Limitations}
This section describes the limitations of our work associated with our multi-stage minimal pair generation approach and computational costs. Noise in the publicly available data and automatic data extraction and annotation errors can generate implausible pairs. However, each stage is highly customizable based on the user needs, and expert validation of our approach shows that roughly 2,235 out of 2,350 generated minimal pairs unambiguously isolate a target phenomenon and display the required grammaticality contrast (\S\ref{subsec:human_validation}).

\paragraph{Corpus Annotation} In our work, we utilize data sources that undergo human review and editing (e.g., Wikipedia and Wikinews articles). However, there is still a high chance of noise in the data, such as web page artifacts or errors of optical character recognition systems. Another disadvantage of this stage is errors in the text segmentation tools and morphosyntactic parsers. We use the current state-of-the-art Russian NLP libraries and models and create a set of shallow heuristics to filter out irrelevant sentences through a series of manual data analysis iterations.

\paragraph{Minimal Pair Generation} On the one hand, our multidomain corpus represents a large-scale source of sentences with a high degree of diversity in terms of lexis, length, frequency, and linguistic structures. On the other hand, there are a few challenges due to the rich Russian morphology, a high degree of ambiguity, and a flexible word order. In particular, not all grammatical sentences with relevant linguistic constructions can be perturbed into ungrammatical ones, e.g., many word perturbations still result in plausible sentences and require additional heuristics to prevent semantic and syntactic felicity, which is not always possible. This is the main reason for narrowing down a set of linguistic structures and contexts to ensure control over the perturbations. We limit the number of the (i) phenomena criteria (e.g., considering nominalizations only with a specific set of endings), (ii) perturbation options (e.g., discarding ambiguous case forms during the government violations), or both (i) and (ii) (e.g., selecting verbs with only two prefixes during the word formation violations and only changing their order instead of adding more prefixes). Last but not least, the search for relevant lexical units and linguistic structures depends on the domain, which limits the scope of the domain-specific performance analysis (e.g., the temporal markers describing the duration or repetition of an event are primarily found in the news domain).

\paragraph{Minimal Pair Curation} Recent research has proposed a broad range of pretraining data detection methods. Our work does not aim to compare different solutions to this problem; we recognize that more advanced methods can be applied \citep[e.g., \textsc{Min-K\%++;}][]{zhang2024min}. We also acknowledge that the \textsc{Min-K\% Prob} method may still identify sentences that \emph{do appear} in the LMs' pretraining corpora as non-pretraining examples and select sentences with rare vocabulary items, which may lead to the performance decrease. Naturally, the effectiveness of the curation stage and the resulting LM's performance depends on the quality of the pretraining data detection method, which is an open question in the LM evaluation \& benchmarking research direction \cite{oren2023proving}. However, our approach allows one to continuously update \rublimp and create multiple versions of the benchmark, which can be decontaminated w.r.t. a set of LMs and another test data decontamination methods (or their ensemble).

\paragraph{Domain Shifts} Many studies report that LMs can judge frequent linguistic patterns in their pretraining corpora as grammatical and perform worse on rare sentences with low probabilities~\cite{marvin-linzen-2018-targeted,linzen2021syntactic}. Our benchmark design implies potential word frequency and domain distribution shifts between an LM's pretraining corpus and \rublimp, which can introduce bias in the evaluation. Nevertheless, we demonstrate a high diversity of syntactic patterns and a moderate word frequency in \rublimp's sentences (\S\ref{subsec:statistics}), and show that the LMs can generalize well to out-of-domain examples (\S\ref{sec:monolingual}).

\paragraph{Computational Costs} Each stage in our minimal pair generation approach requires efficient computational resources. However, the morphosyntactic parser in \S\ref{subsec:corpora_annotation} can be replaced with a more lightweight one with possible changes in the annotation quality (e.g., \texttt{slovnet}\footnote{\href{https://github.com/natasha/slovnet}{\texttt{github.com/natasha/slovnet}}}). Note that the minimal pair curation stage costs are reduced as follows. First, we filter out pairs based on \textsc{Min-K\% Prob} for decoder-only LMs due to their optimal inference speed. Next, we filter out the remaining pairs based on \textsc{Min-K\% Prob} for encoder-only LMs. Recall that both \textsc{Min-K\% Prob} and a sentence's probability are computed via a single forward pass.

%% file: parts/ethics_statement.tex
\section*{Ethics Statement}
\paragraph{Human Annotation} The annotators' votes 
in our annotation projects (see \S\ref{subsec:human_validation}; \S\ref{sec:exps}) are collected anonymously. The average pay rate significantly exceeds the hourly minimum wage in Russia. The annotators are warned about potentially sensitive topics in the examples, such as politics, culture, and religion.

\paragraph{Inference Costs} Evaluating an LM on \rublimp depends on the LM architecture and size and can be optimized with distributed inference libraries (e.g., \texttt{accelerate}\footnote{\href{https://github.com/huggingface/accelerate}{\texttt{github.com/huggingface/accelerate}}}). Running the complete evaluation experiment on a single V100 GPU takes approx. 1.5h and 11h for a decoder-only and encoder-only LM, respectively.

\paragraph{Potential Misuse} \rublimp can be used as training data for acceptability classifiers, potentially enhancing the quality of generated texts \cite{batra-etal-2021-building}. We acknowledge that these improvements in text generation might lead to the misuse of LMs for harmful purposes \cite{lucas-etal-2023-fighting}. \rublimp's intended use is for \textbf{research and development purposes}, and the potential negative uses are not lost on us. 

\paragraph{Transparency} We release \rublimp, our minimal pair generation framework, and all annotation materials under the permissive license following the standard open research practices. Our GitHub repository and HuggingFace dataset card \cite{lhoest-etal-2021-datasets} provide detailed documentation on the codebase, benchmark creation methodology, and human annotation.

\paragraph{Use of AI-assistants} We improve and proofread the text of this paper using Grammarly\footnote{\href{https://app.grammarly.com}{\texttt{grammarly.com}}} to correct grammatical, spelling, and style errors and paraphrasing sentences. Therefore, specific segments of our publication can be detected as AI-generated, AI-edited, or human-AI-generated.

%% file: parts/appendix_examples.tex
\onecolumn
\section{Examples of Minimal Pairs}
\label{app:dataset_examples}

\input{tables/data_examples}

%% file: tables/data_examples.tex
\renewcommand{\arraystretch}{1}
\newcommand\phenomenon{0.07}
\newcommand\example{0.4}
\newcommand\PID{0.25}
\newcommand\prefixes{0.01}
\newcommand\dash{\cdashline{2-4}[0.5pt/2pt]}

\begin{table*}[ht!]\tiny

\begin{adjustbox}{width=\textwidth,center}
    \begin{tabular}{p{\phenomenon\textwidth}p{\PID\textwidth}p{\example\textwidth}p{\example\textwidth}}
    \toprule
    Phenomenon 	&	 PID 	&	 Acceptable Example 	&	 Unacceptable Example 	 	\\	
    \midrule 																						
    \multirow{3}{\phenomenon\linewidth}{\textsc{Word \mbox{Formation}}} 	&	 add\allowbreak \_new\allowbreak \_suffix 	&	Priekhala \textbf{staren'kaya}, malen'kaya, khuden'kaya zhenshchina.  	&    Priekhala \textbf{starsken'kaya}, malen'kaya, khuden'kaya zhenshchina.    \\\dash
    {} 	&	 add\allowbreak \_verb\allowbreak \_prefix 	&  Vesnoy lichinki \textbf{vyedayut} pochki iznutri i v mae okuklivayutsya.   &    Vesnoy lichinki \textbf{vyv"edayut} pochki iznutri i v mae okuklivayutsya.    \\\dash
    {} 	&	 change\allowbreak \_verb\allowbreak \_prefixes\allowbreak \_order 	&	Khochu dolozhit', chto plany po etoy rabote my \textbf{perevypolnili}.  	&  Khochu dolozhit', chto plany po etoy rabote my \textbf{vyperepolnili}.      \\
    \midrule 

    \multirow{3}{\phenomenon\linewidth}{\textsc{Word \mbox{Inflection}}} 	&	 change\allowbreak \_declension\allowbreak \_ending 	&	 Fil'm byl dostatochno podrobno rassmotren v \textbf{zhurnale} "Iskusstvo kino". 	&	 Fil'm byl dostatochno podrobno rassmotren v \textbf{zhurnali} "Iskusstvo kino". \\\dash
    
    {} 	&	 change\allowbreak \_declension\allowbreak \_ending\allowbreak \_has\allowbreak \_dep 	&	 Znachitel'nye ploshchadi pashni podverzheny vodnoy \textbf{erozii}. 	&	 Znachitel'nye ploshchadi pashni podverzheny vodnoy \textbf{eroziu}. 		 	\\\dash
    {} 	&	 change\allowbreak \_verb\allowbreak \_conjugation 	& I nikomu uzhe ne \textbf{dokazhesh'}, chto ty -- eto ty, -- tak on dumal.  & I nikomu uzhe ne \textbf{dokazhish'}, chto ty -- eto ty, -- tak on dumal.  \\
    \midrule 	
    
    \multirow{5}{\phenomenon\linewidth}{\textsc{Government}} 	&	 adposition\allowbreak \_government 	&	Vpervye kosmonavt spal v \textbf{nevesomosti}.  	&	 Vpervye kosmonavt spal v \textbf{nevesomost'yu}.	\\	\dash 
    {} 	&	 verb\allowbreak \_acc\allowbreak \_object 	&	My opishem nashu \textbf{aksiomatizatsiyu} putem opisaniya struktury formul.  	&    My opishem nashu \textbf{aksiomatizatsiya} putem opisaniya struktury formul.    \\	\dash 
    {} 	&	 verb\allowbreak \_gen\allowbreak \_object 	&	Summy ne ochen' bol'shie, no \textbf{azarta} oni dobavlyayut.  	&     Summy ne ochen' bol'shie, no \textbf{azartom} oni dobavlyayut.   \\\dash
    {} 	&	 verb\allowbreak \_ins\allowbreak \_object 	&	Dioksid kremniya obladaet \textbf{polimorfizmom}.  	&    Dioksid kremniya obladaet \textbf{polimorfizma}.    \\	\dash 
    {} 	&	 nominalization\allowbreak \_case 	&	Pakhotnye ploshchadi podverzheny smyvu i vyduvaniyu \textbf{vetrom}.  	&     Pakhotnye ploshchadi podverzheny smyvu i vyduvaniyu \textbf{veter}.    \\
    \midrule
    
    \multirow{11}{\phenomenon\linewidth}{\textsc{Subject-predicate \mbox{Agreement}}} 	&	 noun\allowbreak \_subj\allowbreak \_predicate\allowbreak \_agreement\allowbreak \_number 	&	Pua-Katiki — potukhshij shchitovidnyj \textbf{vulkan} na ostrove Paskhi.  	&	Pua-Katiki — potukhshij shchitovidnyj \textbf{vulkany} na ostrove Paskhi.   \\ \dash
    {} 	&	 genitive\allowbreak \_subj\allowbreak \_predicate\allowbreak \_agreement\allowbreak \_number	&	 Predposylok dlya muzykal'noy kar'jery v ee sem'je ne \textbf{bylo}. 	&	Predposylok dlya muzykal'noy kar'jery v ee sem'je ne \textbf{byli}.   \\\dash
    {} 	&	 clause\allowbreak \_subj\allowbreak \_predicate\allowbreak \_agreement\allowbreak \_number	&	 Takim obrazom, dlya bol'shikh programm \textbf{prikhodilos'} ispol'zovat' overlei. 	&	Takim obrazom, dlya bol'shikh programm \textbf{prikhodilis'} ispol'zovat' overlei.   \\\dash
    {} 	&	 subj\allowbreak \_predicate\allowbreak \_agreement\allowbreak \_number\allowbreak \_attractor	&	Rasprostranennost' drugikh yazykov \textbf{nevelika}.  	&	Rasprostranennost' drugikh yazykov \textbf{neveliki}.   \\ \dash
    
    
    {} 	&	 noun\allowbreak \_subj\allowbreak \_predicate\allowbreak \_agreement\allowbreak \_gender 	&	 Na territorii kompleksa \textbf{postroen} Kongress-tsentr. 	&	Na territorii kompleksa \textbf{postroena} Kongress-tsentr.   \\ \dash
    {} 	&	 genitive\allowbreak \_subj\allowbreak \_predicate\allowbreak \_agreement\allowbreak \_gender	&	 Ushedshikh iz kluba v dannyj transfernyj period ne \textbf{bylo}. 	&	Ushedshikh iz kluba v dannyj transfernyj period ne \textbf{byla}.   \\ \dash
    {} 	&	 clause\allowbreak \_subj\allowbreak \_predicate\allowbreak \_agreement\allowbreak \_gender	&	Dalee \textbf{neobkhodimo} sdelat' obratnuyu zamenu.  	&	Dalee \textbf{neobkhodima} sdelat' obratnuyu zamenu.   \\ \dash
    {} 	&	 subj\allowbreak \_predicate\allowbreak \_agreement\allowbreak \_gender\allowbreak \_attractor	&	Mestnost' vokrug sela sil'no \textbf{zabolochena}.  	&	Mestnost' vokrug sela sil'no \textbf{zabolocheno}.   \\ \dash
    

    {} 	&	 noun\allowbreak \_subj\allowbreak \_predicate\allowbreak \_agreement\allowbreak \_person 	&	Liturgicheskaya komissiya \textbf{rabotaet} v Monreale.  	&	Liturgicheskaya komissiya \textbf{rabotayu} v Monreale.    \\ \dash
    {} 	&	 genitive\allowbreak \_subj\allowbreak \_predicate\allowbreak \_agreement\allowbreak \_person	&	 Detey u Magnusa i Elizavety ne \textbf{bylo}.	&	Detey u Magnusa i Elizavety ne \textbf{budu}.  \\ \dash
    {} 	&	 clause\allowbreak \_subj\allowbreak \_predicate\allowbreak \_agreement\allowbreak \_person	&	Po otsenkam, \textbf{ostaetsya} raskopat' okolo 350 m.  &	  Po otsenkam, \textbf{ostaesh'sya} raskopat' okolo 350 m. \\
    \midrule

    \multirow{2}{\phenomenon\linewidth}{\textsc{Anaphor \mbox{Agreement}}}  &	 anaphor\allowbreak \_agreement\allowbreak \_number 	&	 Est' neskol'ko rastenij, \textbf{kotorye} mozhno nayti tol'ko v Velikobritanii. 	&	 Est' neskol'ko rastenij, \textbf{kotoroe} mozhno nayti tol'ko v Velikobritanii. 	\\ \dash
    {}  &	 anaphor\allowbreak \_agreement\allowbreak \_gender 	&	Tekhnika, \textbf{kotoruyu} on izobrel, poluchila nazvanie «skul'ptura sveta».  	&	 Tekhnika, \textbf{kotoryj} on izobrel, poluchila nazvanie «skul'ptura sveta». 	\\\dash
    \midrule
    
    \multirow{3}{\phenomenon\linewidth}{\textsc{Noun Phrase \mbox{Agreement}}}  &	 np\allowbreak \_agreement\allowbreak \_number 	&	 No \textbf{malen'kaya} geroinya vashego naroda ostalas' tverda. 	&	 No \textbf{malen'kie} geroinya vashego naroda ostalas' tverda. 	\\ \dash
    &	 np\allowbreak \_agreement\allowbreak \_gender 	&	 Titul luchshej komandy Anglii \textbf{togo} sezona takzhe otoshel «osam». 	&	 Titul luchshej komandy Anglii \textbf{toj} sezona takzhe otoshel «osam». 	\\ \dash
    &	 np\allowbreak \_agreement\allowbreak \_case 	&	Zoloto bylo obnaruzheno v \textbf{etom} rajone v 1923 godu.  	&	 Zoloto bylo obnaruzheno v \textbf{etogo} rajone v 1923 godu . 	\\
    \midrule

    \multirow{3}{\phenomenon\linewidth}{\textsc{Floating Quant. \mbox{Agreement}}}  &	 floating\_quantifier\allowbreak \_agreement\allowbreak \_number 	&	Informatsiyu podtverdili i v \textbf{samoj} shkole.  	&	  Informatsiyu podtverdili i v samoj shkolakh.	\\ \dash
    &	 floating\_quantifier\allowbreak \_agreement\allowbreak \_gender 	&	 Pri etom \textbf{samo} povestvovanie nikuda vas ne gonit. 	&	 Pri etom \textbf{sama} povestvovanie nikuda vas ne gonit. 	\\ \dash
    &	 floating\_quantifier\allowbreak \_agreement\allowbreak \_case 	&	Ego \textbf{samogo} uzhe malo kto priznaet avtoritetom.  	&	 Ego \textbf{samomu} uzhe malo kto priznaet avtoritetom. 	\\
    \midrule
    
    \multirow{1}{\phenomenon\linewidth}{\textsc{Reflexives}} 	&	 external\allowbreak \_possessor 	&	Potomki metsenata perebralis' v Moskvu, gde u \textbf{Zhivago} byl biznes.  	&	 Potomki metsenata perebralis' v Moskvu, gde u \textbf{sebya} byl biznes. 	\\
    \midrule
    	 
    \multirow{3}{\phenomenon\linewidth}{\textsc{Negation}} 	&	 negative\allowbreak \_concord 	& I konechno, nikto \textbf{ne} toropilsya vzyat' vinu na sebya.    & I konechno, nikto toropilsya vzyat' vinu \textbf{ne} na sebya.  \\\dash	 
    {} 	&	 negative\_pronoun\allowbreak \_to\allowbreak \_indefinite 	& Poetomu \textbf{nikogda} ne ostanavlivaytes', vsegda idite vpered!    & Poetomu \textbf{kogda-libo} ne ostanavlivaytes', vsegda idite vpered!    \\	 \dash
    {} 	&	 indefinite\_pronoun\allowbreak \_to\allowbreak \_negative 	&	\textbf{Chto-to} podskazyvaet, chto nechto pokhozhee my uvidim i v Parizhe.  	&   \textbf{Nichto} podskazyvaet, chto nechto pokhozhee my uvidim i v Parizhe. \\
    \midrule																						
    \multirow{5}{\phenomenon\linewidth}{\textsc{Argument \mbox{Structure}}} 	&	 transitive\_verb 	&	Ya rasshchityval, chto budet mnogo smaylov i vse \textbf{zatsenyat} sarkazm. 	&	Ya rasshchityval, chto budet mnogo smaylov i vse \textbf{voskhodyat} sarkazm.   \\ \dash	 
    {} 	&	 transitive\_verb\allowbreak \_subject 	&	\textbf{Shante} teryaet soznanie i snova prosypaetsya v svoej krovati.  	&	\textbf{Khimiya} teryaet soznanie i snova prosypaetsya v svoej krovati.   \\	\dash
    {} 	&	 transitive\_verb\allowbreak \_passive 	& 	Al'tron byl unichtozhen \textbf{Vizhenom}, kotoryj prines sebya v zhertvu. &	Al'tron byl unichtozhen \textbf{navykom}, kotoryj prines sebya v zhertvu.   \\ \dash
    {} 	&	 transitive\_verb\allowbreak \_object 	&	Professor Farnsvort naznachaet \textbf{Lilu} kapitanom kosmicheskogo korablya.  	&	Professor Farnsvort naznachaet \textbf{krug} kapitanom kosmicheskogo korablya.   \\\dash
    {} 	&	 transitive\_verb\allowbreak \_iobject 	&	Nasledniki posle ego smerti  prodali dvorets \textbf{Oginskim}.  	&	Nasledniki posle ego smerti  prodali dvorets \textbf{fragmentam}.   \\ 	
    \midrule 																						
    \multirow{3}{\phenomenon\linewidth}{\textsc{Aspect}} 	&	 change\allowbreak \_duration\allowbreak \_aspect 	&	Pri etom vopros avtorstva dolgo \textbf{ostavalsya} otkrytym.  	&	Pri etom vopros avtorstva dolgo \textbf{ostalsya} otkrytym.   \\ \dash	 
    {} 	&	 change\allowbreak \_repetition\allowbreak \_aspect 	&	Boll kazhdyj god \textbf{posylala} tsvety na den' rozhdeniya svoej podruge.  	&	Boll kazhdyj god \textbf{poslala} tsvety na den' rozhdeniya svoej podruge.   \\ \dash	 
    {} 	&	 deontic\_imperative\allowbreak \_aspect 	&	Vse serii kogda-to zakanchivayutsya, ne stoit etomu \textbf{udelyat'} vnimanie.  	&	Vse serii kogda-to zakanchivayutsya, ne stoit etomu \textbf{udelit'} vnimanie.   \\
    \midrule																						
    \multirow{3}{\phenomenon\linewidth}{\textsc{Tense}} 	&	 single\allowbreak \_verb\allowbreak \_tense 	&	 A vchera on \textbf{dopustil} ochen' grubuyu oshibku. 	&	 A vchera on \textbf{dopustit} ochen' grubuyu oshibku.  \\	 \dash
    {} 	&	 conj\allowbreak \_verb\allowbreak \_tense 	&	 Poslezavtra utrom on uzhe \textbf{pokinet} MKS i budet na Zemle. 	&	 Poslezavtra utrom on uzhe \textbf{pokinul} MKS i budet na Zemle. 	\\	\dash
    {} 	&	 tense\allowbreak \_marker 	&	 Tonnel' na Sinopskoy naberezhnoy otkroyut na \textbf{budushchey} nedele. 	&	 Tonnel' na Sinopskoy naberezhnoy otkroyut na \textbf{minuvshey} nedele. 	\\											
        \bottomrule
    \end{tabular}
\end{adjustbox}
\caption{Examples of all \Nparadigms\ paradigms in \rublimp.}\label{tab:dataset_examples}
\end{table*}

%% file: parts/appendix_documentation.tex
\twocolumn
\section{Minimal Pair Generation}
\label{app:generation}

In this section, we provide a detailed description of the minimal pair generation procedure for each phenomenon in \rublimp. 

\subsection{Morphology}

\subsubsection{\textsc{Word Formation}}
The minimal pairs in this phenomenon are created to violate the principles of affix ordering, namely (i-ii) prefix stacking rules \cite{reynolds2013order}, and (iii) suffixation universals \cite{Greenberg_63}. 

\paragraph{Contexts} We create a list of affixes (including all their possible allomorphs) that we can add or swap and manually annotate, dividing them into two subtypes: derivation and inflection. Thus, we limit the contexts to sentences where at least one word has one or several affixes from the list. This gives us more control when generating minimal pairs since not every random affix change leads to ungrammaticality. Additionally, we limit the number of prefixes a target verb can have to two since having more prefixes is less common in Russian.

\paragraph{Implementation Details}
We search the sentences for a possible target word (e.g., a verb with a prefix) and segment it into morphological elements using \texttt{pymorphy2} and dictionaries from \citet{Bolshakova2021}. To generate the minimal pairs we then (i) add a new prefix to a verb (e.g., \textit{za-pisat'} `to write down' {$\rightarrow$} \textit{*\textbf{pro}-za-pisat'}); swap verb prefixes to change their order (\textit{pri-\textbf{u}-krasit'} `to embellish' {$\rightarrow$} \textit{*\textbf{u}-pri-krasit'}); or add a derivational suffix between the root and existing suffixes (\textit{vodoprovod-n-aya} `tap [water]' {$\rightarrow$} *\textit{vodoprovod-\textbf{ist}-n-aya}). We check that the added affixes co-occur with the root to make the examples more probable, w.r.t. co-occurrence frequency. Finally, we check that the target word does not exist in the \texttt{pymoprhy2} dictionaries to ensure that the obtained word is ungrammatical.

\subsubsection{\textsc{Word Inflection}}
\textsc{Word Inflection} phenomenon includes errors in (i) verb conjugation and declension of (ii) a single noun or (iii) a noun with modifiers.

\paragraph{Contexts} Since the list of inflectional affixes is not unique to every declension and conjugation (i.e., there are intersections between classes), we curate a dictionary of possible suffix perturbations. We create the dictionary so that the new suffix will not be interpreted as a different form of the same conjugation/declension. This way, each suffix replacement will lead to ungrammatical forms.

\paragraph{Implementation Details}
We use the manually crafted dictionaries to violate declension or conjugation of target words. In the verb conjugation violations (i) we replace the verb's inflection with inflection of the opposite conjugation (\texttt{I} {$\leftrightarrow$} \texttt{II}) with the same tense, number and person values. For example, the affix \textit{\textbf{-et}} (\textit{fut.3sg}) of the \texttt{I} conjugation verb \textit{chita-\textbf{et}} `is reading' is replaced with \textit{\textbf{-it}}, the \texttt{II} conjugation affix for the \textit{fut.3sg} verb form.

For the declension violations (ii-iii), we change the inflectional suffixes of a noun to the suffixes of another declension. Similarly to (i), we ensure that the new inflection suffixes preserve the gender and case values of the word. For example, \textit{stol-\textbf{a}} `table' (\texttt{m.sg.gen}, \texttt{II} declension) is changed to \textit{stol-\textbf{i}}, where \textit{\textbf{-i}} is the \texttt{m.sg.gen} affix of the \texttt{III} declension). 

We then check that the resulting word does not contain any combinations of letters that do not exist in Russian. We created a list of non-occurring letter sequences based on RNC data. Finally, we check that the new word form is ungrammatical, using the \texttt{pymoprhy2} dictionaries.

\subsection{Syntax}
\label{app:syntax_docs}

\subsubsection{\textsc{Government}}
\textsc{Government} refers to the government of the grammatical case of a noun, wherein a verb or a preposition determines the grammatical case of its noun phrase complement. We violate the government rules by changing the case of the objects of verbs governing (i) Accusative, ii) Instrumental, (iii) Genitive, (iv) a (pro)noun in a prepositional phrase, or (v) a dependent of the nominalization.

\paragraph{Contexts} Since several adpositions allow different cases (e.g., \textit{v} `in' allows both,  \textit{v dome} `in the house (Locative)' and \textit{v litso} `in the face (Accusative)'), we create a list of adpositions and their allowed cases based on \citet{sichinava2018preposition}. To find nominalizations, we check for words ending with \textit{-nie} as in \textit{odobrenie} `blessing'. Since many modifiers in Russian agree with their heads in number, case, and gender, a change in any of those categories will lead to agreement violations. To ensure that the phenomenon is isolated, we only include the sentences where the target word (i.g. a verb's object, a dependent of a nominalization, or an adposition) has no modifiers. 

\paragraph{Implementation Details} We search the sentences for required constructions (e.g., a noun with a preposition in its dependents) and use \texttt{pymorphy2} to change the form of the target word. Notably, to isolate the phenomenon, we ensure that the resulting word form is not ambiguous, i.e., it cannot be interpreted as two different forms (e.g., \texttt{acc.sg} is often the same as \texttt{nom.pl}).

\subsubsection{\textsc{Subject-Predicate Agreement}}
\textsc{Subject-Predicate Agreement} phenomenon includes agreement errors in the domain of the clause, where the subject controls agreement on the predicate. The predicate is often a verb, but sometimes it is an adjective, a participle, or, rarely, a noun. Subject kinds are described below. Our paradigms include violations of agreement in one of the three features: number, gender, and person, which happen in one of the four contexts: with the nominal subject, with the genitive subject, with the clausal subject, and with any subject, but in the presence of an attractor. 

\paragraph{Contexts}

In an ungrammatical sentence, a single feature of the predicate or the subject is altered in the following contexts:

\begin{itemize}[leftmargin=*,topsep=0pt]
    \setlength\itemsep{0pt}
    \item \underline{Nominal subject}: The subject is a noun phrase (including pronouns) in the nominative case. The predicate agrees with it for number and gender (past tense verbs and adjectives) or number and person (present tense verbs).
    \item \underline{Genitive subject}: The subject is nominal in the genitive case with the predicate negated. The predicate must have default features (\textit{3sg.N}). Only the predicate is altered here (to features other than \textit{3sg.N}).
    \item \underline{Clausal subject}: The subject is a clause. The predicate must have default features (\textit{3sg.N}). Only the predicate is altered here (to features other than \textit{3sg.N}).
    \item \underline{Subject with attractor}: The subject is nominal or a clause, and there exists an attractor in terms of \citet{slioussar2016gender} -- a nominal in the subject tree with features distinct from those of the actual subject. Only the predicate is altered here (to features that match those of the attractor) to produce an attraction error.
\end{itemize}

\paragraph{Implementation Details}
We determine the subject and the predicate via a syntactic parser from \cite{anastasyev2020exploring}. We ensure the subject and the predicate can inflect for at least some features like number, gender, or person. Using \texttt{pymorphy2}, we match syntactic and morphological analyses. We ensure the subject and the predicate agree according to \texttt{pymorphy2} feature analysis. We then perform minimal alternation, one feature at a time, for number, gender, and person. In principle, the subject and the predicate can be alternated. However, we never alternate the subject if it controls the agreement of any other word, except for the predicate, so the change is minimal, and the phenomena are kept distinct. We ensure the changed form is not ambiguous: a changed nominal form should have no homonyms in its declension paradigm. Also, we do not alternate the predicate for gender in sentences where the subject is a proper noun or a word denoting jobs, as these can plausibly agree for either feminine or masculine. A curated list of job words from RNC is used.

\subsubsection{\textsc{Anaphor Agreement}}

\textsc{Anaphor Agreement} phenomenon includes errors in agreement of a relative pronoun (anaphor) with its head noun. These pronouns do not inflect for person, so there are two paradigms: incorrect (i) number or (ii) gender

\paragraph{Contexts} For this phenomenon, we search for relative clauses with a pronoun \textit{kotoryj} `which,' that is not a subject of this relative clause (such nominative relative pronoun fits neither this phenomenon nor \textsc{Subject-Predicate Agreeement} because it always enters two agreement relations simultaneously).

\paragraph{Implementation Details}

We determine the head noun and relative pronoun via a morphosyntactic parser by \citet{anastasyev2020exploring}. Using \texttt{pymorphy2}, we match syntactic and morphological analyses. We ensure the head noun and relative pronoun agree according to \texttt{pymorphy2} feature analysis. We then perform minimal alternation, one feature at a time, for number and gender. In principle, the head noun and the relative pronoun can be alternated. However, we never alternate the head noun or the pronoun if they enter several agreement relations simultaneously, so the change is minimal, and the phenomena are kept distinct.

\subsubsection{\textsc{Noun Phrase Agreement}}

\textsc{Noun Phrase Agreement} phenomenon includes agreement errors in the domain of the noun phrase. Adjectives and adjectival pronouns agree with their head noun; as such, the violations include errors in agreement for (i) number, (ii) gender, and (iii) case.

\paragraph{Contexts}

Here, we search for clauses with noun phrases with a single modifier, as we only alter a word in our phenomena. These modifiers could be adjectives, adjective-like pronouns, numerals, and participles that agree with their head nouns.

\paragraph{Implementation Details}

We determine the head noun and modifier via a morphosyntactic parser by \citet{anastasyev2020exploring}. Using \texttt{pymorphy2}, we match syntactic and morphological analyses. We ensure the head noun and modifier agree according to \texttt{pymorphy2} feature analysis. We then perform minimal alternation, one feature at a time, for number, gender, and case. In principle, the head noun and the modifier can be alternated (the head noun can be alternated for number, but not for the case as that would be the \textsc{Government} phenomenon). We never alternate the head noun if it enters several agreement relations simultaneously, so the change is minimal, and the phenomena are kept distinct.

\subsubsection{\textsc{Floating Quantifier Agreement}}

\textsc{Floating Quantifier Agreement} phenomenon includes errors in agreement of a floating quantifier (or ``intensifier'') \textit{sam} `self' with its antecedent head noun. The violations include incorrect (i) number, (ii) gender, and (iii) case.

\paragraph{Contexts}

For this phenomenon, we search for sentences with a floating quantifier \textit{sam} `self.' We determine its antecedent head noun heuristically (see below). The floating quantifier has some freedom to appear in different spots in the sentence for which we account.

\paragraph{Implementation Details}

The syntactic analysis does not connect the floating modifier to its antecedent noun. In each sentence, we heuristically search the whole clause for a single verbal argument (a subject or an object: direct, indirect, or oblique) that has all the same features as a floating quantifier: the number, the gender, and the case -- this will be its antecedent head noun (highlighted brown in Example \ref{ex:float_quant}). (If such a noun is not found, or more than one is found, we discard the sentence). Then, using \texttt{pymorphy2}, we match syntactic and morphological analyses. We ensure the antecedent noun and modifier agree according to \texttt{pymorphy2} feature analysis. We then perform minimal alternation, one feature at a time, for number, gender, and case. In principle, the head noun and the relative pronoun can be alternated (the head noun can be alternated for number, but not for case as that would be the \textsc{Government} phenomenon). We never alternate the head noun if it enters several agreement relations simultaneously, so the change is minimal, and the phenomena are kept distinct.

\pex[*,labeloffset=0.3em,interpartskip=0.2ex,aboveexskip=2ex,belowexskip=2ex]\label{ex:float_quant}%
\a \textit{Zhdali \colorbox{cb-grey}{samogo} \colorbox{cb-brown!20}{bossa} kompanii.} \\
    `They waited for the company \colorbox{cb-brown!20}{boss[m.sg]} \colorbox{cb-grey}{himself}.'

\a \ljudge{*}\textit{Zhdali\colorbox{cb-grey}{samo} \colorbox{cb-brown!20}{bossa} kompanii.} \\
    `They waited for the company \colorbox{cb-brown!20}{boss[m.sg]} \colorbox{cb-grey}{itself}.'

\xe

\subsubsection{\textsc{Reflexives}}
We only consider the case of an external possessor, a so-called \textit{u}-phrase inside the existential \textit{be}-possessive construction that allows a noun phrase or a personal pronoun but cannot bind a reflexive; see Example (\ref{ex:reflexive}) \cite{arylova2013possession,stassen2013predicative}.

\paragraph{Contexts} We define the appropriate external possessor contexts as sentences with a \textit{be}-verb (\textit{byt'}, \textit{est'}), where a noun phrase or a personal pronoun has the preposition \textit{u} in its dependents. Additionally, we limit the contexts to those sentences where the \textit{u}-phrase preceded the verb. This is required because noun phrases following can be used with a preposition \textit{u} in other contexts, namely locative (e.g., \textit{On byl \textbf{u} \underline{doma}} `He was \textbf{by} \underline{the house}'). However, this interpretation is less common for cases when the \textit{u}-phrase precedes the verb.

\paragraph{Implementation Details}
To create violations, we change the noun phrase or a pronoun to a reflexive pronoun \textit{sebya} `self'. Since the reflexive has no gender, number, or case features, we do not need to inflect it.

\pex[*,labeloffset=0.3em,interpartskip=0.2ex,aboveexskip=2ex,belowexskip=2ex]\label{ex:reflexive}
\a \textit{U \colorbox{cb-grey}{nego} byli druz'ya.}\\
    `\colorbox{cb-grey}{He} had friends.'
\a \textit{U \colorbox{cb-grey}{sebya} byli druz'ya.}\\
    `\colorbox{cb-grey}{Himself} had friends.'
\xe

\subsubsection{\textsc{Negation}}

We implement several ways to violate the rules of \textit{negative concord}, namely (i) shifting the negative particle \textit{ne} from a negated verb to another word in the sentence; replacement of (ii) a negative pronoun with an indefinite one, and (iii) an indefinite pronoun with a negative one.

\paragraph{Contexts}
For this paradigm, we search sentences containing a verb under negation used with a negative pronoun (i-ii) or an indefinite pronoun used with a non-negated verb (iii). We do not consider interrogative and conditional sentences and sentences containing an imperative, as their syntactic structures differ from affirmative sentences. 

\paragraph{Implementation Details}
To create violations for paradigm (i), we move the negative particle \textit{ne} `not' from a verb to the head of another noun, adjective, or another phrase. We ensure that the particle is moved not randomly but to specific syntactic constructions to avoid non-logical combinations of words. Such constructions can be negated in other contexts. Thus, the resulting combinations are more plausible and natural. Our systematic approach to replacing a negative pronoun with an indefinite one (and vice versa) ensures that only some replacements lead to ungrammatical sentences. We curate a list of possible replacements, which consistently lead to the violation of negative concord. This list is then systematically applied to paradigms (ii-iii), resulting in the necessary changes to the pronouns.

\subsection{Semantics}
\subsubsection{\textsc{Argument Structure}}
\textsc{Argument Structure} phenomenon includes errors in the verb's argument structure. Similarly to \blimp, we focus on cases where the animacy requirement for the arguments of a transitive verb (from now on in this section -- TV) is violated due to the verb, subject, and object replacement. Additionally, we include a more straightforward case, employing the differences between the argument structure of a transitive and an intransitive verb. Thus, the paradigms include swapping: (i) a TV with an intransitive one; an animate subject of a TV in (ii) active or (iii) passive voice with its inanimate object or replacing it with a random inanimate word; (iv) animate direct object of a TV with a random inanimate word; (v) animate indirect object of a TV with an inanimate subject, or replacing it with a random inanimate word.

\paragraph{Contexts} We consider sentences with a transitive verb in finite form, active or passive, with an inanimate object, both direct and indirect. TVs sometimes allow inanimate subjects, typically metaphorically, so we limit allowed contexts using the RNC semantic annotation. We avoid subjects with semantics of heterogeneous groups of people (e.g., \textit{crowd}); organizations (\textit{bank}); events (\textit{elections}); instruments,  weapons, and their parts (\textit{gun, bullet}); means of transport (\textit{bus}); space, place, and time (\textit{planet, spring}); and proper nouns (\textit{Moscow}). For paradigm (v), we search for sentences with an open clausal complement (\texttt{xcomp}) dependent on an animate object and following the said object.

\paragraph{Implementation Details} To generate minimal pairs for this paradigm, we filter the sentences with a transitive verb and check their dependents for the required arguments. In cases where several arguments are swapped places (paradigms ii-v), to isolate the phenomenon, we ensure that the words to be swapped do not have any modifiers, ensuring that no agreement errors appear after the perturbation. We also make sure to inflect the swapped words to preserve sentence structure. For transitivity (i), that includes replacing a verb with a verb of the same aspect, tense, number, person, and gender values. Subject and object swaps include sampling the nouns with the same number and gender features as the original. See Example (\ref{ex:arg_str}), the TV is \underline{underlined}, the original subject and object are highlighted in \colorbox{cb-grey}{gray} and \colorbox{cb-brown!20}{brown}, respectively. Both subject and object have the same gender category (feminine) and number (singular), so we can swap them. In the generated sentence (b), the original object \textit{sumku} `the bag' takes the Agent argument of the TV, which requires it to be in Nominative, so we change its case from Accusative to Nominative and do the opposite for the object \textit{ona} `she' (Nominative), which becomes \textit{ee} `her' (Accusative).

\pex[*,labeloffset=0.3em,interpartskip=0.2ex,aboveexskip=2ex,belowexskip=2ex]\label{ex:arg_str}
\a \textit{\colorbox{cb-grey}{Ona} \underline{ostavila} \colorbox{cb-brown!20}{sumku} na stole.}\\
    `\colorbox{cb-grey}{She} \underline{left} \colorbox{cb-brown!20}{the bag} on the table.'
\a \ljudge{*}\textit{\colorbox{cb-brown!20}{Sumka} \underline{ostavila} \colorbox{cb-grey}{ee} na stole.}\\
    `\colorbox{cb-brown!20}{The bag} left \colorbox{cb-grey}{her} on the table.'
\xe

\subsubsection{\textsc{Aspect}}
\textsc{Aspect} is the grammatical category of verbs that indicates whether an action is complete (perfective) or incomplete (imperfective) at a particular time. Such semantic difference limits the contexts where each category of verbs can be used, so we employ this to generate minimal pairs for this phenomenon. We replace an imperfective verb with a perfective one in the following contexts, which do not allow a perfective verb: (i) duration; (ii) repetition; contexts with a negated  deontic verb, which only allows a (iii) single or (iv) conjoined imperfective \cite{DeHaan2002,Paducheva:2010}.

\paragraph{Contexts} We curate a list of words and constructions that indicate the required semantics and use them to filter the contexts. The following lexical cues are used:
\begin{itemize}[leftmargin=*,topsep=0pt]
    \setlength\itemsep{0pt}
    \item \underline{Duration} (i): \textit{dolgo}, \textit{dlitel'no}, \textit{prodolzhitel'no}, all with the semantics of `continuously, for a long time'.
    \item \underline{Repetition} (ii): \textit{kazhdyj} `every' + X construction, where X is a noun denoting a time period, such as \textit{kazhdyj den'/god} `every day/year', etc.; and adverbs like \textit{ezhechasno} /\textit{ezheminutno} `ocurring every hour/minute'.
    \item \underline{Deontic modality} (iii-iv): \textit{stoit} and \textit{sleduet} `should', \textit{nado} and \textit{nuzhno} `need'.
\end{itemize}

\paragraph{Implementation Details} To generate minimal pairs, we find sentences with an imperfective verb and check its dependents for one of the lexical cues from the list. We then use a dictionary of aspect pairs \cite{Zaliznyak1987} to change the verb with its perfective counterpart. Note that for some verbs, the dictionary presents several possible versions of pairs (e.g., \textit{sbrasyvat'} `to throw' has two perfective forms: \textit{sbrosit'} and \textit{sbrosat'})). We filter the dictionary by IPM and only leave the pairs with the higher frequency.

\subsubsection{\textsc{Tense}}
The phenomenon focuses on the semantics of tense, expressed in sentences with a tense-marked verb in the presence of a temporal adverbial. We include three paradigms: incorrect choice of a (i) single or (ii) conjoined verb form in a sentence with temporal adverbial, and (iii) wrong temporal adverbial in a sentence with a tense-marked verb.

\paragraph{Contexts} 
We only consider sentences with a perfective verb in future or past tense. This way, we ensure that the pairs are minimal and that the perturbations would lead to ungrammaticality. Additionally, we filter out clausal complements that are verbs to account for constructions like \textit{sobirayus' sdelat'} ``am going to do'', which can be used with markers of both past and future tenses when changed.

To find sentences with the required semantics, we look for a temporal adverbial -- a word or an expression that specifies the time of the event. We include several types of such expressions: 
\begin{itemize}[leftmargin=*,topsep=0pt]
    \setlength\itemsep{0pt}
    \item \underline{Adverbs}: simple one word expressions like \textit{vchera} `yesterday', \textit{zavtra} `tomorrow', etc. We curate a list of adverbs using RNC.
    \item \underline{Adpositional Phrases}: \texttt{PREP + ADJ + NOUN} constructions, such as \textit{v sleduyushchij raz} `next time', \textit{na proshloj nedele} `last week', etc. 
    \item \underline{Numerical Phrases}: constructions of the type \texttt{NUM + NOUN(pl) + ADP}, e.g.,  \textit{neskol'ko dnej nazad} `a few days ago', \textit{paru nedel' nazad} `a couple of weeks ago', etc. 
\end{itemize}

\paragraph{Implementation Details}
\noindent To introduce ungrammaticality, we find sentences that include a verb in past or future tense and check its dependents for one of the temporal adverbials from the list. We change the verb form or the temporal adverbial to the one of the `opposite' tense (future {$\leftrightarrow$} past). Example (\ref{ex:tense}) illustrates the two possible perturbations. We can either change the verb form \textit{poletit} `will fly' to \textit{poletel} `flew', or \textit{zavtra} `tomorrow' to \textit{vhera} `yesterday'. Both alterations result in ungrammatical sentences.

\pex[*,labeloffset=0.3em,interpartskip=0.2ex,aboveexskip=2ex,belowexskip=2ex]\label{ex:tense}
\a \textit{\colorbox{cb-brown!20}{Zavtra} on \colorbox{cb-grey}{poletit} v Italiyu.}\\
    `\colorbox{cb-brown!20}{Tomorrow} he \colorbox{cb-grey}{will fly} to Italy.'
\a \ljudge{*}\textit{Zavtra on \colorbox{cb-grey}{poletel} v Italiyu.}
\a \ljudge{*}\textit{\colorbox{cb-brown!20}{Vchera} on poletit v Italiyu.}

\xe

%% file: parts/appendix_data_validation_guidelines.tex
\onecolumn
\section{Human Validation}
\label{sec: app_data_validation}
\subsection{Annotation Guidelines}
\label{sec: app_data_validation_guidelines}

\subsubsection*{Annotation Task: Verify the quality of a linguistic minimal pair}

\paragraph{Overview} Judge the correctness of a given minimal pair in which the grammatical sentence is taken from the corpus of natural texts, and the ungrammatical sentence is automatically generated using expert-written rules and natural language processing tools. 

\paragraph{What is a minimal pair?} A minimal pair consists of two sentences that differ in grammatical acceptability due to a single morphological, syntactic, or semantic feature. Please note that the minimal pair should isolate only one linguistic feature, such as number, gender, case, and more. The ungrammatical sentence is obtained by perturbing the grammatical one using one of the following operations.:
\begin{itemize}[topsep=5pt]
    \setlength\itemsep{0pt}
    \item Changing a feature, e.g., changing of one inflectional category: number, case, gender, tense, etc.
    \item Replacing a word, e.g., replacing a lexeme while maintaining the original grammatical form;
    \item Swapping two words in a sentence;
    \item Moving a word to another position.
\end{itemize}

\noindent \textbf{Your task} 
\begin{enumerate}[topsep=5pt]
    \setlength\itemsep{0pt}
    \item Carefully read the grammatical and ungrammatical sentences and the linguistic feature that should be isolated.
    \item Decide whether the minimal pair is designed correctly. Does it isolate the specified linguistic feature?
    \item If everything is correct, select ``Yes''.
    \item If the minimal pair is implausible, does not isolate the mentioned feature, contains two grammatical sentences, perturbs multiple sentence units or linguistic features, select ``No''.
    \item If the original sentence is ungrammatical, select ``N/A''.
    \item If there are any typos, please state them in the box.
\end{enumerate}

\noindent Do you have any questions or difficulties with completing your task? Reach out in our group chat.

\vspace{0.5cm}

\noindent \textit{The guidelines further provide an extensive list of minimal pair examples for each paradigm and annotation examples for each answer option. You can access the complete guidelines in our GitHub repository.}

\vspace{1cm}
{\begin{minipage}{8cm}
\par\noindent\rule{\textwidth}{1pt}
\textbf{\large Example of web interface}
\vspace{5pt}

Minimal pair

\vspace{0.1cm}

\colorbox{cb-grey}{This is a toy grammatical sentence.}

\colorbox{cb-grey}{*This \textbf{are} a toy ungrammatical sentence.}

\vspace{10pt}

Phenomenon

\vspace{0.1cm}

\colorbox{cb-grey}{This is the linguistic feature.}

\vspace{10pt}

Is the minimal pair designed correctly?

\vspace{0.2cm}

\setlength\parindent{50pt}$\ocircle$ Yes  $\ocircle$ No $\ocircle$ N/A

\vspace{5pt}

\noindent
Comment
\vspace{0.1cm}

\noindent\framebox{\parbox{0.95\textwidth}{\color{Gray!125}\centering Enter your comment}}

\vspace{0.1cm}

\par\noindent\rule{\textwidth}{1pt}

\end{minipage}}

%% file: parts/appendix_data_validation_results.tex
\subsection{Data validation results} \label{sec: app_data_validation_results}

\input{tables/david_skene_by_phenomenon}

%% file: tables/david_skene_by_phenomenon.tex
\begin{table}[ht!]
\centering
\resizebox{0.9 \textwidth}{!}{
\begin{tabular}{p{0.15\textwidth}lrr}
\toprule
\textbf{Phenomenon} & \textbf{Paradigm} &\textbf{\%} & \textbf{WAWA} \\
 \midrule
    \multirow{3}{\phenomenon\linewidth}{\textsc{Word \mbox{Formation}}} 
        & Addition of Extra Morphemes: Uninterpretable Suffix Combinations & 93.48 & 91.1 \\
        & Addition of Extra Morphemes: Verb Prefixes & 97.83 & 93.6 \\
        & Morpheme Permutation: Verb Prefixes & 96.00 & 93.8 \\
 
 \midrule
    \multirow{3}{\phenomenon\linewidth}{\textsc{Word \mbox{Inflection}}}
        & Replacement of Inflectional Affixes: Noun Declensions (Simple) &98.00 & 96.0 \\
        & Replacement of Inflectional Affixes: Declensions of Nouns With Agreeing Dependents & 94.00 & 89.7 \\
        & Inflectional Affixes: Verbal Conjugation Swap & 94.00 & 96.0 \\
 \midrule
 
    \multirow{5}{\phenomenon\linewidth}{\textsc{Government}} 
        & Prepositional Government & 100.00 & 94.0 \\
        & Verbal Government: Direct Object & 87.50 & 89.7 \\
        & Verbal Government: Genitive Object & 93.62 & 88.8 \\
        & Verbal Government: Object in Instrumental Case & 100.00 &100.0 \\
        & Verbal Government: Nominalizations & 78.05 & 86.7 \\  
 \midrule
 
    \multirow{11}{\phenomenon\linewidth}{\textsc{Subject-predicate \mbox{Agreement}}} 
        & Subject-Predicate Agreement (Number) & 96.00 & 96.0 \\
        & Genitive Subject-Predicate Agreement (Number) & 85.71 & 89.2 \\
        & Clausal Subject-Predicate Agreement (Number) & 97.83 & 88.0 \\
        & Subject-Predicate Agreement in Presence of an Attractor (Number) & 100.00 & 93.6 \\

        & Subject-Predicate Agreement (Gender) & 97.96 & 94.5 \\
        & Genitive Subject-Predicate Agreement (Gender) & 91.84 & 93.6 \\
        & Clausal Subject-Predicate Agreement (Gender) & 100.00 & 91.5 \\
        & Subject-Predicate Agreement in Presence of an Attractor (Gender) & 97.96 & 92.4 \\

        & Subject-Predicate Agreement (Person) & 100.00 & 99.3 \\
        & Genitive Subject-Predicate Agreement (Person) & 89.36 & 85.8 \\
        & Clausal Subject-Predicate Agreement (Person) & 97.96 & 93.2 \\   
 \midrule

    \multirow{2}{\phenomenon\linewidth}{\textsc{Anaphor \mbox{Agreement}}}
        & Anaphor Agreement (Number) & 92.68 & 93.2 \\
        & Anaphor Agreement (Gender) & 95.45 & 92.8 \\
 \midrule
 
    \multirow{3}{\phenomenon\linewidth}{\textsc{Noun Phrase \mbox{Agreement}}} 
        & Noun Phrase Agreement (Number) & 91.49 & 92.3 \\
        & Noun Phrase Agreement (Gender) & 98.00 & 95.5 \\
        & Noun Phrase Agreement (Case) & 100.00 & 95.2 \\
 \midrule

    \multirow{3}{\phenomenon\linewidth}{\textsc{Floating Quant. \mbox{Agreement}}} 
        & Floating Quantifier Agreement (Number) & 95.92 & 87.8 \\
        & Floating Quantifier Agreement (Gender) & 97.92 & 96.0 \\
        & Floating Quantifier Agreement (Case) & 98.00 & 93.3 \\
 \midrule

  \multirow{1}{\phenomenon\linewidth}{\textsc{Reflexives}} 
        & External Possessor & 100.00 & 96.5 \\
  \midrule

    \multirow{3}{\phenomenon\linewidth}{\textsc{Negation}} 
        & Negative Concord & 100.00 & 95.6 \\
        & Replacement of a Negative Pronoun with an Indefinite One & 80.00 & 87.8 \\
        & Replacement of an Indefinite Pronoun with a Negative One & 100.00 & 94.4 \\
 \midrule
 
    \multirow{5}{\phenomenon\linewidth}{\textsc{Argument \mbox{Structure}}} 
        & Transitivity & 97.67 & 91.4 \\
        & Animate Subject of a Transitive Verb & 94.00 & 86.4 \\
        & Animate Subject of a Passive Verb & 93.88 & 92.7 \\
        & Animate Direct Object of a Transitive Verb & 82.00 & 82.4 \\
        & Animate Indirect Object of a Transitive Verb & 100.00 & 96.8 \\
        
 \midrule 
 
    \multirow{3}{\phenomenon\linewidth}{\textsc{Aspect}} 
        & Incompatibility of the Perfective with the Semantics of Duration & 92.00 & 92.7 \\
        & Impossibility of the Perfective in Repetitive Situations & 97.83 & 91.2 \\
        & Impossibility of the Perfective Under Negated Strong Deontic Verbs & 96.00 & 95.0 \\
 \midrule 
 
    \multirow{3}{\phenomenon\linewidth}{\textsc{Tense}}
        & Tense & 95.92 & 92.6 \\
        & Tense (Coordination) & 87.50 & 89.3 \\
        & Tense Markers & 97.96 & 94.4 \\
\bottomrule

\end{tabular}}
\caption{The per-paradigm ratios of plausible minimal pairs (\%) and WAWA inter-annotator agreement rates.}
\label{tab:ds_paradigm}
\end{table}

%% file: parts/appendix_data_stats.tex
\onecolumn
\section{Statistics for Syntactic Patterns}
\label{app:dataset_stats}
We extract syntactic structures from a grammatical sentence's dependency tree to compute a high-level diversity w.r.t. syntactic patterns in \rublimp. Using expert-written rules, we linearize the dependency tree by merging its subtrees into a single constituent. We never merge the verb arguments with it and parse the main and dependent clauses similarly. We then compute the total number of unique patterns and the pattern frequency at the benchmark level. Consider Example \ref{ex:tree_1} for the sentence \textit{Poiski novogo oruzhiya zaderzhali ubijstvo Potioreka} ``The searches for a new weapon slowed down the  murder of Potiorek'', where we extract the sentence's syntactic structure as \textsc{NP V-trans NP} (transitive verb). We provide the word translations with the articles and prepositions in the same nodes for illustration purposes.

\ex\label{ex:tree_1}
\begin{forest}
[$\text{NP V-trans NP}$ 
    [NP 
        [N [Poiski,tier=ru [The\_searches\_for, tier=transl]]]
        [NP 
            [A [novogo,tier=ru [a\_new, tier=transl]]]
            [N [oruzhiya,tier=ru [weapon, tier=transl]]]
        ]
    ]
    [VP
        [V [zaderzhali,tier=ru [slowed\_down, tier=transl]]]
        [NP 
            [N [ubijstvo, tier=ru [the\_murder, tier=transl]]]
            [N [Potioreka, tier=ru [of\_Potiorek, tier=transl]]]
        ]
    ]
]
\end{forest}
\xe

%% file: parts/appendix_human_baseline.tex
\onecolumn
\section{Human Baseline}
\label{sec: app_human_baseline}
\subsection{Annotation Guidelines}

\subsubsection*{Select a Grammatical Sentence}

\noindent \textbf{Your task} 
\begin{enumerate}
    \item Carefully read two sentences.
    \item Determine which of the two sentences is grammatical (a Russian native speaker would say or write like this).
    \item Choose ``Sentence \#1'' if the first sentence is grammatical, or choose ``Sentence \#2'' otherwise.
    \item If there are any typos, please state them in the box.
\end{enumerate}

\noindent Below, you can find annotation examples and examples of possible grammatical errors. For clarity, we mark the sentences with a grammatical error with the ``*'' symbol and highlighted the word in bold.

\noindent Choose the sentence that has \emph{no} grammatical errors. If you find a given pair of sentences difficult, choose the sentence that seems \textit{more} natural and \textit{more} grammatically correct from your perspective.

\vspace{0.5cm}

\noindent \textit{The guidelines further provide an extensive list of minimal pair examples for each paradigm and annotation examples for each answer option. You can access the complete guidelines in our GitHub repository.}

\vspace{1cm}
\begin{minipage}{8cm}

\par\noindent\rule{\textwidth}{1pt}
\textbf{Example of web interface}
\vspace{5pt}

Which of the two sentences has no errors? 
\vspace{0.1cm}

\colorbox{cb-grey}{1. This is a toy sentence \#1.}

\colorbox{cb-grey}{2. This is a toy sentence \#2.}

\vspace{5pt}
\begin{itemize}[noitemsep,topsep=0pt]
    \item[$\ocircle$] Sentence \#1
    \item[$\ocircle$] Sentence \#2
\end{itemize}

\vspace{5pt}
\noindent
Comment
\vspace{0.1cm}

\framebox{\parbox{0.95\textwidth}{\color{Gray!125}\centering Enter your comment}}

\vspace{0.1cm}

\par\noindent\rule{\textwidth}{1pt}
\end{minipage}

%% file: parts/appendix_add_results.tex
\section{Fine-grained Results}
\label{app:add_res}

\input{tables/monolingual_paradigms_scores}

\input{tables/multi_rublimp_paradigms_p1}
\input{tables/multi_rublimp_paradigms_p2}

\begin{figure*}[t!]
    \centering
    \subfloat[Wikipedia]{\includegraphics[width=0.32\textwidth]{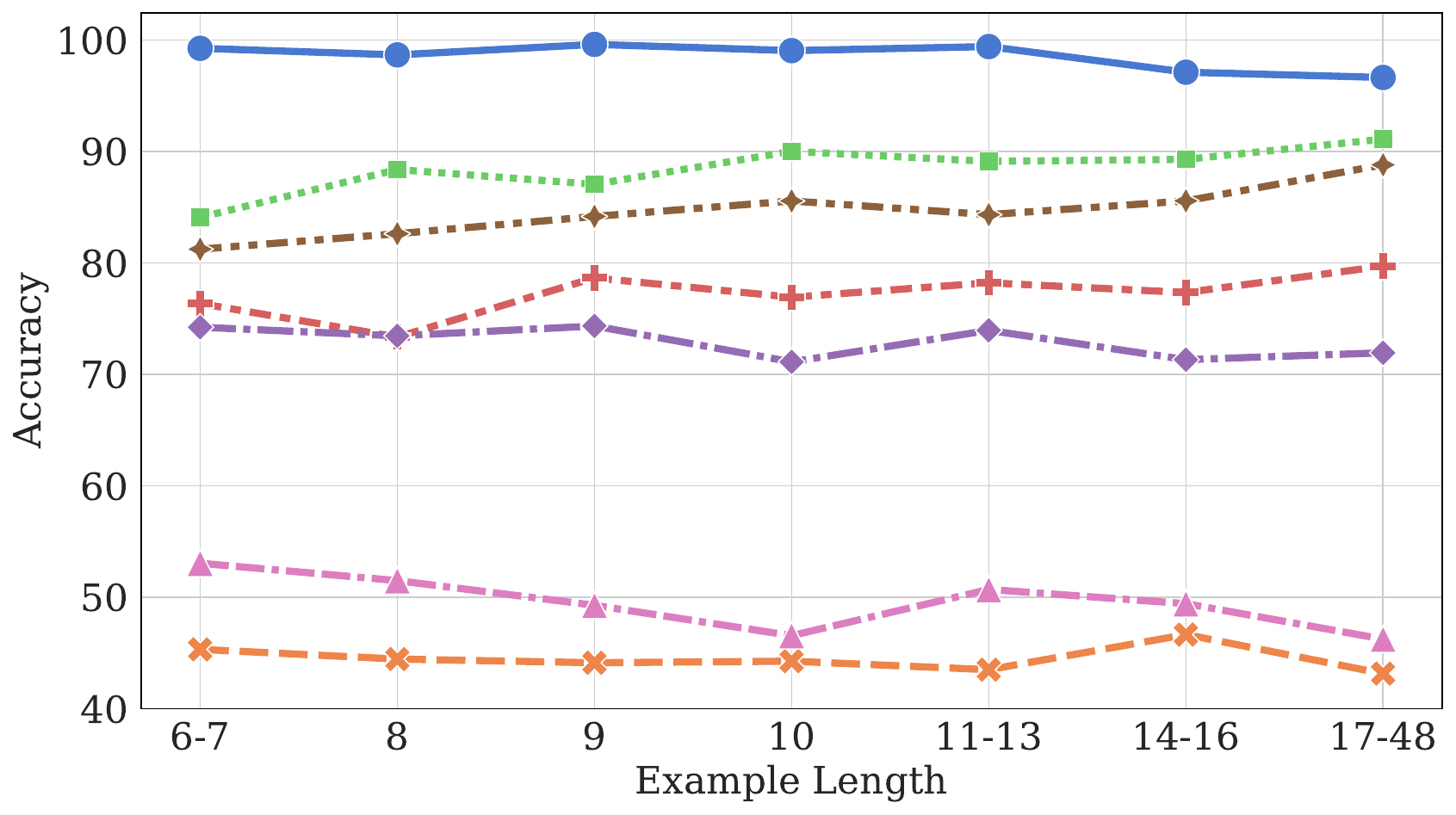}}
    \subfloat[Wikinews]{\includegraphics[width=0.32\textwidth]{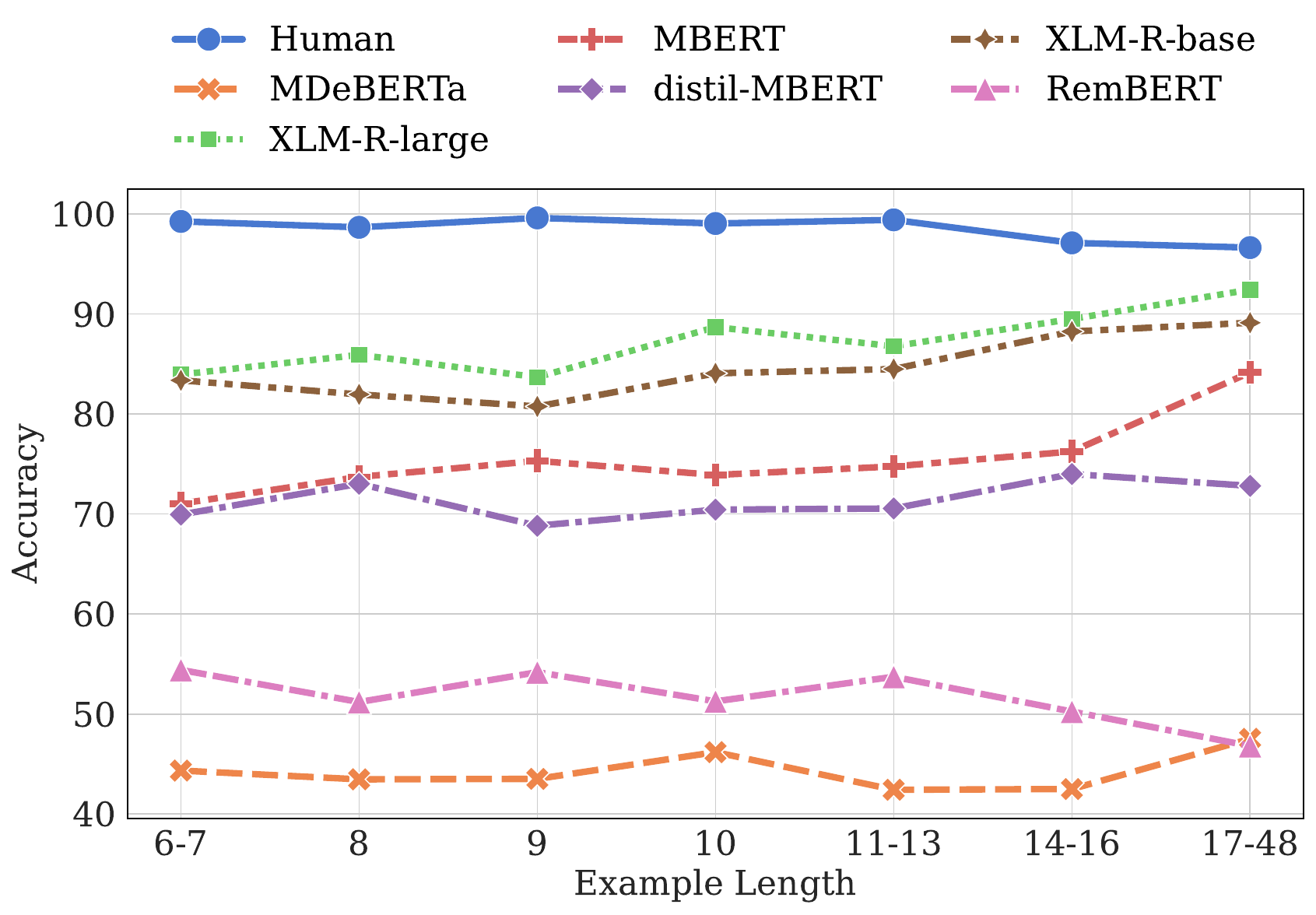}}
    \subfloat[Librusec]{\includegraphics[width=0.32\textwidth]{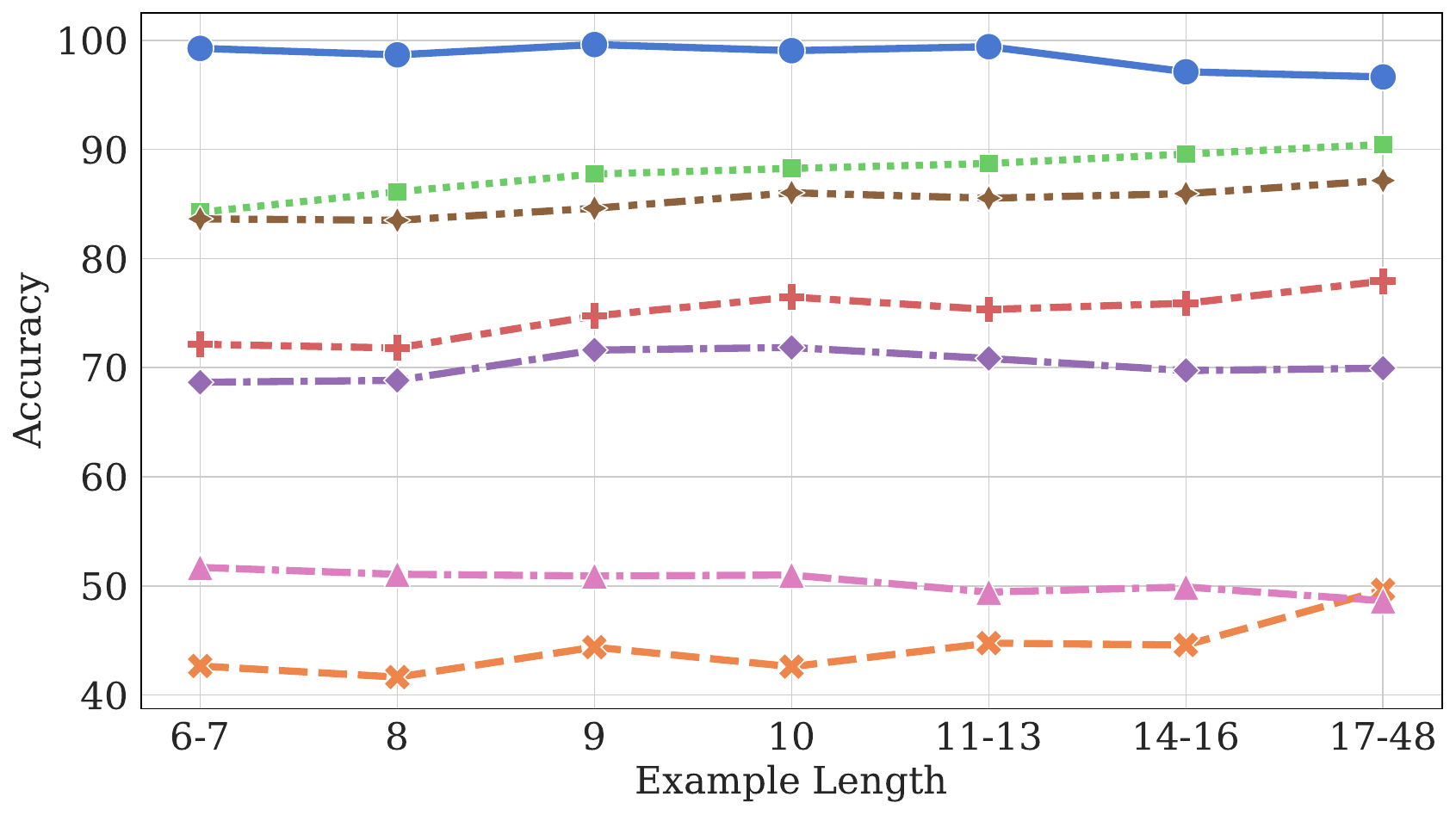}}
    \caption{Results on \rublimp for the multilingual encoder-only LMs per domain grouped by seven quintiles of the length.}
    \label{fig:length_multi_enc}
\end{figure*}

\begin{figure*}[t!]
    \centering
    \subfloat[Wikipedia]{\includegraphics[width=0.32\textwidth]{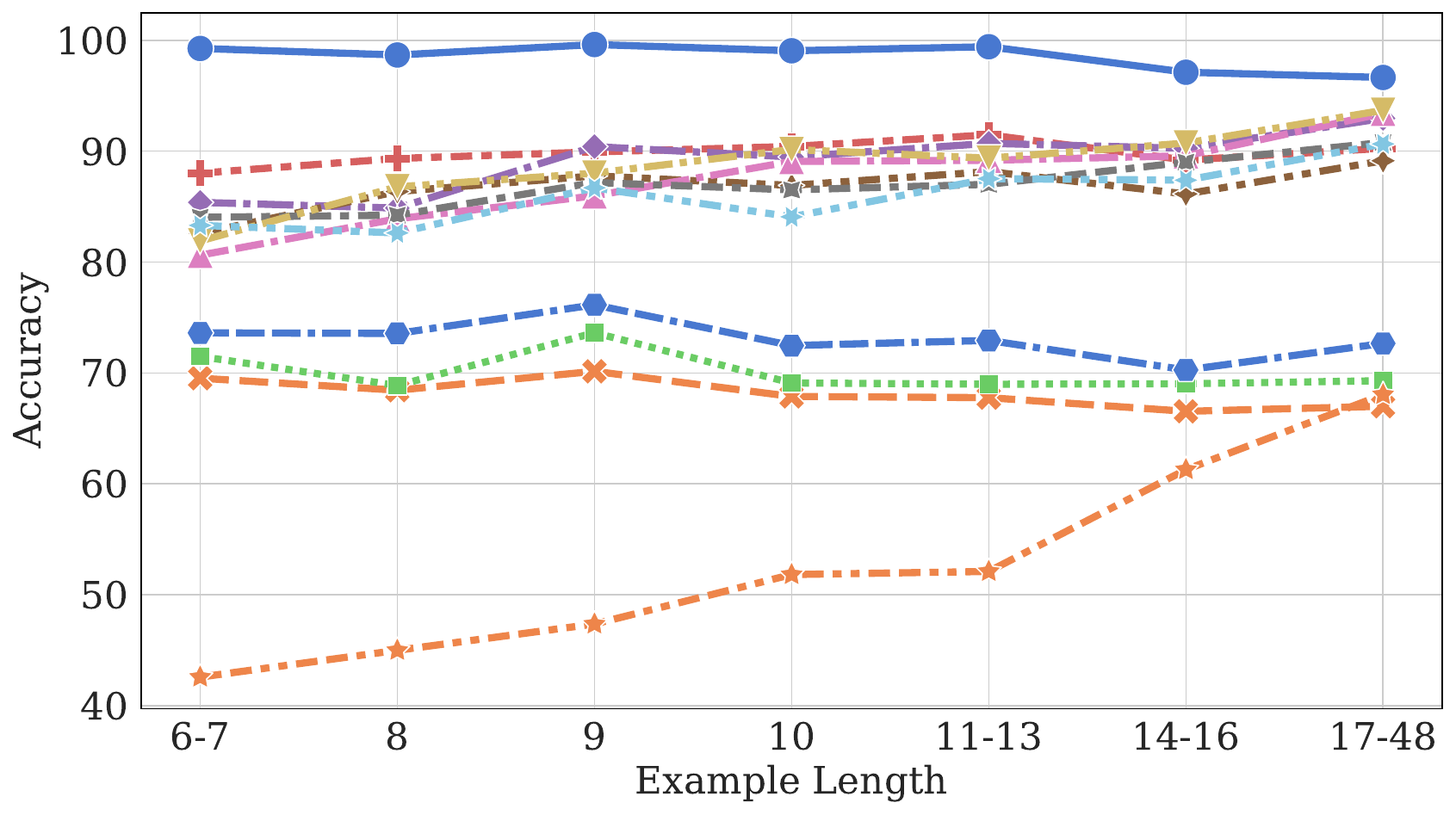}}
    \subfloat[Wikinews]{\includegraphics[width=0.32\textwidth]{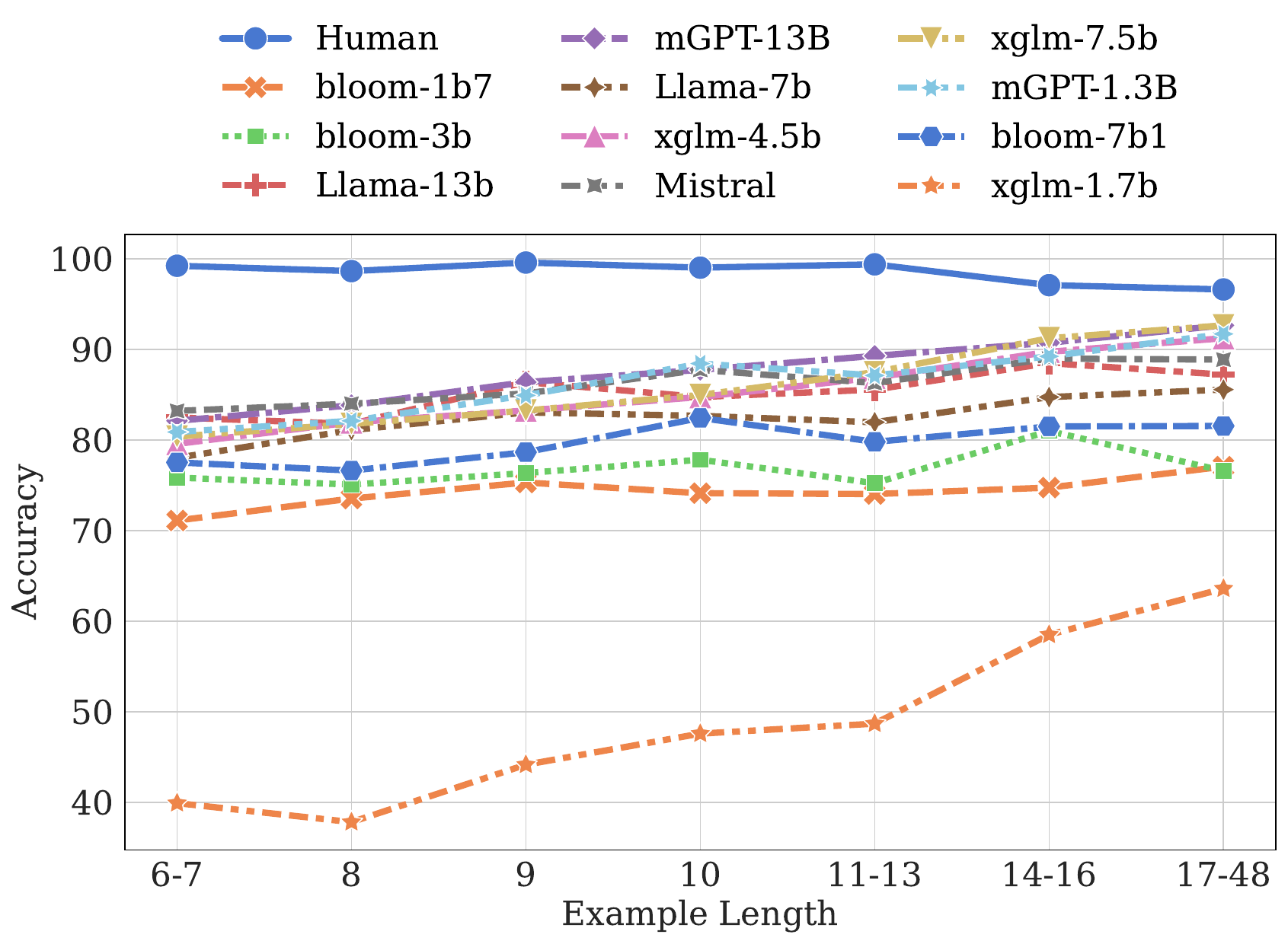}}
    \subfloat[Librusec]{\includegraphics[width=0.32\textwidth]{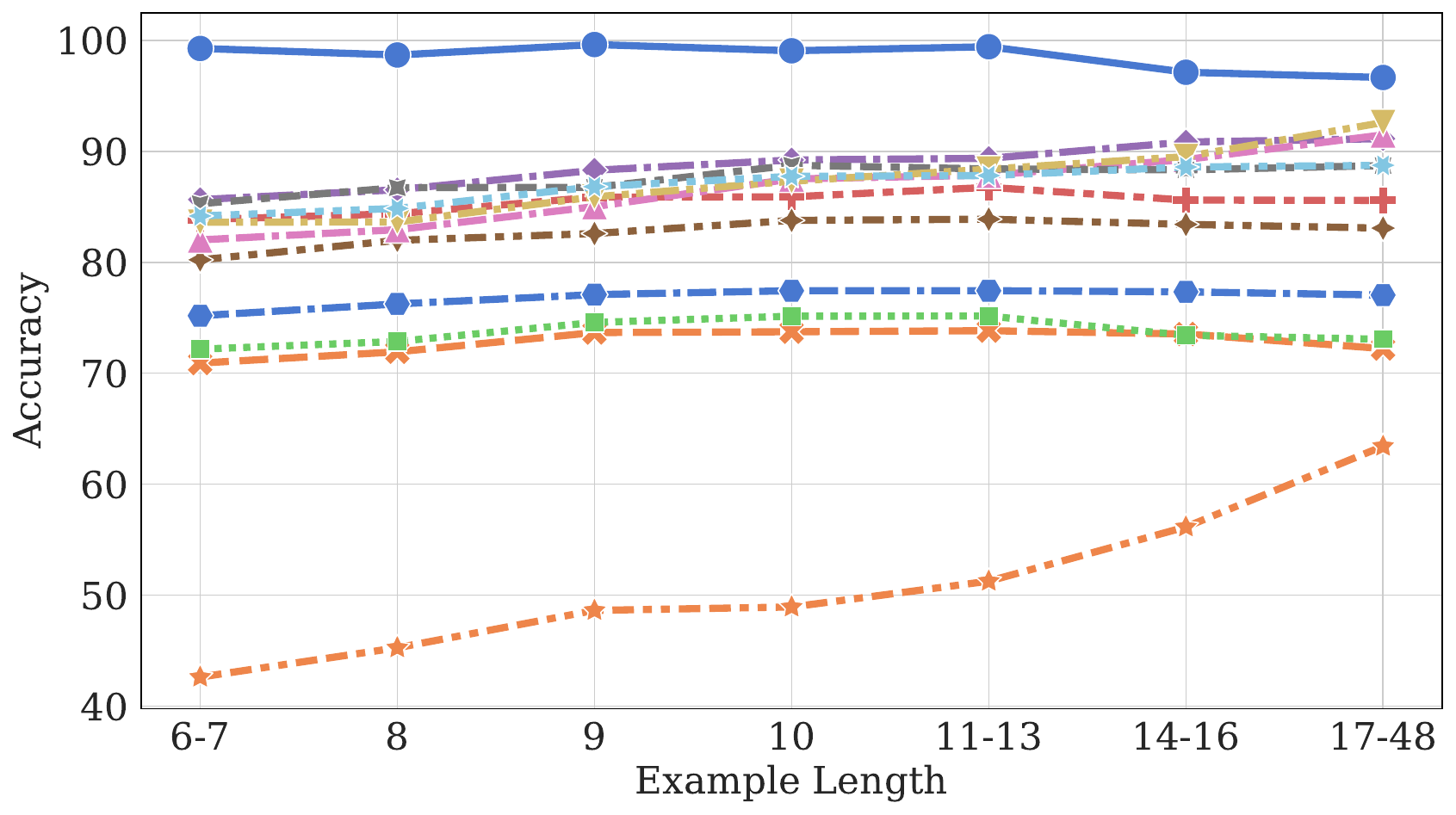}}
    \caption{Results on \rublimp for the multilingual decoder-only LMs per domain grouped by seven quintiles of the length.}
    \label{fig:length_multi_dec}
\end{figure*}

%% file: tables/monolingual_paradigms_scores.tex
\begin{table*}[ht!]

\resizebox{\textwidth}{!}{%
    \begin{tabular}{llRRRRRRRRR}
Phenomenon & PID & \multicolumn{1}{p{2ex}}{\rotatebox{\rotation}{\textsc{\rubertb}}} & \multicolumn{1}{p{2ex}}{\rotatebox{\rotation}{\textsc{\rubertl}}} & \multicolumn{1}{p{2ex}}{\rotatebox{\rotation}{\textsc{\ruroberta}}} & \multicolumn{1}{p{2ex}}{\rotatebox{\rotation}{\textsc{\rugpts}}} & \multicolumn{1}{p{2ex}}{\rotatebox{\rotation}{\textsc{\rugptm}}} & \multicolumn{1}{p{2ex}}{\rotatebox{\rotation}{\textsc{\rugptl}}} & \multicolumn{1}{p{2ex}}{\rotatebox{\rotation}{\textsc{\rugpttpf}}} & \multicolumn{1}{p{2ex}}{\rotatebox{\rotation}{\textsc{\smblingorb}}} & \multicolumn{1}{p{2.5ex}}{\rotatebox{\rotation}{\textbf{Human}}}\\
\midrule
\multirow{3}{\phenomenon\linewidth}{\textsc{Word \mbox{Formation}}} 
    & add\allowbreak \_new\allowbreak \_suffix & 78.30 & 79.50 & 95.30 & 94.80 & 95.80 & 94.30 & 97.70 & 80.60 & 100.0 \\
    & add\allowbreak \_verb\allowbreak \_prefix & 74.40 & 74.80 & 77.70 & 72.90 & 81.30 & 74.40 & 86.90 & 61.50 & 100.0 \\
    & change\allowbreak \_verb\allowbreak \_prefixes\allowbreak \_order & 93.00 & 94.20 & 96.00 & 97.90 & 98.20 & 99.00 & 98.40 & 97.50 & 100.0 \\
\midrule
\multirow{3}{\phenomenon\linewidth}{\textsc{Word \mbox{Inflection}}}
    & change\allowbreak \_declension\allowbreak \_ending & 83.50 & 84.40 & 92.00 & 90.40 & 85.60 & 90.80 & 95.30 & 86.80 & 100.0 \\
    & change\allowbreak \_declension\allowbreak \_ending\allowbreak \_has\allowbreak \_dep & 86.50 & 88.10 & 94.90 & 94.30 & 86.80 & 95.30 & 97.40 & 92.80 & 100.0 \\
    & change\allowbreak \_verb\allowbreak \_conjugation & 84.60 & 85.60 & 88.00 & 90.00 & 86.70 & 88.00 & 92.90 & 77.60 & 98.00 \\
\midrule

\multirow{5}{\phenomenon\linewidth}{\textsc{Government}}
    & adposition\allowbreak \_government & 95.80 & 96.40 & 97.80 & 94.60 & 95.80 & 95.40 & 97.30 & 90.30 & 100.0 \\
    & verb\allowbreak \_acc\allowbreak \_object & 93.80 & 93.50 & 96.10 & 92.20 & 93.60 & 93.90 & 97.10 & 89.70 & 100.0 \\
    & verb\allowbreak \_gen\allowbreak \_object & 93.20 & 94.10 & 94.30 & 88.30 & 91.90 & 90.90 & 95.70 & 77.10 & 98.00 \\
    & verb\allowbreak \_ins\allowbreak \_object & 78.70 & 77.30 & 99.20 & 94.50 & 96.60 & 96.60 & 98.00 & 96.50 & 100.0 \\
    & nominalization\allowbreak \_case & 92.10 & 92.00 & 96.00 & 95.10 & 96.50 & 96.10 & 97.40 & 92.40 & 96.00 \\
\midrule

\multirow{11}{\phenomenon\linewidth}{\textsc{Subject-predicate \mbox{Agreement}}}
    & noun\allowbreak \_subj\allowbreak \_predicate\allowbreak \_agreement\allowbreak \_number & 91.70 & 92.70 & 95.40 & 90.30 & 92.20 & 92.20 & 96.00 & 86.60 & 98.00 \\
    & genitive\allowbreak \_subj\allowbreak \_predicate\allowbreak \_agreement\allowbreak \_number & 95.30 & 95.60 & 96.00 & 95.60 & 96.20 & 96.70 & 97.90 & 82.90 & 97.96 \\
    & clause\allowbreak \_subj\allowbreak \_predicate\allowbreak \_agreement\allowbreak \_number & 90.60 & 89.60 & 89.80 & 91.40 & 93.80 & 91.90 & 96.00 & 73.50 & 97.96 \\
    & subj\allowbreak \_predicate\allowbreak \_agreement\allowbreak \_number\allowbreak \_attractor & 92.70 & 92.30 & 96.50 & 90.50 & 91.40 & 92.10 & 96.20 & 87.60 & 100.0 \\
    & noun\allowbreak \_subj\allowbreak \_predicate\allowbreak \_agreement\allowbreak \_gender & 83.00 & 83.60 & 88.40 & 81.50 & 83.70 & 84.00 & 90.90 & 80.20 & 98.00 \\
    & genitive\allowbreak \_subj\allowbreak \_predicate\allowbreak \_agreement\allowbreak \_gender & 97.10 & 97.40 & 97.00 & 96.40 & 96.00 & 96.80 & 98.60 & 89.50 & 98.00 \\
    & clause\allowbreak \_subj\allowbreak \_predicate\allowbreak \_agreement\allowbreak \_gender & 97.00 & 96.10 & 94.00 & 94.30 & 95.30 & 94.60 & 97.90 & 79.90 & 100.0 \\
    & subj\allowbreak \_predicate\allowbreak \_agreement\allowbreak \_gender\allowbreak \_attractor & 88.40 & 89.50 & 92.60 & 84.90 & 86.80 & 87.10 & 94.90 & 83.90 & 98.00 \\
    & noun\allowbreak \_subj\allowbreak \_predicate\allowbreak \_agreement\allowbreak \_person & 86.40 & 86.90 & 93.40 & 86.10 & 86.40 & 87.00 & 94.60 & 79.40 & 100.0 \\
    & genitive\allowbreak \_subj\allowbreak \_predicate\allowbreak \_agreement\allowbreak \_person & 87.80 & 89.60 & 92.60 & 92.50 & 92.50 & 93.10 & 97.90 & 78.90 & 98.00 \\
    & clause\allowbreak \_subj\allowbreak \_predicate\allowbreak \_agreement\allowbreak \_person & 92.80 & 92.40 & 94.10 & 90.10 & 93.00 & 91.10 & 96.40 & 66.90 & 97.96 \\
\midrule

\multirow{2}{\phenomenon\linewidth}{\textsc{Anaphor \mbox{Agreement}}}
    & anaphor\allowbreak \_agreement\allowbreak \_number & 84.10 & 83.70 & 93.20 & 92.70 & 93.80 & 93.90 & 95.30 & 87.50 & 98.00 \\

    & anaphor\allowbreak \_agreement\allowbreak \_gender & 87.70 & 89.00 & 98.00 & 95.90 & 98.00 & 97.70 & 98.80 & 98.40 & 98.00 \\
\midrule

\multirow{3}{\phenomenon\linewidth}{\textsc{Noun Phrase \mbox{Agreement}}}
    & np\allowbreak \_agreement\allowbreak \_number & 82.80 & 84.90 & 84.20 & 94.70 & 97.20 & 96.70 & 98.60 & 90.60 & 100.0 \\
    & np\allowbreak \_agreement\allowbreak \_gender & 79.50 & 80.90 & 97.00 & 93.40 & 96.20 & 95.10 & 97.40 & 83.20 & 98.00 \\
    & np\allowbreak \_agreement\allowbreak \_case & 88.40 & 88.30 & 85.30 & 97.90 & 98.70 & 98.50 & 99.40 & 95.70 & 98.00 \\
\midrule

\multirow{3}{\phenomenon\linewidth}{\textsc{Floating Quant. \mbox{Agreement}}}
    & floating\allowbreak \_quantifier\allowbreak \_agreement\allowbreak \_number & 83.30 & 85.20 & 96.60 & 89.50 & 93.60 & 93.00 & 98.10 & 83.20 & 100.0 \\
    & floating\allowbreak \_quantifier\allowbreak \_agreement\allowbreak \_gender & 95.40 & 94.30 & 93.30 & 79.80 & 96.50 & 83.30 & 97.70 & 97.60 & 98.00 \\
    & floating\allowbreak \_quantifier\allowbreak \_agreement\allowbreak \_case & 95.50 & 94.20 & 98.60 & 93.60 & 98.40 & 96.20 & 98.70 & 90.50 & 100.0 \\
\midrule

\multirow{1}{\phenomenon\linewidth}{\textsc{Reflexives}}
    & external\allowbreak \_posessor & 78.70 & 81.00 & 91.10 & 83.20 & 79.90 & 87.80 & 94.70 & 96.20 & 98.00 \\
\midrule

\multirow{3}{\phenomenon\linewidth}{\textsc{Negation}}
    & negative\allowbreak \_concord & 99.50 & 99.20 & 99.70 & 99.90 & 99.90 & 99.90 & 100.0 & 99.80 & 100.0 \\
    & negative\allowbreak \_pronoun\allowbreak \_to\allowbreak \_indefinite & 33.90 & 47.40 & 71.10 & 19.90 & 41.80 & 36.40 & 62.60 & 33.40 & 100.0 \\
    & indefinite\allowbreak \_pronoun\allowbreak \_to\allowbreak \_negative & 99.90 & 99.80 & 98.70 & 99.70 & 99.90 & 99.50 & 100.0 & 99.70 & 100.0 \\
    \midrule

\multirow{5}{\phenomenon\linewidth}{\textsc{Argument \mbox{Structure}}}
    & transitive\allowbreak \_verb & 96.50 & 96.40 & 98.60 & 93.00 & 95.40 & 95.40 & 98.60 & 77.90 & 100.0 \\
    & transitive\allowbreak \_verb\allowbreak \_subject & 83.60 & 85.40 & 81.10 & 79.00 & 83.50 & 84.40 & 90.30 & 74.60 & 100.0 \\
    & transitive\allowbreak \_verb\allowbreak \_passive & 90.00 & 90.50 & 93.10 & 89.90 & 94.10 & 93.80 & 98.20 & 91.40 & 100.0 \\
    & transitive\allowbreak \_verb\allowbreak \_object & 88.00 & 87.20 & 93.60 & 94.30 & 96.50 & 97.10 & 98.40 & 86.60 & 100.0 \\
    & transitive\allowbreak \_verb\allowbreak \_iobject & 84.50 & 86.50 & 91.80 & 87.90 & 90.40 & 90.50 & 96.20 & 83.20 & 100.0 \\
    \midrule

\multirow{3}{\phenomenon\linewidth}{\textsc{Aspect}}
    & change\allowbreak \_duration\allowbreak \_aspect & 96.20 & 96.50 & 97.10 & 92.60 & 94.60 & 94.60 & 97.00 & 85.40 & 100.0 \\
    & change\allowbreak \_repetition\allowbreak \_aspect & 95.20 & 95.50 & 97.00 & 94.30 & 95.60 & 95.30 & 97.80 & 91.00 & 100.0 \\
    & deontic\allowbreak \_imperative\allowbreak \_aspect & 96.80 & 97.40 & 97.50 & 94.90 & 96.60 & 96.70 & 98.50 & 85.80 & 100.0 \\
    \midrule

\multirow{3}{\phenomenon\linewidth}{\textsc{Tense}}
    & single\allowbreak \_verb\allowbreak \_tense & 85.00 & 86.80 & 87.80 & 76.30 & 82.60 & 81.70 & 90.70 & 69.30 & 100.0 \\
    & conj\allowbreak \_verb\allowbreak \_tense & 93.00 & 93.00 & 96.90 & 92.60 & 94.60 & 94.60 & 98.50 & 85.40 & 100.0 \\
    & tense\allowbreak \_marker & 83.50 & 84.70 & 92.50 & 84.00 & 88.90 & 84.80 & 96.90 & 86.70 & 98.00 \\
\midrule
\multicolumn{2}{r}{\textbf{Average}} & 87.95 & 88.74 & 93.13 & 89.28 & 91.62 & 91.29 & 95.86 & 84.56 & 99.15 \\
\bottomrule
\end{tabular}
}
\caption{Accuracy scores (\%) for the monolingual LMs by paradigm. Random baseline is 50\%.}
\label{tab:rublimp_phenomena_mono}
\end{table*}

%% file: tables/multi_rublimp_paradigms_p1.tex
\begin{table*}

\resizebox{\textwidth}{!}{
    \begin{tabular}{llRRRRRRRRR}
Phenomenon & PID & \multicolumn{1}{p{2.5ex}}{\rotatebox{\rotation}{\textsc{\dmbert}}} & \multicolumn{1}{p{2.5ex}}{\rotatebox{\rotation}{\textsc{\mbert}}} & \multicolumn{1}{p{2.5ex}}{\rotatebox{\rotation}{\textsc{\xlmrb}}} & \multicolumn{1}{p{2.5ex}}{\rotatebox{\rotation}{\textsc{\xlmrl}}} & \multicolumn{1}{p{2.5ex}}{\rotatebox{\rotation}{\textsc{\rembert}}} & \multicolumn{1}{p{2.5ex}}{\rotatebox{\rotation}{\textsc{\mdeberta}}} & \multicolumn{1}{p{2.5ex}}{\rotatebox{\rotation}{\textsc{\mgpt}}} & \multicolumn{1}{p{2.5ex}}{\rotatebox{\rotation}{\textsc{\mgptb}}} & \multicolumn{1}{p{2.5ex}}{\rotatebox{\rotation}{\textbf{Human}}} \\
\midrule
\multirow{3}{\phenomenon\linewidth}{\textsc{Word \mbox{Formation}}} 
    & add\allowbreak \_new\allowbreak \_suffix & 86.70 & 91.20 & 93.20 & 94.10 & 48.30 & 49.40 & 97.20 & 97.20 & 100.0 \\
    & add\allowbreak \_verb\allowbreak \_prefix & 81.50 & 86.70 & 82.20 & 81.30 & 53.20 & 62.30 & 87.30 & 87.30 & 100.0 \\
    & change\allowbreak \_verb\allowbreak \_prefixes\allowbreak \_order & 83.70 & 88.60 & 90.30 & 91.00 & 52.70 & 46.00 & 98.60 & 99.10 & 100.0 \\
\midrule
\multirow{3}{\phenomenon\linewidth}{\textsc{Word \mbox{Inflection}}}
    & change\allowbreak \_declension\allowbreak \_ending & 79.80 & 81.30 & 89.40 & 90.20 & 56.30 & 43.80 & 92.70 & 92.30 & 100.0 \\
    & change\allowbreak \_declension\allowbreak \_ending\allowbreak \_has\allowbreak \_dep & 86.10 & 89.50 & 93.90 & 94.80 & 58.80 & 40.00 & 97.40 & 97.30 & 100.0 \\
    & change\allowbreak \_verb\allowbreak \_conjugation & 73.00 & 83.10 & 88.40 & 88.10 & 49.00 & 47.10 & 97.80 & 97.00 & 98.00 \\
\midrule

\multirow{5}{\phenomenon\linewidth}{\textsc{Government}}
    & adposition\allowbreak \_government & 80.00 & 85.90 & 91.10 & 92.00 & 55.10 & 47.60 & 92.70 & 93.20 & 100.0 \\
    & verb\allowbreak \_acc\allowbreak \_object & 63.20 & 67.30 & 82.50 & 85.90 & 43.40 & 48.60 & 84.90 & 89.90 & 100.0 \\
    & verb\allowbreak \_gen\allowbreak \_object & 57.00 & 64.90 & 82.20 & 84.70 & 47.70 & 31.70 & 83.60 & 87.40 & 98.00 \\
    & verb\allowbreak \_ins\allowbreak \_object & 72.20 & 89.70 & 94.30 & 96.80 & 44.20 & 49.80 & 95.20 & 95.70 & 100.0 \\
    & nominalization\allowbreak \_case & 81.80 & 86.60 & 92.00 & 93.20 & 54.10 & 59.80 & 91.80 & 94.20 & 96.00 \\
\midrule

\multirow{11}{\phenomenon\linewidth}{\textsc{Subject-predicate \mbox{Agreement}}}
    & noun\allowbreak \_subj\allowbreak \_predicate\allowbreak \_agreement\allowbreak \_number & 74.50 & 79.20 & 87.70 & 89.30 & 51.70 & 34.80 & 88.60 & 89.70 & 98.00 \\
    & genitive\allowbreak \_subj\allowbreak \_predicate\allowbreak \_agreement\allowbreak \_number & 85.70 & 86.90 & 91.10 & 92.00 & 45.70 & 43.00 & 95.50 & 96.20 & 97.96 \\
    & clause\allowbreak \_subj\allowbreak \_predicate\allowbreak \_agreement\allowbreak \_number & 72.50 & 82.80 & 87.90 & 80.70 & 38.40 & 26.00 & 94.60 & 95.60 & 97.96 \\
    & subj\allowbreak \_predicate\allowbreak \_agreement\allowbreak \_number\allowbreak \_attractor & 69.10 & 76.90 & 88.90 & 89.90 & 49.10 & 40.10 & 84.50 & 87.20 & 100.0 \\
    & noun\allowbreak \_subj\allowbreak \_predicate\allowbreak \_agreement\allowbreak \_gender & 72.10 & 72.70 & 82.00 & 84.20 & 52.50 & 36.50 & 82.00 & 83.80 & 98.00 \\
    & genitive\allowbreak \_subj\allowbreak \_predicate\allowbreak \_agreement\allowbreak \_gender & 86.80 & 92.10 & 94.50 & 94.80 & 42.60 & 51.40 & 98.10 & 98.20 & 98.00 \\
    & clause\allowbreak \_subj\allowbreak \_predicate\allowbreak \_agreement\allowbreak \_gender & 74.50 & 82.90 & 89.90 & 86.50 & 43.80 & 27.10 & 87.60 & 88.00 & 100.0 \\
    & subj\allowbreak \_predicate\allowbreak \_agreement\allowbreak \_gender\allowbreak \_attractor & 70.30 & 78.50 & 85.10 & 86.50 & 54.70 & 34.40 & 84.20 & 86.40 & 98.00 \\
    & noun\allowbreak \_subj\allowbreak \_predicate\allowbreak \_agreement\allowbreak \_person & 67.30 & 67.50 & 84.40 & 85.10 & 52.70 & 39.60 & 74.20 & 76.90 & 100.0 \\
    & genitive\allowbreak \_subj\allowbreak \_predicate\allowbreak \_agreement\allowbreak \_person & 80.50 & 76.00 & 86.20 & 88.40 & 62.20 & 39.20 & 82.70 & 82.70 & 98.00 \\
    & clause\allowbreak \_subj\allowbreak \_predicate\allowbreak \_agreement\allowbreak \_person & 81.60 & 88.60 & 85.30 & 87.60 & 54.10 & 32.40 & 92.60 & 91.50 & 97.96 \\
\midrule

\multirow{2}{\phenomenon\linewidth}{\textsc{Anaphor \mbox{Agreement}}}
    & anaphor\allowbreak \_agreement\allowbreak \_number & 74.70 & 82.30 & 89.60 & 89.80 & 45.30 & 65.40 & 90.80 & 92.40 & 98.00 \\

    & anaphor\allowbreak \_agreement\allowbreak \_gender & 30.00 & 90.40 & 94.10 & 96.50 & 18.80 & 85.30 & 93.50 & 96.30 & 98.00 \\

\midrule

\multirow{3}{\phenomenon\linewidth}{\textsc{Noun Phrase \mbox{Agreement}}}
    & np\allowbreak \_agreement\allowbreak \_number & 78.80 & 88.40 & 93.20 & 94.70 & 48.20 & 51.60 & 95.30 & 96.20 & 100.0 \\
    & np\allowbreak \_agreement\allowbreak \_gender & 69.10 & 80.10 & 88.90 & 91.60 & 55.70 & 26.50 & 89.50 & 91.10 & 98.00 \\
    & np\allowbreak \_agreement\allowbreak \_case & 90.40 & 92.70 & 95.90 & 96.80 & 49.60 & 45.00 & 97.80 & 98.70 & 98.00 \\
\midrule

\multirow{3}{\phenomenon\linewidth}{\textsc{Floating Quant. \mbox{Agreement}}}
    & floating\allowbreak \_quantifier\allowbreak \_agreement\allowbreak \_number & 71.00 & 77.20 & 88.40 & 90.40 & 57.40 & 40.20 & 88.20 & 90.70 & 100.0 \\
    & floating\allowbreak \_quantifier\allowbreak \_agreement\allowbreak \_gender & 88.40 & 83.30 & 94.90 & 95.70 & 79.10 & 33.40 & 74.80 & 83.00 & 98.00 \\
    & floating\allowbreak \_quantifier\allowbreak \_agreement\allowbreak \_case & 90.00 & 87.80 & 95.60 & 93.10 & 51.40 & 38.70 & 92.40 & 91.80 & 100.0 \\
\midrule

\multirow{1}{\phenomenon\linewidth}{\textsc{Reflexives}}
    & external\allowbreak \_posessor & 56.00 & 52.90 & 69.90 & 79.10 & 45.30 & 40.20 & 82.50 & 85.60 & 98.00 \\
\midrule

\multirow{3}{\phenomenon\linewidth}{\textsc{Negation}}
    & negative\allowbreak \_concord & 96.30 & 97.90 & 99.20 & 98.80 & 54.60 & 55.30 & 99.80 & 99.70 & 100.0 \\
    & negative\allowbreak \_pronoun\allowbreak \_to\allowbreak \_indefinite & 50.80 & 17.30 & 19.10 & 43.90 & 93.00 & 54.90 & 3.90 & 5.40 & 100.0 \\
    & indefinite\allowbreak \_pronoun\allowbreak \_to\allowbreak \_negative & 78.70 & 83.70 & 99.20 & 99.30 & 5.90 & 20.50 & 99.70 & 99.70 & 100.0 \\
    \midrule

\multirow{5}{\phenomenon\linewidth}{\textsc{Argument \mbox{Structure}}}
    & transitive\allowbreak \_verb & 62.40 & 71.70 & 84.80 & 89.40 & 46.50 & 38.40 & 82.20 & 86.00 & 100.0 \\
    & transitive\allowbreak \_verb\allowbreak \_subject & 53.80 & 56.00 & 64.80 & 69.60 & 49.60 & 42.70 & 70.40 & 73.70 & 100.0 \\
    & transitive\allowbreak \_verb\allowbreak \_passive & 63.60 & 69.40 & 79.70 & 85.50 & 44.80 & 39.40 & 87.00 & 91.20 & 100.0 \\
    & transitive\allowbreak \_verb\allowbreak \_object & 48.40 & 53.60 & 73.80 & 80.70 & 52.40 & 45.40 & 80.20 & 86.00 & 100.0 \\
    & transitive\allowbreak \_verb\allowbreak \_iobject & 51.20 & 55.40 & 74.30 & 81.30 & 53.10 & 43.60 & 79.90 & 85.40 & 100.0 \\
    \midrule

\multirow{3}{\phenomenon\linewidth}{\textsc{Aspect}}
    & change\allowbreak \_duration\allowbreak \_aspect & 58.80 & 62.20 & 80.70 & 87.50 & 54.90 & 43.80 & 83.00 & 85.00 & 100.0 \\
    & change\allowbreak \_repetition\allowbreak \_aspect & 58.70 & 62.90 & 80.40 & 87.70 & 48.10 & 47.20 & 86.90 & 89.00 & 100.0 \\
    & deontic\allowbreak \_imperative\allowbreak \_aspect & 60.90 & 54.20 & 83.60 & 87.90 & 54.20 & 41.30 & 86.40 & 89.10 & 100.0 \\
    \midrule

\multirow{3}{\phenomenon\linewidth}{\textsc{Tense}}
    & single\allowbreak \_verb\allowbreak \_tense & 63.10 & 58.00 & 66.40 & 72.40 & 54.00 & 48.90 & 65.70 & 71.50 & 100.0 \\
    & conj\allowbreak \_verb\allowbreak \_tense & 71.60 & 70.20 & 79.30 & 86.60 & 47.80 & 55.50 & 83.00 & 87.10 & 100.0 \\
    & tense\allowbreak \_marker & 32.80 & 29.90 & 78.30 & 80.30 & 54.80 & 56.20 & 89.90 & 90.50 & 98.00 \\
    \midrule
    
\multicolumn{2}{r}{\textbf{Average}} & 70.65 & 75.03 & 84.81 & 87.46 & 50.55 & 44.22 & 86.37 & 88.26 & 99.15 \\
\bottomrule
\end{tabular}
}
\caption{Accuracy scores (\%) for the multilingual LMs by paradigm (part 1). Random baseline is 50\%. }
\label{tab:rublimp_phenomena_multi_p1}
\end{table*}

%% file: tables/multi_rublimp_paradigms_p2.tex
\begin{table*}

\resizebox{\textwidth}{!}{
    \begin{tabular}{llRRRRRRRRRRR}
Phenomenon & PID & \multicolumn{1}{p{2.5ex}}{\rotatebox{\rotation}{\textsc{\blooms}}} & \multicolumn{1}{p{2.5ex}}{\rotatebox{\rotation}{\textsc{\bloomm}}} & \multicolumn{1}{p{2.5ex}}{\rotatebox{\rotation}{\textsc{\blooml}}} & \multicolumn{1}{p{2.5ex}}{\rotatebox{\rotation}{\textsc{\xglms}}} & \multicolumn{1}{p{2.5ex}}{\rotatebox{\rotation}{\textsc{\xglmm}}} & \multicolumn{1}{p{2.5ex}}{\rotatebox{\rotation}{\textsc{\xglml}}} & \multicolumn{1}{p{2.5ex}}{\rotatebox{\rotation}{\textsc{\llamas}}} & \multicolumn{1}{p{2.5ex}}{\rotatebox{\rotation}{\textsc{\llamam}}} & \multicolumn{1}{p{2.5ex}}{\rotatebox{\rotation}{\textsc{\mistral}}} & \multicolumn{1}{p{2.5ex}}{\rotatebox{\rotation}{\textbf{Human}}} \\
\midrule
\multirow{3}{\phenomenon\linewidth}{\textsc{Word \mbox{Formation}}} 
    & add\allowbreak \_new\allowbreak \_suffix & 90.30 & 90.90 & 91.40 & 23.80 & 96.20 & 96.60 & 93.30 & 94.90 & 96.50 & 100.00 \\
    & add\allowbreak \_verb\allowbreak \_prefix & 90.50 & 92.10 & 92.30 & 17.20 & 82.00 & 83.10 & 92.10 & 93.40 & 95.40 & 100.00 \\
    & change\allowbreak \_verb\allowbreak \_prefixes\allowbreak \_order & 77.50 & 85.60 & 83.00 & 72.10 & 99.00 & 98.70 & 98.70 & 99.20 & 98.70 & 100.00 \\
\midrule
\multirow{3}{\phenomenon\linewidth}{\textsc{Word \mbox{Inflection}}}
    & change\allowbreak \_declension\allowbreak \_ending & 86.00 & 85.40 & 87.80 & 43.10 & 91.30 & 92.20 & 90.50 & 92.10 & 93.50 & 100.0 \\
    & change\allowbreak \_declension\allowbreak \_ending\allowbreak \_has\allowbreak \_dep & 90.20 & 91.40 & 93.90 & 47.90 & 95.30 & 96.20 & 93.30 & 96.20 & 97.00 & 100.0 \\
    & change\allowbreak \_verb\allowbreak \_conjugation & 92.90 & 93.90 & 93.90 & 44.10 & 89.90 & 91.90 & 88.70 & 92.20 & 94.50 & 98.00 \\
\midrule

\multirow{5}{\phenomenon\linewidth}{\textsc{Government}}
    & adposition\allowbreak \_government & 70.00 & 73.50 & 77.00 & 56.90 & 90.40 & 91.80 & 88.40 & 90.90 & 92.50 & 100.0 \\
    & verb\allowbreak \_acc\allowbreak \_object & 65.80 & 69.30 & 71.50 & 43.90 & 81.80 & 82.30 & 85.70 & 88.70 & 87.50 & 100.0 \\
    & verb\allowbreak \_gen\allowbreak \_object & 60.80 & 64.30 & 63.90 & 35.40 & 81.60 & 79.30 & 69.70 & 73.80 & 77.40 & 98.00 \\
    & verb\allowbreak \_ins\allowbreak \_object & 65.30 & 69.70 & 71.40 & 68.90 & 93.60 & 95.10 & 90.30 & 93.50 & 87.50 & 100.0 \\
    & nominalization\allowbreak \_case & 77.40 & 80.40 & 84.30 & 53.50 & 92.40 & 93.80 & 91.90 & 95.60 & 95.90 & 96.00 \\
\midrule

\multirow{11}{\phenomenon\linewidth}{\textsc{Subject-predicate \mbox{Agreement}}}
    & noun\allowbreak \_subj\allowbreak \_predicate\allowbreak \_agreement\allowbreak \_number & 80.90 & 79.50 & 85.00 & 45.20 & 87.20 & 87.80 & 85.40 & 86.00 & 90.40 & 98.00 \\
    & genitive\allowbreak \_subj\allowbreak \_predicate\allowbreak \_agreement\allowbreak \_number & 89.10 & 89.90 & 91.70 & 48.40 & 89.10 & 90.90 & 90.00 & 91.40 & 95.80 & 97.96 \\
    & clause\allowbreak \_subj\allowbreak \_predicate\allowbreak \_agreement\allowbreak \_number & 93.50 & 95.00 & 95.00 & 31.90 & 79.40 & 83.80 & 95.60 & 96.50 & 96.90 & 97.96 \\
    & subj\allowbreak \_predicate\allowbreak \_agreement\allowbreak \_number\allowbreak \_attractor & 75.10 & 76.20 & 83.20 & 56.90 & 84.50 & 85.90 & 84.20 & 86.50 & 87.00 & 100.0 \\
    & noun\allowbreak \_subj\allowbreak \_predicate\allowbreak \_agreement\allowbreak \_gender & 70.50 & 72.00 & 74.90 & 44.90 & 79.00 & 78.80 & 86.30 & 88.20 & 90.00 & 98.00 \\
    & genitive\allowbreak \_subj\allowbreak \_predicate\allowbreak \_agreement\allowbreak \_gender & 95.50 & 95.80 & 94.10 & 51.30 & 91.20 & 91.90 & 95.70 & 96.70 & 96.60 & 98.00 \\
    & clause\allowbreak \_subj\allowbreak \_predicate\allowbreak \_agreement\allowbreak \_gender & 91.80 & 94.60 & 94.70 & 45.60 & 88.80 & 89.70 & 95.20 & 96.10 & 96.70 & 100.0 \\
    & subj\allowbreak \_predicate\allowbreak \_agreement\allowbreak \_gender\allowbreak \_attractor & 69.80 & 73.10 & 75.80 & 51.90 & 81.40 & 81.20 & 84.90 & 87.60 & 88.00 & 98.00 \\
    & noun\allowbreak \_subj\allowbreak \_predicate\allowbreak \_agreement\allowbreak \_person & 85.10 & 87.20 & 91.10 & 43.60 & 76.00 & 76.90 & 82.70 & 86.50 & 87.30 & 100.0 \\
    & genitive\allowbreak \_subj\allowbreak \_predicate\allowbreak \_agreement\allowbreak \_person & 93.10 & 93.00 & 94.80 & 37.20 & 72.30 & 73.50 & 89.30 & 91.70 & 97.00 & 98.00 \\
    & clause\allowbreak \_subj\allowbreak \_predicate\allowbreak \_agreement\allowbreak \_person & 96.70 & 97.00 & 97.70 & 30.00 & 80.80 & 80.80 & 94.70 & 96.30 & 97.20 & 97.96 \\
\midrule

\multirow{2}{\phenomenon\linewidth}{\textsc{Anaphor \mbox{Agreement}}}
    & anaphor\allowbreak \_agreement\allowbreak \_number & 69.70 & 67.90 & 74.40 & 57.80 & 90.50 & 90.70 & 74.20 & 78.50 & 82.80 & 98.00 \\

    & anaphor\allowbreak \_agreement\allowbreak \_gender & 69.80 & 63.80 & 71.80 & 73.60 & 95.10 & 96.20 & 22.50 & 33.50 & 61.40 & 98.00 \\

\midrule

\multirow{3}{\phenomenon\linewidth}{\textsc{Noun Phrase \mbox{Agreement}}}
    & np\allowbreak \_agreement\allowbreak \_number & 75.00 & 78.80 & 83.80 & 56.80 & 92.30 & 93.00 & 91.80 & 94.50 & 94.80 & 100.0 \\
    & np\allowbreak \_agreement\allowbreak \_gender & 76.40 & 76.00 & 80.10 & 55.80 & 88.40 & 90.60 & 81.10 & 84.20 & 87.80 & 98.00 \\
    & np\allowbreak \_agreement\allowbreak \_case & 85.90 & 88.50 & 90.00 & 72.10 & 97.40 & 97.50 & 94.80 & 95.90 & 97.00 & 98.00 \\
\midrule

\multirow{3}{\phenomenon\linewidth}{\textsc{Floating Quant. \mbox{Agreement}}}
    & floating\allowbreak \_quantifier\allowbreak \_agreement\allowbreak \_number & 79.90 & 77.70 & 80.80 & 47.10 & 85.10 & 87.60 & 87.00 & 89.80 & 90.90 & 100.0 \\
    & floating\allowbreak \_quantifier\allowbreak \_agreement\allowbreak \_gender & 48.80 & 52.80 & 64.50 & 50.50 & 94.70 & 94.00 & 54.30 & 62.00 & 90.30 & 98.00 \\
    & floating\allowbreak \_quantifier\allowbreak \_agreement\allowbreak \_case & 71.90 & 74.50 & 80.80 & 44.10 & 94.10 & 91.50 & 75.00 & 79.10 & 82.30 & 100.0 \\
\midrule

\multirow{1}{\phenomenon\linewidth}{\textsc{Reflexives}}
    & external\allowbreak \_posessor & 13.10 & 19.30 & 23.30 & 64.10 & 82.90 & 90.80 & 84.80 & 89.00 & 32.40 & 98.00 \\
\midrule

\multirow{3}{\phenomenon\linewidth}{\textsc{Negation}}
    & negative\allowbreak \_concord & 98.30 & 98.80 & 99.00 & 75.50 & 100.00 & 100.00 & 99.50 & 99.60 & 99.80 & 100.0 \\
    & negative\allowbreak \_pronoun\allowbreak \_to\allowbreak \_indefinite & 12.40 & 16.10 & 11.70 & 2.60 & 21.90 & 23.30 & 20.80 & 25.20 & 19.60 & 100.0 \\
    & indefinite\allowbreak \_pronoun\allowbreak \_to\allowbreak \_negative & 84.90 & 88.60 & 94.50 & 85.80 & 100.00 & 100.00 & 96.90 & 97.20 & 97.80 & 100.0 \\
    \midrule

\multirow{5}{\phenomenon\linewidth}{\textsc{Argument \mbox{Structure}}}
    & transitive\allowbreak \_verb & 74.80 & 74.80 & 76.20 & 28.10 & 83.50 & 83.00 & 81.70 & 85.90 & 87.20 & 100.0 \\
    & transitive\allowbreak \_verb\allowbreak \_subject & 56.30 & 56.70 & 58.40 & 33.30 & 71.40 & 72.60 & 70.10 & 73.40 & 73.70 & 100.0 \\
    & transitive\allowbreak \_verb\allowbreak \_passive & 56.30 & 54.30 & 60.10 & 54.10 & 89.90 & 91.60 & 90.30 & 92.40 & 91.10 & 100.0 \\
    & transitive\allowbreak \_verb\allowbreak \_object & 34.80 & 38.90 & 43.50 & 38.60 & 88.00 & 87.20 & 82.30 & 83.90 & 83.20 & 100.0 \\
    & transitive\allowbreak \_verb\allowbreak \_iobject & 46.70 & 46.40 & 50.00 & 36.90 & 80.30 & 81.20 & 75.40 & 79.80 & 81.20 & 100.0 \\
    \midrule

\multirow{3}{\phenomenon\linewidth}{\textsc{Aspect}}
    & change\allowbreak \_duration\allowbreak \_aspect & 50.00 & 48.20 & 53.20 & 74.10 & 91.00 & 90.40 & 81.40 & 87.00 & 84.60 & 100.0 \\
    & change\allowbreak \_repetition\allowbreak \_aspect & 57.00 & 57.40 & 61.30 & 74.90 & 90.60 & 91.50 & 86.40 & 91.50 & 91.10 & 100.0 \\
    & deontic\allowbreak \_imperative\allowbreak \_aspect & 57.00 & 47.80 & 50.70 & 76.80 & 89.80 & 88.90 & 75.80 & 78.00 & 84.10 & 100.0 \\
    \midrule

\multirow{3}{\phenomenon\linewidth}{\textsc{Tense}}
    & single\allowbreak \_verb\allowbreak \_tense & 75.50 & 76.60 & 85.80 & 49.70 & 71.40 & 75.00 & 78.90 & 84.40 & 84.80 & 100.0 \\
    & conj\allowbreak \_verb\allowbreak \_tense & 73.50 & 75.60 & 83.70 & 50.60 & 87.20 & 88.50 & 86.90 & 91.30 & 92.40 & 100.0 \\
    & tense\allowbreak \_marker & 57.60 & 64.70 & 62.80 & 55.50 & 83.10 & 83.60 & 80.00 & 81.20 & 87.20 & 98.00 \\
\midrule
\multicolumn{2}{r}{\textbf{Average}} & 71.85 & 73.20 & 76.20 & 50.05 & 86.04 & 86.91 & 83.08 & 86.03 & 87.04 & 99.15 \\
\bottomrule
\end{tabular}
}
\caption{Accuracy scores (\%) for the monolingual LMs by paradigm (part 2). Random baseline is 50\%.  }
\label{tab:rublimp_phenomena_multi_p2}
\end{table*}

%% file: tables/cross_ling_all.tex
\begin{table*}[!ht]
    \centering
    \resizebox{0.97\textwidth}{!}{\begin{tabular}{lRRRRRRRRRRR}
        \toprule
       \multirow{2}{*}{\textbf{Model}}  
       & \multicolumn{1}{c}{\textbf{\rublimp}}
       & \multicolumn{1}{c}{\textbf{\blimp}}
       & \multicolumn{1}{c}{\textbf{\climp}}
       & \multicolumn{1}{c}{\textbf{\sling}}
       & \multicolumn{1}{c}{\textbf{\jblimp}}
       & \multicolumn{5}{c}{\textbf{\clams}}

       & \multicolumn{1}{c}{\multirow{2}{*}{\textbf{Avg.}}} \\
    
     \cmidrule(lr){2-2} \cmidrule(lr){3-3} \cmidrule(lr){4-4} \cmidrule(lr){5-5} \cmidrule(lr){6-6} \cmidrule(lr){7-11}
     
        & \multicolumn{1}{c}{ru} & \multicolumn{1}{c}{en} & \multicolumn{1}{c}{zh} & \multicolumn{1}{c}{zh}  & \multicolumn{1}{c}{ja} & \multicolumn{1}{c}{en} & \multicolumn{1}{c}{ru} & \multicolumn{1}{c}{de} & \multicolumn{1}{c}{fr} & \multicolumn{1}{c}{he}  \\
       \midrule
       

       \dmbert & 70.65 & 66.49 & 69.07 & 75.25 & 59.72 & 69.42 & 73.52 & 84.50 & 75.04 & 65.12 & 70.88 \\

       \mbert & 75.03 & 68.45 & 72.95 & 74.01 & 65.49 & 66.95 & 83.33 & 84.09 & 75.96 & 66.57 & 73.28 \\

       \xlmrb & 84.81 & 76.69 & 72.84 & 73.63 & 66.28 & 73.25 & 78.02 & 86.40 & 70.16 & 68.66 & 75.07 \\

       \xlmrl & 87.46 & 79.60 & 74.85 & 73.72 & 71.96 & 82.09 & 78.17 & 87.62 & 72.53 & 75.66 & 78.37 \\

       \rembert & 50.55 & 45.42 & 51.61 & 48.34 & 43.57 & 47.38 & 51.51 & 40.28 & 46.48 & 49.05 & 47.42 \\

       \mdeberta & 44.22 & 54.02 & 49.45 & 48.72 & 44.84 & 54.28 & 60.72 & 56.45 & 64.53 & 52.20 & 52.94 \\

       \midrule
       
        \mgpt & 86.37 & 76.21 & 75.85 & 68.80 & 71.86 & 73.31 & 76.31 & 91.40 & 75.76 & 66.81 & 76.27 \\
        
        \mgptb & 88.26 & 76.45 & 76.21 & 72.30 & 63.29 & 83.15 & 77.56 & 92.65 & 80.18 & 68.72 & 77.88 \\

        \blooms & 71.85 & 78.16 & 73.44 & 66.00 & 49.91 & 79.69 & 73.49 & 65.24 & 82.95 & 53.04 & 69.38 \\

        \bloomm & 73.20 & 78.69 & 74.25 & 64.67 & 60.47 & 81.87 & 76.18 & 67.33 & 83.85 & 58.50 & 71.90 \\

        \blooml & 76.20 & 79.54 & 73.66 & 66.80 & 65.83 & 84.70 & 79.72 & 69.53 & 86.37 & 56.08 & 73.84 \\

        \xglms & 50.05 & 78.56 & 77.01 & 65.18 & 72.87 & 75.94 & 81.95 & 91.87 & 79.24 & 52.60 & 72.53 \\

        \xglmm & 86.04 & 77.94 & 76.07 & 67.36 & 71.06 & 75.18 & 83.53 & 91.75 & 81.60 & 70.58 & 78.11 \\

        \xglml & 86.91 & 78.99 & 77.83 & 66.49 & 73.77 & 76.57 & 83.72 & 93.22 & 81.81 & 52.00 & 77.13 \\

        \llamas & 83.08 & 79.46 & 63.89 & 74.75 & 70.36 & 78.14 & 80.73 & 89.52 & 84.39 & 53.96 & 75.83 \\
        \llamam & 86.03 & 79.11 & 64.53 & 75.32 & 69.20 & 78.79 & 82.62 & 87.19 & 82.59 & 54.08 & 75.95 \\
        \mistral & 87.04 & 80.66 & 72.03 & 79.51 & 69.15 & 86.01 & 87.04 & 84.01 & 82.91 & 56.78 & 78.51 \\

        \bottomrule
       
    \end{tabular}}
    \caption{Accuracy scores (\%) for the multilingual experiments on RuBLiMP, BLiMP, CLiMP, SLING, JBLiMP, and CLAMS. Random baseline is 50\%. The line separates the encoder-only and decoder-only LMs.}
    \label{tab:crossling_overall}
\end{table*}

%% file: parts/appendix_multilingual_experiments.tex
\section{Multilingual Experiments}
\label{appendix:multi}

\subsection{Experimental Setup}\label{sec: app_benchmarks_description} We evaluate 17 multilingual LMs on six benchmarks as shown in \S\ref{sec:exps}. The benchmarks can be characterized by the minimal pair generation method: (i) using a dictionary and linguistic templates (BLiMP), (ii) translating an English dictionary and adapting the linguistic templates (CLiMP, CLAMS), (iii) collecting examples from linguistic publications (JBLiMP), (iv) extracting sentences from a Universal Dependencies treebank and using linguistic templates (SLING), and (v) extracting sentences from open text corpora, using linguistic perturbations, and decontaminating test data (ours). The benchmark details are given below:

\begin{itemize}[leftmargin=*,topsep=1pt]
    \setlength\itemsep{0pt}
    
    \item \blimp \cite{warstadt-etal-2020-blimp-benchmark} comprises 67 paradigms for English, 1k minimal pairs each. It covers 12 representative phenomena in English, including anaphor agreement, argument structure, binding, control/raising, determiner-noun agreement, ellipsis, filler gap dependencies, irregular verb forms, island effects, NPI licensing, quantifiers, and subject-verb agreement.
    
    \item \climp \cite{xiang-etal-2021-climp} includes 16 paradigms nine phenomena in Chinese, such as anaphor agreement, binding, argument structure, and classifier-noun agreement.

    \item \sling \cite{song-etal-2022-sling} includes nine high-level linguistic phenomena in Mandarin Chinese, present in \climp (e.g., anaphor agreement, classifier-noun agreement, binding) and new ones (aspect, polarity items, relative clauses, and wh-fronting, among others).

    \item \jblimp \cite{someya-oseki-2023-jblimp} comprises 11 phenomena in Japanese: argument structure, binding, control/raising, ellipsis, filler gap dependencies, island effects, morphology, nominal structures, NPI licensing, verbal agreement, and quantifiers.
    
    \item \clams \cite{mueller-etal-2020-cross} is a syntactic evaluation suite in five languages (English, Russian, French, German, and Hebrew) that covers different paradigms of subject-verb agreement.
    
\end{itemize}

\subsection{Results}
The results are summarized in \autoref{tab:crossling_overall}. Overall, we find that \rembert and \mdeberta perform at the level of a random baseline on all benchmarks. We also observe an unsatisfactory performance of most decoder-only LMs on CLAMS (Hebrew) and JBLiMP (Japanese), with the scores ranging between approx. 50\% (\xglms) to 70.6\% (\xglmm). No single LM performs consistently well in all languages.

\paragraph{Larger $\neq$ Better} Similar to our findings on \rublimp (\S\ref{sec:monolingual}), the LMs' performance does not always improve with the number of parameters, e.g.: \texttt{XLMR} (BLiMP, SLING, CLAMS), \texttt{mGPT} (BLiMP, CLiMP, SLING), and \texttt{bloom} (CLiMP and SLING).

\paragraph{Sensitivity to Agreement} For a more fine-grained analysis, we select \textsc{Agreement} as one of the most well-represented phenomena in all considered benchmarks. We report the results in \autoref{tab:crossling_agr} and describe them by phenomenon and language. The general trend here is that model performance in a given language depends on the benchmark. In particular, the $\Delta$-scores between the benchmarks for the \textsc{Subject-predicate Agreement} in Russian can range from 2.4\% (\texttt{distil-MBERT}) to 37\% (\xglms). However, some LMS perform consistently w.r.t. this phenomenon on both RuBLiMP and BLiMP (e.g., \texttt{bloom}, \texttt{xglm}, \texttt{MBERT}). The LMs identify the \textsc{Anaphor Agreement} contrast reliably on BLiMP and demonstrate lower performance on CLAMS, with the 
$\Delta$-score in the range between 2.72\% and 15.03\% For Chinese, the $\Delta$-scores vary between 4\% and 21\%. We assume that the result differences are attributed to the minimal pair generation method and quality, which is analyzed in detail for SLING and CLiMP \cite{song-etal-2022-sling}. We provide the results of the CLAMS' manual analysis below.

\vspace{0.6em}

\noindent Now, we focus on the performance analysis for Chinese and Russian since both languages have benchmarks created through the translation of an English vocabulary and linguistic templates (CLiMP and CLAMS) and usage of open text corpora, linguistic resources, and linguistic perturbations (SLING and RuBLiMP). 

\paragraph{CLiMP vs SLING} We find that the decoder-only LMs generally perform worse on SLING, with the accuracy $\Delta$-score of up to 12\% (e.g., \texttt{xglm} and \texttt{bloom}). A high-level analysis indicates that SLING does overcome the limitations of CLiMP and represents a more challenging benchmark of linguistic minimal pairs for Chinese. We refer the reader to \citet{song-etal-2022-sling} for a detailed comparison of these two evaluation resources. 

\paragraph{CLAMS vs RuBLiMP} We are interested in analyzing the LMs' performance differences on CLAMS and RuBLiMP in more detail. Three authors of this paper conduct a manual analysis of 50 random examples in CLAMS (approx. 17 examples per author) and the paper's appendices \cite{mueller-etal-2020-cross}. The results show there are:

\begin{enumerate}[leftmargin=*,topsep=1pt]
    \setlength\itemsep{0pt}
    \item 60\% of plausible minimal pairs; the minimum length is 2 tokens (e.g., \textit{Khudozhnik stariy/$^{*}$stariye} ``The painter is/$^{*}$are old'').
    \item 20\% of semantically implausible or uninterpretable pairs (e.g., \textit{Vrachi, kotorykh lidery hotyat/$^{*}$hochet, bol'schiye} ``The doctors that the leaders want/$^{*}$wants are big'').
    \item 15\% of pairs do not isolate a target phenomenon, which means that the grammatical sentence is implausible or the ungrammatical sentence can have multiple errors. E.g., \textit{Klienty govoryat i zhdali/$^{*}$zhdal} ``The clients are speaking and were/$^{*}$was waiting''. Here, the tense concord rules are violated in the grammatical sentence, which leads to the perturbation of both number and tense verb forms in the ungrammatical sentence).
    \item 5\% of pairs contain repetitive constructions or abruptly break off (e.g.,  \textit{Senator lyubit smotret' teleperedachi and lyubit/$^{*}$lyubyat smotret' teleperedachi} ``The senator likes to watch TV and likes/$^{*}$like to watch TV'').
\end{enumerate}

The primary reason behind these errors is that the word vocabulary is translated from English, and the contextual ambiguity is not controlled. There are 126 unique tokens (including the punctuation marks) in the 40.1k grammatical sentences in CLAMS, which significantly limits the diversity of the minimal pairs. Besides, some minimal pairs are plausible from the perspective of well-formedness and acceptability. However, a native Russian speaker -- at least the authors performing the analysis -- is unlikely to say or write a sentence this way. We conclude that these factors contribute to the performance differences.

\input{tables/agreement_cross_ling}

%% file: tables/agreement_cross_ling.tex
\begin{table*}[ht!]
    \centering
    \resizebox{\textwidth}{!}{
    \begin{tabular}{lRRRRRRRRRRRRRR}
        \toprule
        \multirow{3}{*}{\textbf{Model}} 
        & \multicolumn{3}{c}{\textbf{\rublimp}} 
        & \multicolumn{3}{c}{\textbf{\blimp}} 
        & \multicolumn{1}{c}{\textbf{\climp}} 
        & \multicolumn{1}{c}{\textbf{\sling}} 
        & \multicolumn{6}{c}{\textbf{\clams}}
        \\
    \cmidrule(lr){2-4} \cmidrule(lr){5-7} \cmidrule(lr){8-8} \cmidrule(lr){9-9} \cmidrule(lr){10-15}
        
        & \multicolumn{3}{c}{ru} 
        & \multicolumn{3}{c}{en} 
        & \multicolumn{1}{c}{zh} 
        & \multicolumn{1}{c}{zh} 
        & \multicolumn{2}{c}{en} 
        & \multicolumn{1}{c}{ru} 
        & \multicolumn{1}{c}{de}
        & \multicolumn{1}{c}{fr}
        & \multicolumn{1}{c}{he}\\

    \cmidrule(lr){2-4} \cmidrule(lr){5-7} \cmidrule(lr){8-8} \cmidrule(lr){9-9} \cmidrule(lr){10-11} \cmidrule(lr){12-12} \cmidrule(lr){13-13} \cmidrule(lr){14-14} \cmidrule(lr){15-15}
        & \multicolumn{1}{c}{SPA} 
        & \multicolumn{1}{c}{AA} 
        & \multicolumn{1}{c}{NPA}
        & \multicolumn{1}{c}{SPA}
        & \multicolumn{1}{c}{AA}
        & \multicolumn{1}{c}{DNA}
        & \multicolumn{1}{c}{AA}
        & \multicolumn{1}{c}{AA}
        & \multicolumn{1}{c}{SPA}
        & \multicolumn{1}{c}{AA} 
        & \multicolumn{1}{c}{SPA}
        & \multicolumn{1}{c}{SPA}
        & \multicolumn{1}{c}{SPA}
        & \multicolumn{1}{c}{SPA} \\
        
        \midrule

       \dmbert & 75.90 & 52.35 & 79.43 & 75.30 & 94.70 & 87.11 & 83.40 & 83.37 & 67.83 & 85.26 & 73.52 & 84.50 & 75.04 & 65.12\\

       \mbert & 80.37 & 86.35 & 87.07 & 80.28 & 89.75 & 88.45 & 73.00 & 89.11 & 65.70	& 79.54 & 83.33 & 84.09 & 75.96 & 66.57\\

       \xlmrb &  87.55 & 91.85 & 92.67 & 80.08 & 92.00 & 90.46 & 82.10 & 75.35 & 72.55 & 80.32 & 78.02 & 86.40 & 70.16 & 68.66\\

       \xlmrl & 87.73 & 93.15 & 94.37 & 83.42 & 94.95 & 92.98 & 78.00 & 67.17 & 81.48 & 88.22 & 78.17 & 87.62 & 72.53 & 75.66\\

       \rembert & 49.77 & 32.05 & 51.17 & 50.18 & 49.30 & 45.01 & 52.90 & 36.95 & 47.46 & 46.58 & 51.51 & 40.28 & 46.48 & 49.05\\

       \mdeberta & 36.77 & 75.35 & 41.03 & 53.79 & 45.45 & 48.54 & 72.00 & 36.31 & 54.67 & 50.32 & 60.72 & 56.45 & 64.53 & 52.20 \\

       \midrule
       
        \mgpt &  87.69 & 92.15 & 94.2 & 79.20 & 98.00 & 88.80 & 86.80 & 65.67 & 71.97 & 86.66 & 76.31 & 91.40 & 75.76 & 66.81\\

        \mgptb &  88.75 & 94.35 & 95.33 & 73.55 & 98.95 & 87.80 & 86.90 & 70.29 & 82.71 & 87.56 & 77.56 & 92.65 & 80.18 & 68.72 \\

        \blooms & 85.55 & 69.75 & 79.10 & 85.50 & 98.40 & 93.20 & 61.80 & 57.77 & 78.58 & 90.76 & 73.49 & 65.24 & 82.95 & 53.40\\

        \bloomm & 86.66 & 65.85 & 81.10 & 84.80 & 98.80 & 91.90 & 62.40 & 56.27 & 81.02 & 90.40 & 76.18 & 67.33 & 83.85 & 58.50 \\

        \blooml & 88.90 & 73.1 & 84.63 & 85.80 & 99.30 & 93.50 & 62.10 & 60.07 & 84.03 & 91.41 & 79.72 & 69.53 & 86.37 & 56.08\\

        \xglms & 44.26 & 65.7 & 61.57 & 84.50 & 99.60 & 89.90 & 77.60 & 58.72 & 75.08 & 84.57 & 81.95 & 91.87 & 79.24 & 52.60 \\

        \xglmm & 82.70 & 92.8  & 92.70 & 84.30 & 99.10 & 89.80 & 78.30 & 60.69 & 74.19 & 85.11 & 83.53 & 91.75 & 81.60 & 70.58 \\

        \xglml & 83.75 & 93.45 & 93.70 & 83.90 & 99.50 & 90.50 & 81.70 & 59.54 & 75.75 & 84.8 & 83.72 & 93.22 & 81.81 & 52.00 \\

        \llamas & 89.45 & 48.35 & 89.23 & 74.50 & 99.45 & 91.15 & 63.30 & 79.48 & 76.52 & 94.40 & 80.73 & 89.52 & 84.39 & 53.96 \\
        \llamam & 91.23 & 56.00 & 91.53 & 78.20 & 99.50 & 90.32 & 64.60 & 79.09 & 77.65 & 90.19 & 82.62 & 87.19 & 82.59 & 54.08 \\
        \mistral & 92.99 & 72.10 & 93.20 & 76.60 & 99.55 & 91.39 & 91.00 & 86.87 & 85.47 & 91.41 & 87.04 & 84.01 & 82.91 & 56.78 \\

        \bottomrule
    \end{tabular}}
    \caption{Results of the multilingual model evaluation on the agreement phenomena. \textbf{Phenomena}: SPA -- Subject-Predicate agreement, AA -- Anaphor Agreement, NPA -- Noun-Phrase Agreement, DNA -- Determiner-Noun Agreement.}
    \label{tab:crossling_agr}
\end{table*}

%% file: main.bbl
\begin{thebibliography}{76}
\providecommand{\natexlab}[1]{#1}

\bibitem[{Anastasyev(2020)}]{anastasyev2020exploring}
D.~G. Anastasyev. 2020.
\newblock \href {http://dx.doi.org/10.28995/2075-7182-2020-19-1-12} {Exploring
  pretrained models for joint morpho-syntactic parsing of {R}ussian}.
\newblock In \emph{Computational Linguistics and Intellectual Technologies},
  pages 1--12.

\bibitem[{Arylova(2013)}]{arylova2013possession}
Aysa Arylova. 2013.
\newblock \emph{Possession in the Russian clause: towards dynamicity in
  syntax}.
\newblock Ph.D. thesis, University of Groningen.
\newblock Relation: http://www.rug.nl/ Rights: University of Groningen.

\bibitem[{Attardi(2015)}]{Wikiextractor2015}
Giusepppe Attardi. 2015.
\newblock Wikiextractor.
\newblock \url{https://github.com/attardi/wikiextractor}.

\bibitem[{Batra et~al.(2021)Batra, Jain, Heidari, Arun, Youngs, Li, Donmez,
  Mei, Kuo, Bhardwaj, Kumar, and White}]{batra-etal-2021-building}
Soumya Batra, Shashank Jain, Peyman Heidari, Ankit Arun, Catharine Youngs,
  Xintong Li, Pinar Donmez, Shawn Mei, Shiunzu Kuo, Vikas Bhardwaj, Anuj Kumar,
  and Michael White. 2021.
\newblock \href {https://doi.org/10.18653/v1/2021.emnlp-main.53} {Building
  adaptive acceptability classifiers for neural {NLG}}.
\newblock In \emph{Proceedings of the 2021 Conference on Empirical Methods in
  Natural Language Processing}, pages 682--697, Online and Punta Cana,
  Dominican Republic. Association for Computational Linguistics.

\bibitem[{Bocharov et~al.(2013)Bocharov, Alexeeva, Granovsky, Protopopova,
  Stepanova, and Surikov}]{bocharov2013crowdsourcing}
Victor Bocharov, Svetlana Alexeeva, Dmitry Granovsky, Ekaterina Protopopova,
  Maria Stepanova, and Aleksei Surikov. 2013.
\newblock Crowdsourcing morphological annotation.
\newblock In \emph{{Computational Linguistics and Intellectual Technologies:
  Papers from the Annual International Conference Dialogue}}, volume~1, pages
  109--114.

\bibitem[{Bolshakova and Sapin(2021)}]{Bolshakova2021}
Elena Bolshakova and Andrey Sapin. 2021.
\newblock \href {https://doi.org/10.28995/2075-7182-2021-20-154-161} {Building
  dataset and morpheme segmentation model for russian word forms}.
\newblock In \emph{Computational Linguistics and Intellectual Technologies:
  Proceedings of the International Conference “Dialogue”}, pages 154--161.

\bibitem[{Bowman and Dahl(2021)}]{bowman-dahl-2021-will}
Samuel~R. Bowman and George Dahl. 2021.
\newblock \href {https://doi.org/10.18653/v1/2021.naacl-main.385} {What will it
  take to fix benchmarking in natural language understanding?}
\newblock In \emph{Proceedings of the 2021 Conference of the North American
  Chapter of the Association for Computational Linguistics: Human Language
  Technologies}, pages 4843--4855, Online. Association for Computational
  Linguistics.

\bibitem[{Brown et~al.(2020)Brown, Mann, Ryder, Subbiah, Kaplan, Dhariwal,
  Neelakantan, Shyam, Sastry, Askell et~al.}]{brown2020language}
Tom Brown, Benjamin Mann, Nick Ryder, Melanie Subbiah, Jared~D Kaplan, Prafulla
  Dhariwal, Arvind Neelakantan, Pranav Shyam, Girish Sastry, Amanda Askell,
  et~al. 2020.
\newblock {Language Models are Few-shot Learners}.
\newblock \emph{{Advances in Neural Information Processing Systems}},
  33:1877--1901.

\bibitem[{Choenni and Shutova(2022)}]{choenni-shutova-2022-investigating}
Rochelle Choenni and Ekaterina Shutova. 2022.
\newblock \href {https://doi.org/10.1162/coli_a_00444} {Investigating language
  relationships in multilingual sentence encoders through the lens of
  linguistic typology}.
\newblock \emph{Computational Linguistics}, 48(3):635--672.

\bibitem[{Chomsky(1965)}]{chomsky1965aspects}
Noam Chomsky. 1965.
\newblock \emph{Aspects of the Theory of Syntax}.
\newblock MIT Press.

\bibitem[{Chung et~al.(2021)Chung, F{\'{e}}vry, Tsai, Johnson, and
  Ruder}]{Chung2021Rethinking}
Hyung~Won Chung, Thibault F{\'{e}}vry, Henry Tsai, Melvin Johnson, and
  Sebastian Ruder. 2021.
\newblock \href {https://openreview.net/forum?id=xpFFI\_NtgpW} {Rethinking
  embedding coupling in pre-trained language models}.
\newblock In \emph{9th International Conference on Learning Representations,
  {ICLR} 2021, Virtual Event, Austria, May 3-7, 2021}. OpenReview.net.

\bibitem[{Conneau et~al.(2020)Conneau, Khandelwal, Goyal, Chaudhary, Wenzek,
  Guzm{\'a}n, Grave, Ott, Zettlemoyer, and
  Stoyanov}]{conneau-etal-2020-unsupervised}
Alexis Conneau, Kartikay Khandelwal, Naman Goyal, Vishrav Chaudhary, Guillaume
  Wenzek, Francisco Guzm{\'a}n, Edouard Grave, Myle Ott, Luke Zettlemoyer, and
  Veselin Stoyanov. 2020.
\newblock \href {https://doi.org/10.18653/v1/2020.acl-main.747} {Unsupervised
  cross-lingual representation learning at scale}.
\newblock In \emph{Proceedings of the 58th Annual Meeting of the Association
  for Computational Linguistics}, pages 8440--8451, Online. Association for
  Computational Linguistics.

\bibitem[{Csaki et~al.(2023)Csaki, Pawakapan, Thakker, and
  Xu}]{csaki2023efficiently}
Zoltan Csaki, Pian Pawakapan, Urmish Thakker, and Qiantong Xu. 2023.
\newblock {Efficiently Adapting Pre-trained Language Models to New Languages}.
\newblock \emph{arXiv preprint arXiv:2311.05741}.

\bibitem[{Dawid and Skene(1979)}]{dawid_skene}
A.~P. Dawid and A.~M. Skene. 1979.
\newblock \href {http://www.jstor.org/stable/2346806} {Maximum likelihood
  estimation of observer error-rates using the em algorithm}.
\newblock \emph{Journal of the Royal Statistical Society. Series C (Applied
  Statistics)}, 28(1):20--28.

\bibitem[{de~Haan(2002)}]{DeHaan2002}
Ferdinand de~Haan. 2002.
\newblock \href {https://doi.org/10.1075/scl.9.07haa} {Strong modality and
  negation in russian}.
\newblock In Randi Reppen, Susan Fitzmaurice, and Douglas Biber, editors,
  \emph{Using Corpora to Explore Linguistic Variation}, pages 91--110. John
  Benjamins, Amsterdam/Philadelphia.

\bibitem[{Devlin et~al.(2019)Devlin, Chang, Lee, and
  Toutanova}]{devlin-etal-2019-bert}
Jacob Devlin, Ming-Wei Chang, Kenton Lee, and Kristina Toutanova. 2019.
\newblock \href {https://doi.org/10.18653/v1/N19-1423} {{BERT}: Pre-training of
  deep bidirectional transformers for language understanding}.
\newblock In \emph{Proceedings of the 2019 Conference of the North {A}merican
  Chapter of the Association for Computational Linguistics: Human Language
  Technologies, Volume 1 (Long and Short Papers)}, pages 4171--4186,
  Minneapolis, Minnesota. Association for Computational Linguistics.

\bibitem[{Gao et~al.(2023)Gao, Tow, Abbasi, Biderman, Black, DiPofi, Foster,
  Golding, Hsu, Le~Noac'h, Li, McDonell, Muennighoff, Ociepa, Phang, Reynolds,
  Schoelkopf, Skowron, Sutawika, Tang, Thite, Wang, Wang, and
  Zou}]{eval-harness}
Leo Gao, Jonathan Tow, Baber Abbasi, Stella Biderman, Sid Black, Anthony
  DiPofi, Charles Foster, Laurence Golding, Jeffrey Hsu, Alain Le~Noac'h,
  Haonan Li, Kyle McDonell, Niklas Muennighoff, Chris Ociepa, Jason Phang,
  Laria Reynolds, Hailey Schoelkopf, Aviya Skowron, Lintang Sutawika, Eric
  Tang, Anish Thite, Ben Wang, Kevin Wang, and Andy Zou. 2023.
\newblock \href {https://doi.org/10.5281/zenodo.10256836} {A framework for
  few-shot language model evaluation}.

\bibitem[{Greenberg(1963)}]{Greenberg_63}
Joseph~H. Greenberg. 1963.
\newblock Some universals of grammar with particular reference to the order of
  meaningful elements.
\newblock In Joseph~H. Greenberg, editor, \emph{Universals of Language},
  chapter~5, pages 58--90. The MIT Press.

\bibitem[{Hartmann et~al.(2021)Hartmann, de~Lhoneux, Hershcovich,
  Kementchedjhieva, Nielsen, Qiu, and
  S{\o}gaard}]{hartmann-etal-2021-multilingual}
Mareike Hartmann, Miryam de~Lhoneux, Daniel Hershcovich, Yova Kementchedjhieva,
  Lukas Nielsen, Chen Qiu, and Anders S{\o}gaard. 2021.
\newblock \href {https://doi.org/10.18653/v1/2021.conll-1.19} {A multilingual
  benchmark for probing negation-awareness with minimal pairs}.
\newblock In \emph{Proceedings of the 25th Conference on Computational Natural
  Language Learning}, pages 244--257, Online. Association for Computational
  Linguistics.

\bibitem[{He et~al.(2022)He, Gao, and Chen}]{he2021debertav3}
Pengcheng He, Jianfeng Gao, and Weizhu Chen. 2022.
\newblock {DeBERTaV3: Improving DeBERTa using ELECTRA-Style Pre-Training with
  Gradient-Disentangled Embedding Sharing}.
\newblock In \emph{The Eleventh International Conference on Learning
  Representations}.

\bibitem[{Hopper and Thompson(1980)}]{hopper1980transitivity}
Paul~J Hopper and Sandra~A Thompson. 1980.
\newblock Transitivity in grammar and discourse.
\newblock \emph{Language}, pages 251--299.

\bibitem[{Jentoft and Samuel(2023)}]{jentoft-samuel-2023-nocola}
Matias Jentoft and David Samuel. 2023.
\newblock \href {https://aclanthology.org/2023.nodalida-1.60} {{N}o{C}o{LA}:
  The {N}orwegian corpus of linguistic acceptability}.
\newblock In \emph{Proceedings of the 24th Nordic Conference on Computational
  Linguistics (NoDaLiDa)}, pages 610--617, T{\'o}rshavn, Faroe Islands.
  University of Tartu Library.

\bibitem[{Jiang et~al.(2023)Jiang, Sablayrolles, Mensch, Bamford, Chaplot,
  Casas, Bressand, Lengyel, Lample, Saulnier et~al.}]{jiang2023mistral}
Albert~Q Jiang, Alexandre Sablayrolles, Arthur Mensch, Chris Bamford,
  Devendra~Singh Chaplot, Diego de~las Casas, Florian Bressand, Gianna Lengyel,
  Guillaume Lample, Lucile Saulnier, et~al. 2023.
\newblock Mistral 7b.
\newblock \emph{arXiv preprint arXiv:2310.06825}.

\bibitem[{Korobov(2015)}]{pymorphy2}
Mikhail Korobov. 2015.
\newblock \href {https://doi.org/10.1007/978-3-319-26123-2_31} {Morphological
  analyzer and generator for {R}ussian and {U}krainian languages}.
\newblock In Mikhail~Yu. Khachay, Natalia Konstantinova, Alexander Panchenko,
  Dmitry~I. Ignatov, and Valeri~G. Labunets, editors, \emph{Analysis of Images,
  Social Networks and Texts}, volume 542 of \emph{Communications in Computer
  and Information Science}, pages 320--332. Springer International Publishing.

\bibitem[{Lau et~al.(2017)Lau, Clark, and Lappin}]{lau2017grammaticality}
Jey~Han Lau, Alexander Clark, and Shalom Lappin. 2017.
\newblock Grammaticality, acceptability, and probability: A probabilistic view
  of linguistic knowledge.
\newblock \emph{Cognitive science}, 41(5):1202--1241.

\bibitem[{Lauren{\c{c}}on et~al.(2022)Lauren{\c{c}}on, Saulnier, Wang, Akiki,
  Villanova~del Moral, Le~Scao, Von~Werra, Mou, Gonz{\'a}lez~Ponferrada, Nguyen
  et~al.}]{laurenccon2022bigscience}
Hugo Lauren{\c{c}}on, Lucile Saulnier, Thomas Wang, Christopher Akiki, Albert
  Villanova~del Moral, Teven Le~Scao, Leandro Von~Werra, Chenghao Mou, Eduardo
  Gonz{\'a}lez~Ponferrada, Huu Nguyen, et~al. 2022.
\newblock The bigscience roots corpus: A 1.6 tb composite multilingual dataset.
\newblock \emph{Advances in Neural Information Processing Systems},
  35:31809--31826.

\bibitem[{Leong et~al.(2023)Leong, Ngui, Susanto, Rengarajan, Sarveswaran, and
  Tjhi}]{leong2023bhasa}
Wei~Qi Leong, Jian~Gang Ngui, Yosephine Susanto, Hamsawardhini Rengarajan,
  Kengatharaiyer Sarveswaran, and William~Chandra Tjhi. 2023.
\newblock \href {https://arxiv.org/abs/2309.06085} {{BHASA: A Holistic
  Southeast Asian Linguistic and Cultural Evaluation Suite for Large Language
  Models}}.
\newblock \emph{Preprint}, arXiv:2309.06085.

\bibitem[{Lhoest et~al.(2021)Lhoest, Villanova~del Moral, Jernite, Thakur, von
  Platen, Patil, Chaumond, Drame, Plu, Tunstall, Davison, {\v{S}}a{\v{s}}ko,
  Chhablani, Malik, Brandeis, Le~Scao, Sanh, Xu, Patry, McMillan-Major, Schmid,
  Gugger, Delangue, Matussi{\`e}re, Debut, Bekman, Cistac, Goehringer, Mustar,
  Lagunas, Rush, and Wolf}]{lhoest-etal-2021-datasets}
Quentin Lhoest, Albert Villanova~del Moral, Yacine Jernite, Abhishek Thakur,
  Patrick von Platen, Suraj Patil, Julien Chaumond, Mariama Drame, Julien Plu,
  Lewis Tunstall, Joe Davison, Mario {\v{S}}a{\v{s}}ko, Gunjan Chhablani,
  Bhavitvya Malik, Simon Brandeis, Teven Le~Scao, Victor Sanh, Canwen Xu,
  Nicolas Patry, Angelina McMillan-Major, Philipp Schmid, Sylvain Gugger,
  Cl{\'e}ment Delangue, Th{\'e}o Matussi{\`e}re, Lysandre Debut, Stas Bekman,
  Pierric Cistac, Thibault Goehringer, Victor Mustar, Fran{\c{c}}ois Lagunas,
  Alexander Rush, and Thomas Wolf. 2021.
\newblock \href {https://doi.org/10.18653/v1/2021.emnlp-demo.21} {Datasets: A
  community library for natural language processing}.
\newblock In \emph{Proceedings of the 2021 Conference on Empirical Methods in
  Natural Language Processing: System Demonstrations}, pages 175--184, Online
  and Punta Cana, Dominican Republic. Association for Computational
  Linguistics.

\bibitem[{Lin et~al.(2022)Lin, Mihaylov, Artetxe, Wang, Chen, Simig, Ott,
  Goyal, Bhosale, Du, Pasunuru, Shleifer, Koura, Chaudhary, O{'}Horo, Wang,
  Zettlemoyer, Kozareva, Diab, Stoyanov, and Li}]{lin-etal-2022-shot}
Xi~Victoria Lin, Todor Mihaylov, Mikel Artetxe, Tianlu Wang, Shuohui Chen,
  Daniel Simig, Myle Ott, Naman Goyal, Shruti Bhosale, Jingfei Du, Ramakanth
  Pasunuru, Sam Shleifer, Punit~Singh Koura, Vishrav Chaudhary, Brian O{'}Horo,
  Jeff Wang, Luke Zettlemoyer, Zornitsa Kozareva, Mona Diab, Veselin Stoyanov,
  and Xian Li. 2022.
\newblock \href {https://doi.org/10.18653/v1/2022.emnlp-main.616} {Few-shot
  learning with multilingual generative language models}.
\newblock In \emph{Proceedings of the 2022 Conference on Empirical Methods in
  Natural Language Processing}, pages 9019--9052, Abu Dhabi, United Arab
  Emirates. Association for Computational Linguistics.

\bibitem[{Linzen and Baroni(2021)}]{linzen2021syntactic}
Tal Linzen and Marco Baroni. 2021.
\newblock Syntactic structure from deep learning.
\newblock \emph{Annu. Rev. Linguist}, 7:2--1.

\bibitem[{Linzen et~al.(2016)Linzen, Dupoux, and
  Goldberg}]{linzen-etal-2016-assessing}
Tal Linzen, Emmanuel Dupoux, and Yoav Goldberg. 2016.
\newblock \href {https://doi.org/10.1162/tacl_a_00115} {Assessing the ability
  of {LSTM}s to learn syntax-sensitive dependencies}.
\newblock \emph{Transactions of the Association for Computational Linguistics},
  4:521--535.

\bibitem[{Lucas et~al.(2023)Lucas, Uchendu, Yamashita, Lee, Rohatgi, and
  Lee}]{lucas-etal-2023-fighting}
Jason Lucas, Adaku Uchendu, Michiharu Yamashita, Jooyoung Lee, Shaurya Rohatgi,
  and Dongwon Lee. 2023.
\newblock \href {https://doi.org/10.18653/v1/2023.emnlp-main.883} {Fighting
  fire with fire: The dual role of {LLM}s in crafting and detecting elusive
  disinformation}.
\newblock In \emph{Proceedings of the 2023 Conference on Empirical Methods in
  Natural Language Processing}, pages 14279--14305, Singapore. Association for
  Computational Linguistics.

\bibitem[{Marvin and Linzen(2018)}]{marvin-linzen-2018-targeted}
Rebecca Marvin and Tal Linzen. 2018.
\newblock \href {https://doi.org/10.18653/v1/D18-1151} {Targeted syntactic
  evaluation of language models}.
\newblock In \emph{Proceedings of the 2018 Conference on Empirical Methods in
  Natural Language Processing}, pages 1192--1202, Brussels, Belgium.
  Association for Computational Linguistics.

\bibitem[{Mikhailov et~al.(2022)Mikhailov, Shamardina, Ryabinin, Pestova,
  Smurov, and Artemova}]{mikhailov-etal-2022-rucola}
Vladislav Mikhailov, Tatiana Shamardina, Max Ryabinin, Alena Pestova, Ivan
  Smurov, and Ekaterina Artemova. 2022.
\newblock \href {https://doi.org/10.18653/v1/2022.emnlp-main.348}
  {{R}u{C}o{LA}: {R}ussian corpus of linguistic acceptability}.
\newblock In \emph{Proceedings of the 2022 Conference on Empirical Methods in
  Natural Language Processing}, pages 5207--5227, Abu Dhabi, United Arab
  Emirates. Association for Computational Linguistics.

\bibitem[{Mikhailov et~al.(2021)Mikhailov, Taktasheva, Sigdel, and
  Artemova}]{mikhailov-etal-2021-rusenteval}
Vladislav Mikhailov, Ekaterina Taktasheva, Elina Sigdel, and Ekaterina
  Artemova. 2021.
\newblock \href {https://aclanthology.org/2021.bsnlp-1.6} {{R}u{S}ent{E}val:
  Linguistic source, encoder force!}
\newblock In \emph{Proceedings of the 8th Workshop on Balto-Slavic Natural
  Language Processing}, pages 43--65, Kiyv, Ukraine. Association for
  Computational Linguistics.

\bibitem[{Mueller et~al.(2020)Mueller, Nicolai, Petrou-Zeniou, Talmina, and
  Linzen}]{mueller-etal-2020-cross}
Aaron Mueller, Garrett Nicolai, Panayiota Petrou-Zeniou, Natalia Talmina, and
  Tal Linzen. 2020.
\newblock \href {https://doi.org/10.18653/v1/2020.acl-main.490}
  {Cross-linguistic syntactic evaluation of word prediction models}.
\newblock In \emph{Proceedings of the 58th Annual Meeting of the Association
  for Computational Linguistics}, pages 5523--5539, Online. Association for
  Computational Linguistics.

\bibitem[{Nangia and Bowman(2019)}]{nangia-bowman-2019-human}
Nikita Nangia and Samuel~R. Bowman. 2019.
\newblock \href {https://doi.org/10.18653/v1/P19-1449} {Human vs. muppet: A
  conservative estimate of human performance on the {GLUE} benchmark}.
\newblock In \emph{Proceedings of the 57th Annual Meeting of the Association
  for Computational Linguistics}, pages 4566--4575, Florence, Italy.
  Association for Computational Linguistics.

\bibitem[{Nguyen et~al.(2024)Nguyen, Nguyen, Lai, Man, Ngo, Dernoncourt, Rossi,
  and Nguyen}]{nguyen-etal-2024-culturax}
Thuat Nguyen, Chien~Van Nguyen, Viet~Dac Lai, Hieu Man, Nghia~Trung Ngo, Franck
  Dernoncourt, Ryan~A. Rossi, and Thien~Huu Nguyen. 2024.
\newblock \href {https://aclanthology.org/2024.lrec-main.377} {{C}ultura{X}: A
  cleaned, enormous, and multilingual dataset for large language models in 167
  languages}.
\newblock In \emph{Proceedings of the 2024 Joint International Conference on
  Computational Linguistics, Language Resources and Evaluation (LREC-COLING
  2024)}, pages 4226--4237, Torino, Italia. ELRA and ICCL.

\bibitem[{Ning et~al.(2018)Ning, Feng, Wu, and Roth}]{ning-etal-2018-joint}
Qiang Ning, Zhili Feng, Hao Wu, and Dan Roth. 2018.
\newblock \href {https://doi.org/10.18653/v1/P18-1212} {Joint reasoning for
  temporal and causal relations}.
\newblock In \emph{Proceedings of the 56th Annual Meeting of the Association
  for Computational Linguistics (Volume 1: Long Papers)}, pages 2278--2288,
  Melbourne, Australia. Association for Computational Linguistics.

\bibitem[{Nivre et~al.(2017)Nivre, Zeman, Ginter, and
  Tyers}]{nivre-etal-2017-universal}
Joakim Nivre, Daniel Zeman, Filip Ginter, and Francis Tyers. 2017.
\newblock \href {https://aclanthology.org/E17-5001} {{U}niversal
  {D}ependencies}.
\newblock In \emph{Proceedings of the 15th Conference of the {E}uropean Chapter
  of the Association for Computational Linguistics: Tutorial Abstracts},
  Valencia, Spain. Association for Computational Linguistics.

\bibitem[{Oren et~al.(2023)Oren, Meister, Chatterji, Ladhak, and
  Hashimoto}]{oren2023proving}
Yonatan Oren, Nicole Meister, Niladri~S Chatterji, Faisal Ladhak, and Tatsunori
  Hashimoto. 2023.
\newblock Proving test set contamination for black-box language models.
\newblock In \emph{The Twelfth International Conference on Learning
  Representations}.

\bibitem[{Paducheva(2010)}]{Paducheva:2010}
Elena Paducheva. 2010.
\newblock \emph{{Semanticheskiye issledovaniya: Semantika vremeni i vida v
  russkom yazyke (in Russian)}}, second edition.
\newblock Languages of Slavonic culture.

\bibitem[{Panchenko et~al.(2017)Panchenko, Ustalov, Arefyev, Paperno,
  Konstantinova, Loukachevitch, and Biemann}]{Panchenko:17:aist}
Alexander Panchenko, Dmitry Ustalov, Nikolay Arefyev, Denis Paperno, Natalia
  Konstantinova, Natalia Loukachevitch, and Chris Biemann. 2017.
\newblock \href {https://doi.org/10.1007/978-3-319-52920-2_21} {\emph{{Human
  and Machine Judgements for Russian Semantic Relatedness}}}, pages 221--235.
\newblock Springer International Publishing, Cham.

\bibitem[{P{\'e}rez-Mayos et~al.(2021)P{\'e}rez-Mayos, T{\'a}boas~Garc{\'\i}a,
  Mille, and Wanner}]{perez-mayos-etal-2021-assessing}
Laura P{\'e}rez-Mayos, Alba T{\'a}boas~Garc{\'\i}a, Simon Mille, and Leo
  Wanner. 2021.
\newblock \href {https://doi.org/10.18653/v1/2021.findings-acl.333} {Assessing
  the syntactic capabilities of transformer-based multilingual language
  models}.
\newblock In \emph{Findings of the Association for Computational Linguistics:
  ACL-IJCNLP 2021}, pages 3799--3812, Online. Association for Computational
  Linguistics.

\bibitem[{Raffel et~al.(2020)Raffel, Shazeer, Roberts, Lee, Narang, Matena,
  Zhou, Li, and Liu}]{raffel2020exploring}
Colin Raffel, Noam Shazeer, Adam Roberts, Katherine Lee, Sharan Narang, Michael
  Matena, Yanqi Zhou, Wei Li, and Peter~J Liu. 2020.
\newblock {Exploring the Limits of Transfer Learning with a Unified
  Text-to-Text Transformer}.
\newblock \emph{Journal of Machine Learning Research}, 21:1--67.

\bibitem[{Ravishankar et~al.(2019)Ravishankar, {\O}vrelid, and
  Velldal}]{ravishankar-etal-2019-probing}
Vinit Ravishankar, Lilja {\O}vrelid, and Erik Velldal. 2019.
\newblock \href {https://doi.org/10.18653/v1/W19-4318} {Probing multilingual
  sentence representations with {X}-probe}.
\newblock In \emph{Proceedings of the 4th Workshop on Representation Learning
  for NLP (RepL4NLP-2019)}, pages 156--168, Florence, Italy. Association for
  Computational Linguistics.

\bibitem[{Reinhart(2016)}]{reinhart2016anaphora}
Tanya Reinhart. 2016.
\newblock \emph{Anaphora and semantic interpretation}.
\newblock Routledge.

\bibitem[{Reynolds(2013)}]{reynolds2013order}
Robert Reynolds. 2013.
\newblock Out of order?: Russian prefixes, complexity-based ordering and
  acyclicity.
\newblock \emph{University of Pennsylvania Working Papers in Linguistics}.

\bibitem[{{\c{S}}ahin et~al.(2020){\c{S}}ahin, Vania, Kuznetsov, and
  Gurevych}]{csahin2020linspector}
G{\"o}zde~G{\"u}l {\c{S}}ahin, Clara Vania, Ilia Kuznetsov, and Iryna Gurevych.
  2020.
\newblock Linspector: Multilingual probing tasks for word representations.
\newblock \emph{Computational Linguistics}, 46(2):335--385.

\bibitem[{Salazar et~al.(2020)Salazar, Liang, Nguyen, and
  Kirchhoff}]{salazar-etal-2020-masked}
Julian Salazar, Davis Liang, Toan~Q. Nguyen, and Katrin Kirchhoff. 2020.
\newblock \href {https://doi.org/10.18653/v1/2020.acl-main.240} {Masked
  language model scoring}.
\newblock In \emph{Proceedings of the 58th Annual Meeting of the Association
  for Computational Linguistics}, pages 2699--2712, Online. Association for
  Computational Linguistics.

\bibitem[{Sanh et~al.(2019)Sanh, Debut, Chaumond, and
  Wolf}]{Sanh2019DistilBERTAD}
Victor Sanh, Lysandre Debut, Julien Chaumond, and Thomas Wolf. 2019.
\newblock {DistilBERT, a Distilled Version of BERT: Smaller, Faster, Cheaper
  and Lighter.}
\newblock In \emph{{Proceedings of Thirty-third Conference on Neural
  Information Processing Systems}}.

\bibitem[{Savchuk et~al.(2024)Savchuk, Arkhangelsky, Bonch-Osmolovskaya,
  Donina, Kuznetsova, Lyashevskaya, Orekhov, and
  Podryadchikova}]{savchuk-ruscorpora-2024}
Svetlana Savchuk, Timofey Arkhangelsky, Anastasia Bonch-Osmolovskaya, Olga
  Donina, Yulia Kuznetsova, Olga Lyashevskaya, Boris Orekhov, and Maria
  Podryadchikova. 2024.
\newblock {Natsionalny Korpus Russkogo Yazyka 2.0: Novye Vozmozhnosti i
  Perspektivy Razvitiya (in Russian)}.
\newblock \emph{Voprosy Yazykoznaniya}, 2:7--34.

\bibitem[{Scao et~al.(2023)Scao, Fan, Akiki, Pavlick, Ilić, Hesslow,
  Castagné, Luccioni, Yvon, Gallé, Tow, Rush, Biderman, Webson, Ammanamanchi,
  Wang, Sagot, Muennighoff, del Moral, Ruwase, Bawden, Bekman, McMillan-Major,
  Beltagy, Nguyen, Saulnier, Tan, Suarez, Sanh, Laurençon, Jernite, Launay,
  Mitchell, Raffel, Gokaslan, Simhi, Soroa, Aji, Alfassy, Rogers, Nitzav, Xu,
  Mou, Emezue, Klamm, Leong, van Strien, Adelani, Radev, Ponferrada, Levkovizh,
  Kim, Natan, Toni, Dupont, Kruszewski, Pistilli, Elsahar, Benyamina, Tran, Yu,
  Abdulmumin, Johnson, Gonzalez-Dios, de~la Rosa, Chim, Dodge, Zhu, Chang,
  Frohberg, Tobing, Bhattacharjee, Almubarak, Chen, Lo, Werra, Weber, Phan,
  allal, Tanguy, Dey, Muñoz, Masoud, Grandury, Šaško, Huang, Coavoux, Singh,
  Jiang, Vu, Jauhar, Ghaleb, Subramani, Kassner, Khamis, Nguyen, Espejel,
  de~Gibert, Villegas, Henderson, Colombo, Amuok, Lhoest, Harliman, Bommasani,
  López, Ribeiro, Osei, Pyysalo, Nagel, Bose, Muhammad, Sharma, Longpre,
  Nikpoor, Silberberg, Pai, Zink, Torrent, Schick, Thrush, Danchev, Nikoulina,
  Laippala, Lepercq, Prabhu, Alyafeai, Talat, Raja, Heinzerling, Si, Taşar,
  Salesky, Mielke, Lee, Sharma, Santilli, Chaffin, Stiegler, Datta, Szczechla,
  Chhablani, Wang, Pandey, Strobelt, Fries, Rozen, Gao, Sutawika, Bari,
  Al-shaibani, Manica, Nayak, Teehan, Albanie, Shen, Ben-David, Bach, Kim,
  Bers, Fevry, Neeraj, Thakker, Raunak, Tang, Yong, Sun, Brody, Uri, Tojarieh,
  Roberts, Chung, Tae, Phang, Press, Li, Narayanan, Bourfoune, Casper, Rasley,
  Ryabinin, Mishra, Zhang, Shoeybi, Peyrounette, Patry, Tazi, Sanseviero, von
  Platen, Cornette, Lavallée, Lacroix, Rajbhandari, Gandhi, Smith, Requena,
  Patil, Dettmers, Baruwa, Singh, Cheveleva, Ligozat, Subramonian, Névéol,
  Lovering, Garrette, Tunuguntla, Reiter, Taktasheva, Voloshina, Bogdanov,
  Winata, Schoelkopf, Kalo, Novikova, Forde, Clive, Kasai, Kawamura, Hazan,
  Carpuat, Clinciu, Kim, Cheng, Serikov, Antverg, van~der Wal, Zhang, Zhang,
  Gehrmann, Mirkin, Pais, Shavrina, Scialom, Yun, Limisiewicz, Rieser,
  Protasov, Mikhailov, Pruksachatkun, Belinkov, Bamberger, Kasner, Rueda,
  Pestana, Feizpour, Khan, Faranak, Santos, Hevia, Unldreaj, Aghagol,
  Abdollahi, Tammour, HajiHosseini, Behroozi, Ajibade, Saxena, Ferrandis,
  McDuff, Contractor, Lansky, David, Kiela, Nguyen, Tan, Baylor, Ozoani, Mirza,
  Ononiwu, Rezanejad, Jones, Bhattacharya, Solaiman, Sedenko, Nejadgholi,
  Passmore, Seltzer, Sanz, Dutra, Samagaio, Elbadri, Mieskes, Gerchick,
  Akinlolu, McKenna, Qiu, Ghauri, Burynok, Abrar, Rajani, Elkott, Fahmy,
  Samuel, An, Kromann, Hao, Alizadeh, Shubber, Wang, Roy, Viguier, Le, Oyebade,
  Le, Yang, Nguyen, Kashyap, Palasciano, Callahan, Shukla, Miranda-Escalada,
  Singh, Beilharz, Wang, Brito, Zhou, Jain, Xu, Fourrier, Periñán, Molano,
  Yu, Manjavacas, Barth, Fuhrimann, Altay, Bayrak, Burns, Vrabec, Bello, Dash,
  Kang, Giorgi, Golde, Posada, Sivaraman, Bulchandani, Liu, Shinzato,
  de~Bykhovetz, Takeuchi, Pàmies, Castillo, Nezhurina, Sänger, Samwald,
  Cullan, Weinberg, Wolf, Mihaljcic, Liu, Freidank, Kang, Seelam, Dahlberg,
  Broad, Muellner, Fung, Haller, Chandrasekhar, Eisenberg, Martin, Canalli, Su,
  Su, Cahyawijaya, Garda, Deshmukh, Mishra, Kiblawi, Ott, Sang-aroonsiri,
  Kumar, Schweter, Bharati, Laud, Gigant, Kainuma, Kusa, Labrak, Bajaj,
  Venkatraman, Xu, Xu, Xu, Tan, Xie, Ye, Bras, Belkada, and
  Wolf}]{workshop2023bloom}
Teven~Le Scao, Angela Fan, Christopher Akiki, Ellie Pavlick, Suzana Ilić,
  Daniel Hesslow, Roman Castagné, Alexandra~Sasha Luccioni, François Yvon,
  Matthias Gallé, Jonathan Tow, Alexander~M. Rush, Stella Biderman, Albert
  Webson, Pawan~Sasanka Ammanamanchi, Thomas Wang, Benoît Sagot, Niklas
  Muennighoff, Albert~Villanova del Moral, Olatunji Ruwase, Rachel Bawden, Stas
  Bekman, Angelina McMillan-Major, Iz~Beltagy, Huu Nguyen, Lucile Saulnier,
  Samson Tan, Pedro~Ortiz Suarez, Victor Sanh, Hugo Laurençon, Yacine Jernite,
  Julien Launay, Margaret Mitchell, Colin Raffel, Aaron Gokaslan, Adi Simhi,
  Aitor Soroa, Alham~Fikri Aji, Amit Alfassy, Anna Rogers, Ariel~Kreisberg
  Nitzav, Canwen Xu, Chenghao Mou, Chris Emezue, Christopher Klamm, Colin
  Leong, Daniel van Strien, David~Ifeoluwa Adelani, Dragomir Radev,
  Eduardo~González Ponferrada, Efrat Levkovizh, Ethan Kim, Eyal~Bar Natan,
  Francesco~De Toni, Gérard Dupont, Germán Kruszewski, Giada Pistilli, Hady
  Elsahar, Hamza Benyamina, Hieu Tran, Ian Yu, Idris Abdulmumin, Isaac Johnson,
  Itziar Gonzalez-Dios, Javier de~la Rosa, Jenny Chim, Jesse Dodge, Jian Zhu,
  Jonathan Chang, Jörg Frohberg, Joseph Tobing, Joydeep Bhattacharjee, Khalid
  Almubarak, Kimbo Chen, Kyle Lo, Leandro~Von Werra, Leon Weber, Long Phan,
  Loubna~Ben allal, Ludovic Tanguy, Manan Dey, Manuel~Romero Muñoz, Maraim
  Masoud, María Grandury, Mario Šaško, Max Huang, Maximin Coavoux, Mayank
  Singh, Mike Tian-Jian Jiang, Minh~Chien Vu, Mohammad~A. Jauhar, Mustafa
  Ghaleb, Nishant Subramani, Nora Kassner, Nurulaqilla Khamis, Olivier Nguyen,
  Omar Espejel, Ona de~Gibert, Paulo Villegas, Peter Henderson, Pierre Colombo,
  Priscilla Amuok, Quentin Lhoest, Rheza Harliman, Rishi Bommasani,
  Roberto~Luis López, Rui Ribeiro, Salomey Osei, Sampo Pyysalo, Sebastian
  Nagel, Shamik Bose, Shamsuddeen~Hassan Muhammad, Shanya Sharma, Shayne
  Longpre, Somaieh Nikpoor, Stanislav Silberberg, Suhas Pai, Sydney Zink,
  Tiago~Timponi Torrent, Timo Schick, Tristan Thrush, Valentin Danchev,
  Vassilina Nikoulina, Veronika Laippala, Violette Lepercq, Vrinda Prabhu, Zaid
  Alyafeai, Zeerak Talat, Arun Raja, Benjamin Heinzerling, Chenglei Si,
  Davut~Emre Taşar, Elizabeth Salesky, Sabrina~J. Mielke, Wilson~Y. Lee,
  Abheesht Sharma, Andrea Santilli, Antoine Chaffin, Arnaud Stiegler, Debajyoti
  Datta, Eliza Szczechla, Gunjan Chhablani, Han Wang, Harshit Pandey, Hendrik
  Strobelt, Jason~Alan Fries, Jos Rozen, Leo Gao, Lintang Sutawika, M~Saiful
  Bari, Maged~S. Al-shaibani, Matteo Manica, Nihal Nayak, Ryan Teehan, Samuel
  Albanie, Sheng Shen, Srulik Ben-David, Stephen~H. Bach, Taewoon Kim, Tali
  Bers, Thibault Fevry, Trishala Neeraj, Urmish Thakker, Vikas Raunak, Xiangru
  Tang, Zheng-Xin Yong, Zhiqing Sun, Shaked Brody, Yallow Uri, Hadar Tojarieh,
  Adam Roberts, Hyung~Won Chung, Jaesung Tae, Jason Phang, Ofir Press, Conglong
  Li, Deepak Narayanan, Hatim Bourfoune, Jared Casper, Jeff Rasley, Max
  Ryabinin, Mayank Mishra, Minjia Zhang, Mohammad Shoeybi, Myriam Peyrounette,
  Nicolas Patry, Nouamane Tazi, Omar Sanseviero, Patrick von Platen, Pierre
  Cornette, Pierre~François Lavallée, Rémi Lacroix, Samyam Rajbhandari,
  Sanchit Gandhi, Shaden Smith, Stéphane Requena, Suraj Patil, Tim Dettmers,
  Ahmed Baruwa, Amanpreet Singh, Anastasia Cheveleva, Anne-Laure Ligozat, Arjun
  Subramonian, Aurélie Névéol, Charles Lovering, Dan Garrette, Deepak
  Tunuguntla, Ehud Reiter, Ekaterina Taktasheva, Ekaterina Voloshina, Eli
  Bogdanov, Genta~Indra Winata, Hailey Schoelkopf, Jan-Christoph Kalo,
  Jekaterina Novikova, Jessica~Zosa Forde, Jordan Clive, Jungo Kasai, Ken
  Kawamura, Liam Hazan, Marine Carpuat, Miruna Clinciu, Najoung Kim, Newton
  Cheng, Oleg Serikov, Omer Antverg, Oskar van~der Wal, Rui Zhang, Ruochen
  Zhang, Sebastian Gehrmann, Shachar Mirkin, Shani Pais, Tatiana Shavrina,
  Thomas Scialom, Tian Yun, Tomasz Limisiewicz, Verena Rieser, Vitaly Protasov,
  Vladislav Mikhailov, Yada Pruksachatkun, Yonatan Belinkov, Zachary Bamberger,
  Zdeněk Kasner, Alice Rueda, Amanda Pestana, Amir Feizpour, Ammar Khan, Amy
  Faranak, Ana Santos, Anthony Hevia, Antigona Unldreaj, Arash Aghagol, Arezoo
  Abdollahi, Aycha Tammour, Azadeh HajiHosseini, Bahareh Behroozi, Benjamin
  Ajibade, Bharat Saxena, Carlos~Muñoz Ferrandis, Daniel McDuff, Danish
  Contractor, David Lansky, Davis David, Douwe Kiela, Duong~A. Nguyen, Edward
  Tan, Emi Baylor, Ezinwanne Ozoani, Fatima Mirza, Frankline Ononiwu, Habib
  Rezanejad, Hessie Jones, Indrani Bhattacharya, Irene Solaiman, Irina Sedenko,
  Isar Nejadgholi, Jesse Passmore, Josh Seltzer, Julio~Bonis Sanz, Livia Dutra,
  Mairon Samagaio, Maraim Elbadri, Margot Mieskes, Marissa Gerchick, Martha
  Akinlolu, Michael McKenna, Mike Qiu, Muhammed Ghauri, Mykola Burynok, Nafis
  Abrar, Nazneen Rajani, Nour Elkott, Nour Fahmy, Olanrewaju Samuel, Ran An,
  Rasmus Kromann, Ryan Hao, Samira Alizadeh, Sarmad Shubber, Silas Wang, Sourav
  Roy, Sylvain Viguier, Thanh Le, Tobi Oyebade, Trieu Le, Yoyo Yang, Zach
  Nguyen, Abhinav~Ramesh Kashyap, Alfredo Palasciano, Alison Callahan, Anima
  Shukla, Antonio Miranda-Escalada, Ayush Singh, Benjamin Beilharz, Bo~Wang,
  Caio Brito, Chenxi Zhou, Chirag Jain, Chuxin Xu, Clémentine Fourrier,
  Daniel~León Periñán, Daniel Molano, Dian Yu, Enrique Manjavacas, Fabio
  Barth, Florian Fuhrimann, Gabriel Altay, Giyaseddin Bayrak, Gully Burns,
  Helena~U. Vrabec, Imane Bello, Ishani Dash, Jihyun Kang, John Giorgi, Jonas
  Golde, Jose~David Posada, Karthik~Rangasai Sivaraman, Lokesh Bulchandani,
  Lu~Liu, Luisa Shinzato, Madeleine~Hahn de~Bykhovetz, Maiko Takeuchi, Marc
  Pàmies, Maria~A Castillo, Marianna Nezhurina, Mario Sänger, Matthias
  Samwald, Michael Cullan, Michael Weinberg, Michiel~De Wolf, Mina Mihaljcic,
  Minna Liu, Moritz Freidank, Myungsun Kang, Natasha Seelam, Nathan Dahlberg,
  Nicholas~Michio Broad, Nikolaus Muellner, Pascale Fung, Patrick Haller, Ramya
  Chandrasekhar, Renata Eisenberg, Robert Martin, Rodrigo Canalli, Rosaline Su,
  Ruisi Su, Samuel Cahyawijaya, Samuele Garda, Shlok~S Deshmukh, Shubhanshu
  Mishra, Sid Kiblawi, Simon Ott, Sinee Sang-aroonsiri, Srishti Kumar, Stefan
  Schweter, Sushil Bharati, Tanmay Laud, Théo Gigant, Tomoya Kainuma, Wojciech
  Kusa, Yanis Labrak, Yash~Shailesh Bajaj, Yash Venkatraman, Yifan Xu, Yingxin
  Xu, Yu~Xu, Zhe Tan, Zhongli Xie, Zifan Ye, Mathilde Bras, Younes Belkada, and
  Thomas Wolf. 2023.
\newblock \href {https://arxiv.org/abs/2211.05100} {Bloom: A 176b-parameter
  open-access multilingual language model}.
\newblock \emph{Preprint}, arXiv:2211.05100.

\bibitem[{Sch{\"u}tze(1996)}]{schutze1996empirical}
Carson~T. Sch{\"u}tze. 1996.
\newblock \emph{{The Empirical Base of Linguistics: Grammaticality Judgments
  and Linguistic Methodology}}.
\newblock University of Chicago Press.

\bibitem[{Serikov et~al.(2022)Serikov, Protasov, Voloshina, Knyazkova, and
  Shavrina}]{serikov-etal-2022-universal}
Oleg Serikov, Vitaly Protasov, Ekaterina Voloshina, Viktoria Knyazkova, and
  Tatiana Shavrina. 2022.
\newblock \href {https://doi.org/10.18653/v1/2022.blackboxnlp-1.37} {Universal
  and independent: Multilingual probing framework for exhaustive model
  interpretation and evaluation}.
\newblock In \emph{Proceedings of the Fifth BlackboxNLP Workshop on Analyzing
  and Interpreting Neural Networks for NLP}, pages 441--456, Abu Dhabi, United
  Arab Emirates (Hybrid). Association for Computational Linguistics.

\bibitem[{Shi et~al.(2023)Shi, Ajith, Xia, Huang, Liu, Blevins, Chen, and
  Zettlemoyer}]{shi2023detecting}
Weijia Shi, Anirudh Ajith, Mengzhou Xia, Yangsibo Huang, Daogao Liu, Terra
  Blevins, Danqi Chen, and Luke Zettlemoyer. 2023.
\newblock {Detecting Pretraining Data from Large Language Models}.
\newblock In \emph{The Twelfth International Conference on Learning
  Representations}.

\bibitem[{Shliazhko et~al.(2024)Shliazhko, Fenogenova, Tikhonova, Kozlova,
  Mikhailov, and Shavrina}]{shliazhko2024mgpt}
Oleh Shliazhko, Alena Fenogenova, Maria Tikhonova, Anastasia Kozlova, Vladislav
  Mikhailov, and Tatiana Shavrina. 2024.
\newblock mgpt: Few-shot learners go multilingual.
\newblock \emph{Transactions of the Association for Computational Linguistics},
  12:58--79.

\bibitem[{Sichinava(2018)}]{sichinava2018preposition}
D.~Sichinava. 2018.
\newblock Preposition.
\newblock Materials for the project of corpus description of Russian grammar
  (http://rusgram.ru).

\bibitem[{Slioussar and Malko(2016)}]{slioussar2016gender}
Natalia Slioussar and Anton Malko. 2016.
\newblock Gender agreement attraction in russian: production and comprehension
  evidence.
\newblock \emph{Frontiers in psychology}, 7:166019.

\bibitem[{Someya and Oseki(2023)}]{someya-oseki-2023-jblimp}
Taiga Someya and Yohei Oseki. 2023.
\newblock \href {https://doi.org/10.18653/v1/2023.findings-eacl.117}
  {{JBL}i{MP}: {J}apanese benchmark of linguistic minimal pairs}.
\newblock In \emph{Findings of the Association for Computational Linguistics:
  EACL 2023}, pages 1581--1594, Dubrovnik, Croatia. Association for
  Computational Linguistics.

\bibitem[{Song et~al.(2022)Song, Krishna, Bhatt, and
  Iyyer}]{song-etal-2022-sling}
Yixiao Song, Kalpesh Krishna, Rajesh Bhatt, and Mohit Iyyer. 2022.
\newblock \href {https://doi.org/10.18653/v1/2022.emnlp-main.305} {{SLING}:
  {S}ino linguistic evaluation of large language models}.
\newblock In \emph{Proceedings of the 2022 Conference on Empirical Methods in
  Natural Language Processing}, pages 4606--4634, Abu Dhabi, United Arab
  Emirates. Association for Computational Linguistics.

\bibitem[{Stassen(2013)}]{stassen2013predicative}
Leon Stassen. 2013.
\newblock \href {https://doi.org/10.5281/zenodo.7385533} {Predicative
  possession}.
\newblock In Matthew~S. Dryer and Martin Haspelmath, editors, \emph{WALS
  Online}, v2020.3 edition. Zenodo.
\newblock Available online at \url{http://wals.info/chapter/117}, Accessed on
  2023-11-16.

\bibitem[{Testelets(2001)}]{Testelets:2001}
Yakov Testelets. 2001.
\newblock \emph{{Vvedeniye v obschiy sintaksis}}.
\newblock Russian State University for the Humanities.

\bibitem[{Touvron et~al.(2023)Touvron, Martin, Stone, Albert, Almahairi,
  Babaei, Bashlykov, Batra, Bhargava, Bhosale et~al.}]{touvron2023llama}
Hugo Touvron, Louis Martin, Kevin Stone, Peter Albert, Amjad Almahairi, Yasmine
  Babaei, Nikolay Bashlykov, Soumya Batra, Prajjwal Bhargava, Shruti Bhosale,
  et~al. 2023.
\newblock Llama 2: Open foundation and fine-tuned chat models.
\newblock \emph{arXiv preprint arXiv:2307.09288}.

\bibitem[{Ustalov et~al.(2024)Ustalov, Pavlichenko, and
  Tseitlin}]{ustalov2024learning}
Dmitry Ustalov, Nikita Pavlichenko, and Boris Tseitlin. 2024.
\newblock {Learning from Crowds with Crowd-Kit}.
\newblock \emph{Journal of Open Source Software}, 9(96):6227.

\bibitem[{Vaswani et~al.(2017)Vaswani, Shazeer, Parmar, Uszkoreit, Jones,
  Gomez, Kaiser, and Polosukhin}]{vaswani2017attention}
Ashish Vaswani, Noam Shazeer, Niki Parmar, Jakob Uszkoreit, Llion Jones,
  Aidan~N Gomez, {\L}ukasz Kaiser, and Illia Polosukhin. 2017.
\newblock {Attention is All you Need}.
\newblock In \emph{{Advances in Neural Information Processing Systems}}, pages
  5998--6008.

\bibitem[{Volodina et~al.(2021)Volodina, Mohammed, and
  Klezl}]{volodina-etal-2021-dalaj}
Elena Volodina, Yousuf~Ali Mohammed, and Julia Klezl. 2021.
\newblock \href {https://aclanthology.org/2021.nlp4call-1.3} {{D}a{LAJ} {--} a
  dataset for linguistic acceptability judgments for {S}wedish}.
\newblock In \emph{Proceedings of the 10th Workshop on NLP for Computer
  Assisted Language Learning}, pages 28--37, Online. LiU Electronic Press.

\bibitem[{Warstadt and Bowman(2022)}]{warstadt2022artificial}
Alex Warstadt and Samuel~R Bowman. 2022.
\newblock What artificial neural networks can tell us about human language
  acquisition.
\newblock In \emph{Algebraic structures in natural language}, pages 17--60. CRC
  Press.

\bibitem[{Warstadt et~al.(2019)Warstadt, Cao, Grosu, Peng, Blix, Nie, Alsop,
  Bordia, Liu, Parrish, Wang, Phang, Mohananey, Htut, Jeretic, and
  Bowman}]{warstadt-etal-2019-investigating}
Alex Warstadt, Yu~Cao, Ioana Grosu, Wei Peng, Hagen Blix, Yining Nie, Anna
  Alsop, Shikha Bordia, Haokun Liu, Alicia Parrish, Sheng-Fu Wang, Jason Phang,
  Anhad Mohananey, Phu~Mon Htut, Paloma Jeretic, and Samuel~R. Bowman. 2019.
\newblock \href {https://doi.org/10.18653/v1/D19-1286} {Investigating
  {BERT}{'}s knowledge of language: Five analysis methods with {NPI}s}.
\newblock In \emph{Proceedings of the 2019 Conference on Empirical Methods in
  Natural Language Processing and the 9th International Joint Conference on
  Natural Language Processing (EMNLP-IJCNLP)}, pages 2877--2887, Hong Kong,
  China. Association for Computational Linguistics.

\bibitem[{Warstadt et~al.(2020)Warstadt, Parrish, Liu, Mohananey, Peng, Wang,
  and Bowman}]{warstadt-etal-2020-blimp-benchmark}
Alex Warstadt, Alicia Parrish, Haokun Liu, Anhad Mohananey, Wei Peng, Sheng-Fu
  Wang, and Samuel~R. Bowman. 2020.
\newblock \href {https://doi.org/10.1162/tacl_a_00321} {{BL}i{MP}: The
  benchmark of linguistic minimal pairs for {E}nglish}.
\newblock \emph{Transactions of the Association for Computational Linguistics},
  8:377--392.

\bibitem[{Wilcox et~al.(2018)Wilcox, Levy, Morita, and
  Futrell}]{wilcox-etal-2018-rnn}
Ethan Wilcox, Roger Levy, Takashi Morita, and Richard Futrell. 2018.
\newblock \href {https://doi.org/10.18653/v1/W18-5423} {What do {RNN} language
  models learn about filler{--}gap dependencies?}
\newblock In \emph{Proceedings of the 2018 {EMNLP} Workshop {B}lackbox{NLP}:
  Analyzing and Interpreting Neural Networks for {NLP}}, pages 211--221,
  Brussels, Belgium. Association for Computational Linguistics.

\bibitem[{Wolf et~al.(2020)Wolf, Debut, Sanh, Chaumond, Delangue, Moi, Cistac,
  Rault, Louf, Funtowicz, Davison, Shleifer, von Platen, Ma, Jernite, Plu, Xu,
  Le~Scao, Gugger, Drame, Lhoest, and Rush}]{wolf-etal-2020-transformers}
Thomas Wolf, Lysandre Debut, Victor Sanh, Julien Chaumond, Clement Delangue,
  Anthony Moi, Pierric Cistac, Tim Rault, Remi Louf, Morgan Funtowicz, Joe
  Davison, Sam Shleifer, Patrick von Platen, Clara Ma, Yacine Jernite, Julien
  Plu, Canwen Xu, Teven Le~Scao, Sylvain Gugger, Mariama Drame, Quentin Lhoest,
  and Alexander Rush. 2020.
\newblock \href {https://doi.org/10.18653/v1/2020.emnlp-demos.6} {Transformers:
  State-of-the-art natural language processing}.
\newblock In \emph{Proceedings of the 2020 Conference on Empirical Methods in
  Natural Language Processing: System Demonstrations}, pages 38--45, Online.
  Association for Computational Linguistics.

\bibitem[{Xiang et~al.(2021)Xiang, Yang, Li, Warstadt, and
  Kann}]{xiang-etal-2021-climp}
Beilei Xiang, Changbing Yang, Yu~Li, Alex Warstadt, and Katharina Kann. 2021.
\newblock \href {https://doi.org/10.18653/v1/2021.eacl-main.242} {{CL}i{MP}: A
  benchmark for {C}hinese language model evaluation}.
\newblock In \emph{Proceedings of the 16th Conference of the European Chapter
  of the Association for Computational Linguistics: Main Volume}, pages
  2784--2790, Online. Association for Computational Linguistics.

\bibitem[{Zaliznyak(1987)}]{Zaliznyak1987}
Andrey Zaliznyak. 1987.
\newblock \emph{Grammatical Dictionary of Russian Language: Word Inflection}.
\newblock Moscow.

\bibitem[{Zhang et~al.(2024)Zhang, Sun, Yeats, Ouyang, Kuo, Zhang, Yang, and
  Li}]{zhang2024min}
Jingyang Zhang, Jingwei Sun, Eric Yeats, Yang Ouyang, Martin Kuo, Jianyi Zhang,
  Hao Yang, and Hai Li. 2024.
\newblock {Min-K\%++: Improved Baseline for Detecting Pre-Training Data from
  Large Language Models}.
\newblock \emph{arXiv preprint arXiv:2404.02936}.

\bibitem[{Zmitrovich et~al.(2024)Zmitrovich, Abramov, Kalmykov, Kadulin,
  Tikhonova, Taktasheva, Astafurov, Baushenko, Snegirev, Shavrina, Markov,
  Mikhailov, and Fenogenova}]{zmitrovich-etal-2024-family-pretrained}
Dmitry Zmitrovich, Aleksandr Abramov, Andrey Kalmykov, Vitaly Kadulin, Maria
  Tikhonova, Ekaterina Taktasheva, Danil Astafurov, Mark Baushenko, Artem
  Snegirev, Tatiana Shavrina, Sergei~S. Markov, Vladislav Mikhailov, and Alena
  Fenogenova. 2024.
\newblock \href {https://aclanthology.org/2024.lrec-main.45} {A family of
  pretrained transformer language models for {R}ussian}.
\newblock In \emph{Proceedings of the 2024 Joint International Conference on
  Computational Linguistics, Language Resources and Evaluation (LREC-COLING
  2024)}, pages 507--524, Torino, Italia. ELRA and ICCL.

\end{thebibliography}
